\documentclass{article}

\usepackage[preprint,nonatbib]{neurips_2024}

\newcommand{\ie}{\emph{i.e., }}
\newcommand{\eg}{\emph{e.g., }}
\usepackage[utf8]{inputenc} 
\usepackage[T1]{fontenc}    
\usepackage{hyperref}       
\usepackage{url}            
\usepackage{amsmath}
\usepackage{booktabs}       
\usepackage{amsfonts}       
\usepackage{nicefrac}       
\usepackage{microtype}      
\usepackage{xcolor}         

\usepackage[utf8]{inputenc} 
\usepackage[T1]{fontenc}    
\usepackage{hyperref}       
\usepackage{url}            
\usepackage{booktabs}       
\usepackage{amsfonts}       
\usepackage{nicefrac}       
\usepackage{microtype}      
\usepackage{xcolor}         
\usepackage{pdfpages}

\usepackage{xspace}
\usepackage{array}
\usepackage{graphicx}
\usepackage{multirow,array}
\usepackage{multicol}
\usepackage{subcaption}
\hypersetup{
	colorlinks=true,%
	citecolor={blue!50!black},
	linkcolor={red!50!black},
	urlcolor={green!50!black}
}

\usepackage{csquotes}
\usepackage{tcolorbox}
\usepackage{wrapfig}
\usepackage[font=small]{caption}
\usepackage[T1]{fontenc}
\usepackage{enumitem}
\usepackage{listings}

\definecolor{codegreen}{rgb}{0,0.6,0}
\definecolor{codegray}{rgb}{0.5,0.5,0.5}
\definecolor{codepink}{RGB}{252, 142, 172}
\definecolor{codepurple}{rgb}{0.58,0,0.82}
\definecolor{backcolour}{RGB}{245,245,245}
\lstdefinestyle{mystyle}{
    backgroundcolor=\color{backcolour},   
    commentstyle=\color{magenta},
    keywordstyle=\color{blue},
    numberstyle=\tiny\color{codegray},
    stringstyle=\color{codepurple},
    basicstyle=\fontfamily{\ttdefault}\footnotesize,
    breakatwhitespace=false,        
    breaklines=true,                
    keepspaces=true,    
    frame=single,
    numbersep=5pt,                  
    showspaces=false,              
    showstringspaces=false,
    showtabs=false,               
    tabsize=2,
    classoffset=1, 
    keywordstyle=\color{violet},
    classoffset=0,
}
\hypersetup{
    colorlinks=true
}

\title{Egocentric Vision Language Planning}
\author{%
    Zhirui Fang\textsuperscript{\rm 2}\thanks{Work performed while an intern at BAAI.}, Ming Yang\textsuperscript{\rm 3}, Weishuai Zeng\textsuperscript{\rm 3}, Boyu Li\textsuperscript{\rm 1}$^*$, Junpeng Yue\textsuperscript{\rm 3}$^*$, \\ \textbf{Ziluo Ding}\textsuperscript{\rm 1}, \textbf{Xiu Li}\textsuperscript{\rm 2}\textbf{,} \textbf{Zongqing Lu}\textsuperscript{\rm 3,1} \\
    \textsuperscript{\rm 1} Beijing Academy of Artificial Intelligence (BAAI)\\
    \textsuperscript{\rm 2} Tsinghua Shenzhen International Graduate School, Tsinghua University \\
   \textsuperscript{\rm 3} School of Computer Science, Peking University\\
   \texttt{\{fzr23@mails.tsinghua.edu.cn, ziluoding@baai.ac.cn} \\
   \texttt{zongqing.lu@pku.edu.cn\}}
}

\begin{document}

\maketitle

\begin{abstract}
We explore leveraging large multi-modal models (LMMs) and text2image models to build a more general embodied agent. LMMs excel in planning long-horizon tasks over symbolic abstractions but struggle with grounding in the physical world, often failing to accurately identify object positions in images. A bridge is needed to connect LMMs to the physical world. The paper proposes a novel approach, egocentric vision language planning (EgoPlan), to handle long-horizon tasks from an egocentric perspective in varying household scenarios. This model leverages a diffusion model to simulate the fundamental dynamics between states and actions, integrating techniques like style transfer and optical flow to enhance generalization across different environmental dynamics. The LMM serves as a planner, breaking down instructions into sub-goals and selecting actions based on their alignment with these sub-goals, thus enabling more generalized and effective decision-making. Experiments show that EgoPlan improves long-horizon task success rates from the egocentric view compared to baselines across household scenarios.

\end{abstract}

\section{Introduction}
The advent of large language models (LLMs)~\cite{openai2024gpt4, touvron2023llama} and large multi-modal models (LMMs)~\cite{2023GPT4VisionSC, girdhar2023imagebind, zhang2023videollama,zhu2023minigpt4} has revolutionized the field of artificial intelligence. Their strong reasoning~\cite{wang2023selfconsistency,wei2023chainofthought} and powerful generalization capabilities allow them to be directly applied in various scenarios. In the next step toward artificial general intelligence (AGI), researchers are considering enabling large models (LMs), especially LMMs, to break through the world expressed by text and images to interact with the physical world. They aim to build a general embodied agent that intelligently interacts with the physical world.

LMMs have demonstrated an impressive capability of planning for long-horizon tasks over symbolic abstraction in the physical world~\cite{wake2024gpt4vision}. However, there's still a piece of the puzzle missing. They have struggled to ground the text world with the physical world. For example, GPT-4V often fails to accurately identify objects' positions in images. LMMs seem to know \textit{what to do next} but do not understand \textit{how the world works}. A world model (dynamics model) is hence needed to connect the LMMs to the physical world. There are two potential solutions. One is to implicitly integrate environmental dynamics into the LMMs, that is, fine-tuning the LMMs based on a vast amount of state-action sequences, such as PaLM-E~\cite{driess2023palme} and RT-2~\cite{brohan2023rt2}. However, directly training large models requires extensive data and computational resources. The other is to explicitly introduce a pre-trained world model, \eg text2image models~\cite{radford2021learning,saharia2022photorealistic}, which can be used by LMMs as an auxiliary tool. Our work explores the second path. We try to answer the question: \textbf{\textit{how do we leverage the LMMs and text2image model to build a more general embodied agent?}}

Some works already train text2image/video models as world models for decision-making. However, there still exist several limitations. First, their task scenarios often involve object manipulation, a fully observable setting. This is uncommon in real-world scenarios, and their methods seem to struggle to adapt to other practical scenes. For example, SuSIE~\cite{black2023zeroshot} and VLP~\cite{du2023video} require generating images several steps ahead, yet the error introduced by long-range predictions is substantial for most partially observed scenarios, \eg autonomous driving.
In contrast, we focus on a more challenging, partially observable setting. The embodied agent, like humans, tends to complete more complex tasks, \eg household tasks, from the egocentric view. Second, their framework has limited generalization capability, mainly reflected in two aspects: (i) Their low-level policies are tailored to specific tasks, and the different dynamics may lead to policy failure; (ii) The dynamics can vary for the same action described by the text, \eg turn left. This is because individuals from different environments, e.g., simulators or the physical world, exhibit differences. The text2image/text2video model lacks individual motion pattern information and cannot be generalized accurately to dynamics of other environments that are out of the training dataset. We hope the agent can generalize to different dynamics within the same type of scenario, \eg household scenario.

In this work, we propose egocentric vision language planning (\textbf{EgoPlan}), a general embodied agent to perform long-horizon tasks from the egocentric view across different household environments. Under the egocentric view setting, predicting an observation even a few steps ahead is unreliable. Hence, text2image model, \ie diffusion model, is adopted to realize the fundamental dynamics model under the partially observable setting, where observation and action are represented by image and text respectively. Furthermore, the main differences between the dynamics of the two different environments stem from two aspects: the style of the environment is different, and the motion pattern, \eg amplitude, of the same type of action is different. To accurately generalize the text2image model to other environments, we can perform style transfer based on LoRA~\cite{hu2021lora} and introduce optical flow into the model to guide the motion pattern. 

For decision-making, we require a generalized policy; hence, we prompt the LMM as the planner. In more detail, given the instruction, the LMM first decomposes it into many sub-goals. For sub-goal representation, we explore text-based and image-based approaches and analyze each form. After knowing the outcome of each action based on the dynamics model, the LMM can choose the proper action by judging which outcome is closer to the current sub-goal. Intuitively, if the planner and dynamics model possess a certain degree of generalization ability, the agent also inherits this ability.

We conduct a comprehensive evaluation and analysis of each module of the embodied agent. Empirically, we demonstrate the high quality of image generation by the world model and the high accuracy of optical flow prediction. Subsequently, we verify the world model's effectiveness in aiding decision-making in more complex tasks. Lastly, we confirm the method’s generalization capabilities in a different environment.
Our major contributions are summarized as follows:
\begin{itemize}[leftmargin=*]
\item We have collected a dataset on Virtualhome, which views an action of the agent as a trajectory and provides egocentric observations, visualising optical flow, depth maps and semantic segmentation maps at each time step in the trajectory, which will provide data support for navigation and manipulation tasks in the embodied environment.
\item We propose \textbf{EgoPlan}, a framework for complex task planning that combines LMM and a world model that predicts an egocentric view of the scene at the next time step after an action is executed. In order to plan more complex tasks with more diverse and different viewpoints (a composite task includes navigation and production tasks), we limit the prediction step size of the world model to avoid the complexity explosion of the prediction algorithm, and introduce optical flow information into the world model to make the world model more sensitive to action position changes and adapt to scene changes during navigation. We demonstrate the effectiveness of our framework through LMM$+$world model planning experiments on comprehensive tasks.
\item 
The egocentric observation considering different actions in different environments consists of the agent motion itself and the fine-grained background information of the environment. The optical flow information represents the motion information itself, which is computationally invariant to different scenes and styles, while the fine-grained background information of the environment can be fine-tuned by a small number of sample images from the perspective of the environment agent. We borrow the idea of style transfer in computer vision and adopt the Lora model to fine-tune our diffusion world model, so as to achieve the ability of our framework to achieve few-shot generalization in different embodied scenarios. Experiments on habitat show that our framework can still assist multi-modal large models for task planning in different environments.
\end{itemize}

\section{Related Work}

\subsection{Diffusion Model}
The diffusion model~\cite{ho2020denoising, song2022denoising} has been extensively studied in the field of image generation~\cite{dhariwal2021diffusion,ho2021cascaded,rombach2022highresolution} and image editing~\cite{gal2022image,hertz2022prompttoprompt,meng2022sdedit}. Diffusion models can achieve a high degree of control during the image generation. In more detail, InstructPix2Pix (InstructP2P)~\cite{brooks2023instructpix2pix} trains a conditional diffusion model that, given an input image and text instruction for how to edit it, generates the edited image. ControlNet~\cite{zhang2023adding} is widely used to control the style of the generated image by using various forms of prior information, \eg edge information and segmentation. By adding LoRA or adapter~\cite{houlsby2019parameterefficient} modules to the network, the model trained on one data distribution can also be transferred to other data distributions (different visual styles) through a few picture examples.

The images produced by current diffusion models are of very high quality, highly realistic, and easily controllable. It prompts various fields to consider using these generated images to assist in accomplishing other tasks. Our paper adopts the diffusion model to generate task subgoals and predict the image of the next state for decision-making. 



\subsection{World Model for Decision-making}
The world model is used to model the dynamics of the environment. It is crucial for building autonomous agents and enabling intelligent interactions in various scenarios. However, developing a precise world model remains a significant challenge in model-based decision-making.

The Dreamer series~\cite{hafner2020dream,hafner2022mastering,hafner2024mastering} model the environment dynamics in the latent space to predict future states within gaming environments. It enables game agents to learn the tasks by imagination, decreasing the interactions for effective learning. However, since the world models are learned in latent space instead of pixel space, these models often lack the generalization of unseen tasks and environments. A world model built in the pixel space may have better generalization capabilities. By carefully orchestrating rich data along different axes, UniSim~\cite{yang2023unisim} can simulate realistic experiences in the visual form in response to actions by humans, robots, and other interactive agents. In summary, we can find out the versatile applications of world models span beyond games and robotics.

The advancements in diffusion-based world models are reshaping how we model physical motion laws in real-world settings, particularly in robotics. UniPi~\cite{du2023learning} frames the decision-making problem in robotics as a text2video task. The generated video is fed into an inverse dynamics model (IDM) that extracts underlying low-level control actions, which are executed in simulation or by a real robot agent. Video Language Planning (VLP)~\cite{du2023video} introduces a novel method for task planning that integrates video generation with tree search algorithms. This methodology lets robots plan over longer horizons by visualizing future actions and outcomes. Unlike previous works, SuSIE~\cite{black2023zeroshot} leverages pre-trained image-editing models to predict the hypothetical future frame. A low-level goal-reaching policy is trained on robot data to reach this hypothetical future frame. Since one goal frame prediction does not require the model to understand the intricacies of the robot’s low-level precisely dynamics, it should facilitate transfer from other data sources, \eg human videos. RoboDreamer~\cite{zhou2024robodreamer} advances the field by utilizing video diffusion to formulate plans combining actions and objects, solving novel tasks in unexplored robotic environments.

We find it unrealistic to apply the text2video model to partially observed scenarios. Moreover, it is still hard to predict the goal frame several steps ahead, as the shift in perspective could be significant. Therefore, we adopt the text2image model to accurately predict the short-range outcome for one-step planning.


\begin{figure*}[t]
    \centering
    \includegraphics[width=\linewidth]{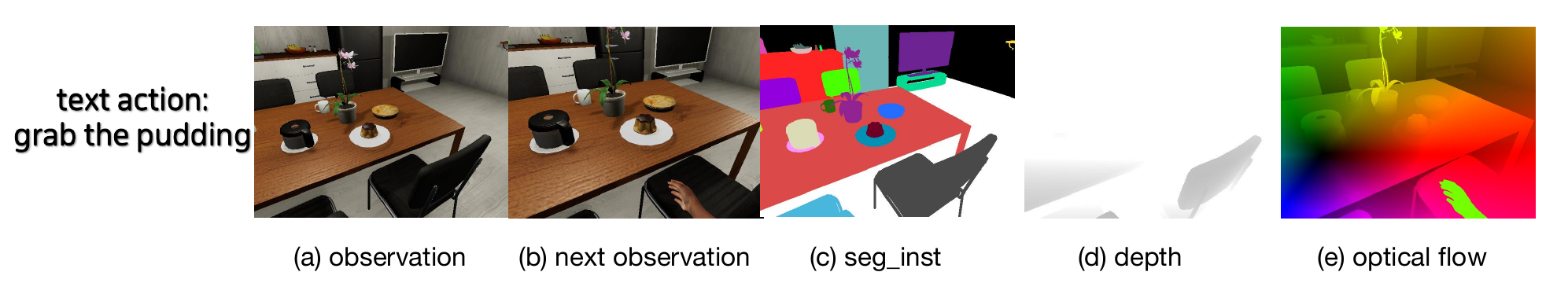}
    \caption{An illustration sample in VH-1.5M, which includes current image observation, next image observation given the text action, semantic segmentation map, depth map, and optical flow map.}
    \label{fig:sample}
    \vspace{-0.2cm}
\end{figure*}

\section{VH-1.5M Dataset}

Most datasets related to embodied agents, \eg RT-X~\cite{embodimentcollaboration2024open} and RH20T~\cite{fang2023rh20t}, employ the third-person view to avoid the visual occlusion issue, thus lacking data regarding the egocentric view (first-person view). There are some datasets, \eg Alfred~\cite{shridhar2020alfred} and Procthor~\cite{deitke2022procthor}, that adopt a first-person perspective, however, they simplify the state transition by assuming instantaneous completion of actions, which fails to mimic the dynamics changes in real-world environments. We propose the VH-1.5M dataset based on the VirtualHome~\cite{puig2018virtualhome, puig2020watchandhelp} environment to address these limitations.



We construct our dataset VH-1.5M in the VirtualHome environment, which comprises 50 distinct houses. Each house contains approximately 300 interactive objects, and the embodied agent can perform more than 10 actions. Note that the VirtualHome environment is a simulator tailored for embodied agents, offering a detailed simulation of a residential living scenario. It enables a range of household tasks, \eg navigation and object manipulation.



The VH-1.5M dataset is organized in a structured manner, encapsulating the relationship between actions, houses, agents, and trajectories. Each task sequence entry follows a hierarchical structure, \eg "/open/house\_0/Female4/2\_fridge" (female4 open the fridge2 in house0). 

\textbf{Dataset Details:} The VH-1.5M dataset consists of:
\begin{itemize}
    \item 13 Actions: Various physical actions and interactions for Agents within the Houses.
    \item 50 Houses: Uniquely designed houses with diverse layouts and object placements.
    \item 4 Agents: Four distinct agents, each capable of performing the full range of actions.
    \item 1.5M Samples: Dateset has numerous detailed sequences, each executing one action. Information from each step in the sequence is stored as one sample. One example is shown in Figure \ref{fig:sample}. We use \textit{House49} as the validation set. 
\end{itemize}

More details of the dataset can be found in the appendix, and \textit{we will open-source the dataset.}



\section{Method}

Our embodied agent, EgoPlan, takes input as a visual observation \(x_t\) of the scene at the current timestep \(t\) and a natural language goal \(g\) and outputs an action \(a_t\) to interact with the environment. Note that the \(x_t\) only partially represents the current environment state. In addition, the agent uses encapsulated skills as actions, such as moving forward, turning, and grabbing objects. 

\begin{figure}[t]
  \centering
  \includegraphics[width=1\textwidth]{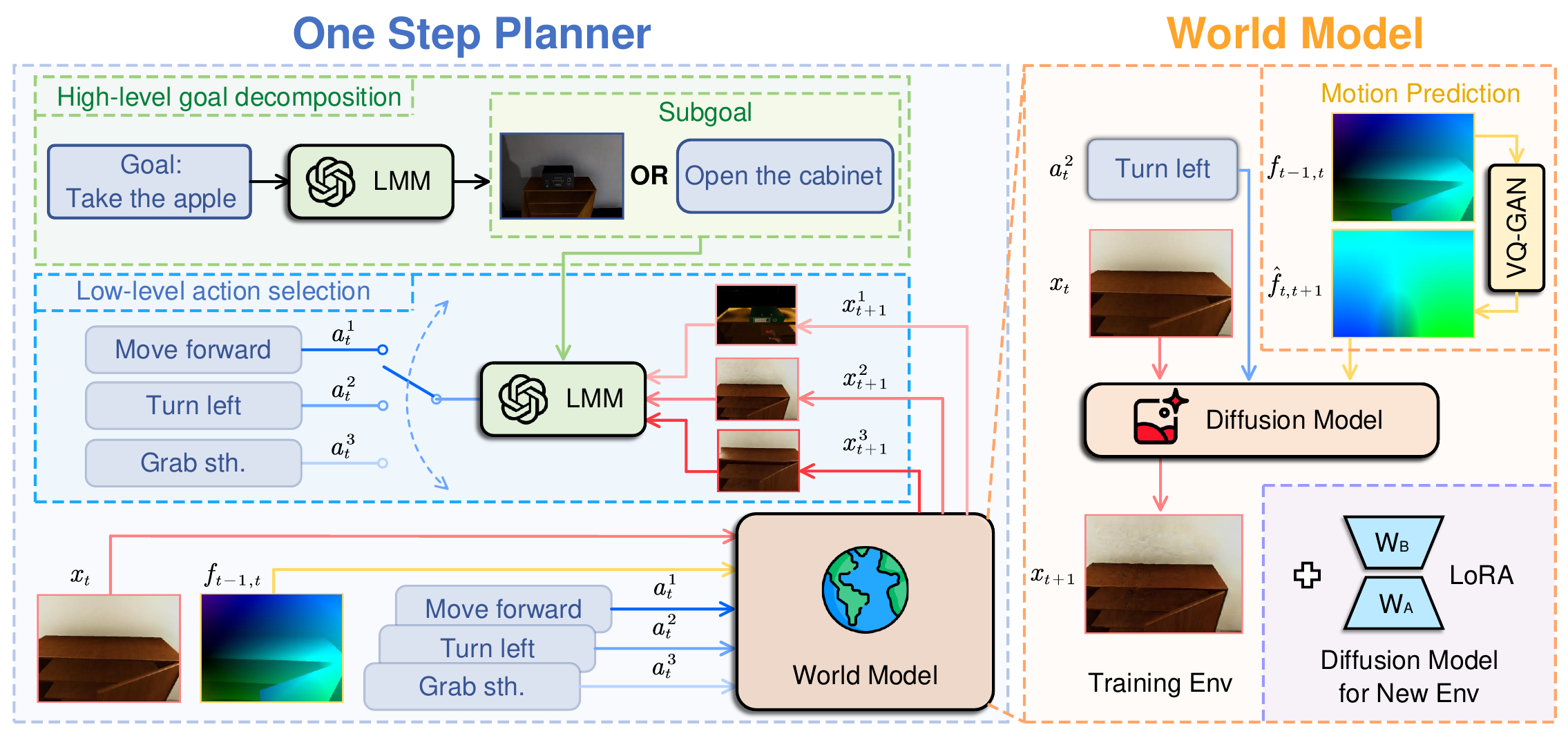}
  \caption{Overview of EgoPlan. The left side features a one-step planner that provides the agent with decision-making capabilities, while the right side includes a world model (dynamics model) that provides the agent with an understanding of the current environment.}
  \label{fig:framework}
\vspace{-0.2cm}
\end{figure}

EgoPlan consists of two parts, as illustrated in Figure \ref{fig:framework}. One is a dynamics model that gives the agent the concept of the current environment, and the other is the planner that endows the agent with decision-making capabilities. Intuitively, we humans first envision the outcomes of each action in our minds, and then, by comparing the results, we make the best decision.

\subsection{Diffusion-Based Dynamics Model}

\subsubsection{Learning Dynamics}\label{LD}
From a first-person perspective, the view after two steps may be completely different, making it difficult to model. Therefore, we aim to model the fundamental dynamics model, \(p_{\theta}(x_{t+1}|x_t,a_t)\), for one-step planning usage. In more detail, we want to generate a new image \(x_{t+1}\), representing the next state given the current visual observation \(x_t\) and the text of the action \(a_t\). Then, we cast our eyes on the text2image model and resort to the diffusion model for modeling specifically. It has an irreplaceable advantage in easily incorporating other modalities as a condition. 

Although the open-sourced diffusion model~\cite{ho2022imagen,luo2023videofusion}, \(p_{\theta}(x_{\rm tar}|x_{\rm src}, l)\), trained on a wealth of online videos, has demonstrated the ability to predict the future, their generated results are hard to control, and most are only semantically reasonable. Moreover, most of the text in the pre-trained dataset consists of image descriptions \(l\) rather than action instructions \(a\). Therefore, supervised fine-tuning is adopted based on our VH-1.5M dataset to better model the dynamics, \(p_{\theta_{\rm sft}}(x_{t+1}|x_t,a_t)\). Formally, the training objective is given by:
\begin{align}
\footnotesize
    \mathcal{L}_{\text{MSE}} &=\left\|\epsilon-\epsilon_{\theta}\left(q\left(x_{t+1}^{(k)}|x_t,a_t\right),k\right)\right\|^2  \\
     &= \left\|\epsilon-\epsilon_{\theta}\left(\sqrt{\overline{\alpha_t}}x_t+\sqrt{1-\overline{\alpha_t}}\epsilon|a_{t}\right)\right\|^2
\end{align}
where $\epsilon_{\theta}$ is a learnable denoising model for reverse process, $k$ is denoising steps, and $\overline{\alpha_t}$ are a set of $K$ different noise levels for each $k \in [1,K]$.
However, we find it difficult to generalize directly to other environments since our dataset only includes VirtualHome scenes. The difference between two environments, \eg Habitat2.0~\cite{savva2019habitat,szot2022habitat} and VirtualHome, primarily lies in their different motion patterns for the same action and distinct visual styles. Especially for the former, the motion pattern, \eg the amplitude of the same action, performed by agents in a different environment can be unpredictable.


\subsubsection{Generalization} 

We want to improve the model's generalization ability from a different perspective. In other words, instead of enhancing generalization through big data and large models, we aim to explicitly address the differences between environments aforementioned at the methodological level.

\textbf{Motion Regularization.} Firstly, we must combine the motion information into the diffusion model to distinguish the different motion patterns. Optical flow has thus caught our attention. It refers to the pattern of apparent motion of image objects between two consecutive frames caused by objects or camera movement. In optical flow maps, colors represent the direction of motion, and the depth or intensity of the colors indicates the magnitude of the motion, which is a general feature across different environments.

However, in practice, in the absence of the next observation, we cannot obtain the current optical flow, \(f_{t,t+1}\). Inspired by other motion estimation works~\cite{7780878,10.5555/1771530.1771554}, we assume motion consistency holds over short intervals, meaning abrupt changes do not occur. Consequently, the consecutive optical flow maps are highly correlated, allowing us to predict the current optical flow map using the previous map. The previous map is calculated from the previous two frames and reflects the actual motion pattern in the current environment. 

We notice that optical flow generation does not require complex texture generation, and it is expected not to cause a significant delay in the pipeline. Therefore, we adopt a less powerful but lightweight generative model, VQ-GAN~\cite{esser2021taming}, and train it on our dataset to predict the optical flow map. Empirically, the generalization ability to predict optical flow is much better than predicting actual images. Formally, the training objective is given by:
\begin{equation}  
\small
\min{}\mathcal{L}_{VQ}(E, G, Z) = \| x - \hat{x} \|_2^2 + \| \text{sg}[E(x)] - z_q \|_2^2 + \beta \| \text{sg}[z_q] - E(x) \|_2^2,
\end{equation}
where \(E\) is the encoder, \(G\) is the generator, \(Z\) represents the latent space, \(x\) is the input image, \(\hat{x}\) is the reconstructed image, \(z_q\) is the quantized latent vector, \(\text{sg}\) denotes the stop-gradient operator, and \(\beta\) is a hyperparameter that balances the commitment loss.

\textit{In summary, we use a simple model to predict motion patterns and then a more complex model to reconstruct real textures based on motion patterns.} Therefore, we adopt ControlNet~\cite{zhang2023adding} to incorporate the optical flow map, \(f_{t,t+1}\), into the default diffusion model, \(p_{\theta_{\rm sft}}(x_{t+1}|x_t,a_t, f_{t,t+1})\). Only the ControlNet part needs to be fine-tuned on VH-1.5M at this stage. Formally, the training objective is given by:
\begin{align}
\footnotesize
    \mathcal{L}_{\text{MSE}} &=\left\|\epsilon-\epsilon_{\theta}\left(q\left(x_{t+1}^{(k)}|x_t,a_t,f_{t,t+1}\right),k\right)\right\|^2  \\
     &= \left\|\epsilon-\epsilon_{\theta}\left(\sqrt{\overline{\alpha_t}}x_t+\sqrt{1-\overline{\alpha_t}}\epsilon|a_{t},f_{t,t+1}\right)\right\|^2.
\end{align}

\textbf{Style Transfer.} Secondly, we use LoRA to fine-tune the diffusion model for visual style transfer. Note that LoRA requires very little data, just dozens of samples. Normally, it is convenient to collect data on such a scale in new environments. We expect the model to achieve generalization with as little effort as possible.

\subsection{Planning with Dynamics Model}

To avoid further training in new environments, we prompt the LMM, \ie GPT-4V, as the planner. The LMM needs to be responsible for high-level goal decomposition as well as low-level action selection. Meanwhile, the pre-trained dynamics model can help the LMM better understand the world.

\subsubsection{Goal Decomposition}

For long-term complex tasks, goal decomposition is an indispensable step. Subgoals can be represented in both text and image forms. For the text-based subgoal \(g_{\rm tar}\), we prompt the LMM to generate a reasonable one.
In addition, we train another diffusion model, \(p_{\theta_{\rm sft}}(x_{\rm tar}|x_t,g_{\rm tar})\), to generate the image-based subgoal \(x_{\rm tar}\) only based on the text-based subgoal and current observation. Note that predicting the image of the subgoal can be more challenging than predicting the next observation, which means the results are not very precise. We plan to delve into the impact of different types of subgoals on tasks. See Section \ref{result}.

\subsubsection{One-Step Planner}

Since we can only ensure that the prediction for the next step is relatively accurate, we adopt a one-step planning method. In more detail, we utilize the pre-trained dynamics model to predict the visual outcomes of all the actions in the next state. Once the text/image-based subgoal is obtained, we send the subgoal and all the visual outcomes to the LMM. Then, we prompt it to compare all the potential outcomes with the subgoal and determine which action can bring the agent closer to the goal. 

\newcommand{\subfigwidth}{0.16}
\begin{figure}
  \centering
  \begin{subfigure}{\subfigwidth\textwidth}
    \centering
    \includegraphics[width=\linewidth]{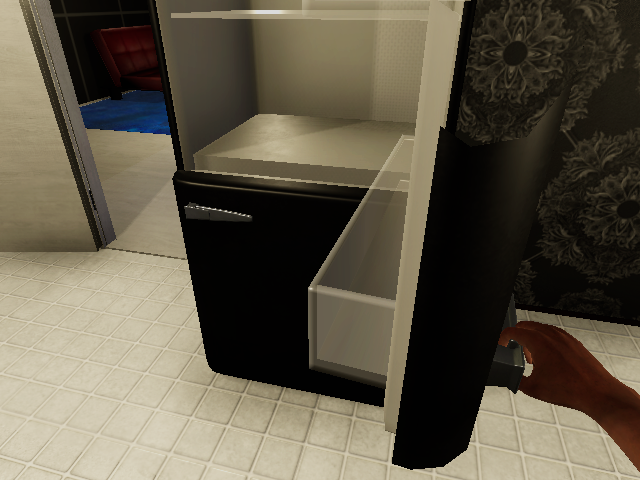}
    \label{subfig:01_Follow_Dutch}
  \end{subfigure}
    \hspace{-0.16cm}
  \begin{subfigure}{\subfigwidth\textwidth}
    \centering
    \includegraphics[width=\linewidth]{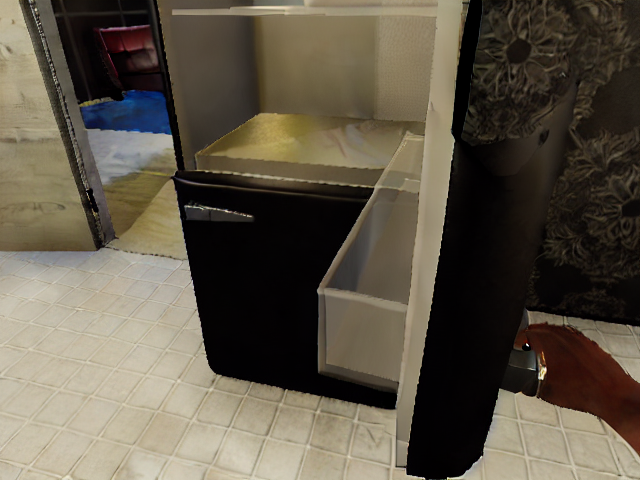} 
    \label{subfig:02_Hitch_horse}
  \end{subfigure}
    \hspace{-0.16cm}
  \begin{subfigure}{\subfigwidth\textwidth}
    \centering
    \includegraphics[width=\linewidth]{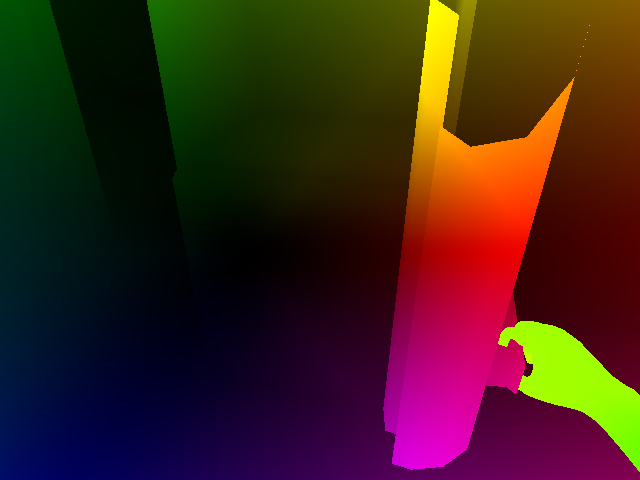}
    \label{subfig:03_Go_to_shed}
  \end{subfigure}
  \hspace{-0.16cm}
  \begin{subfigure}{\subfigwidth\textwidth}
    \centering
    \includegraphics[width=\linewidth]{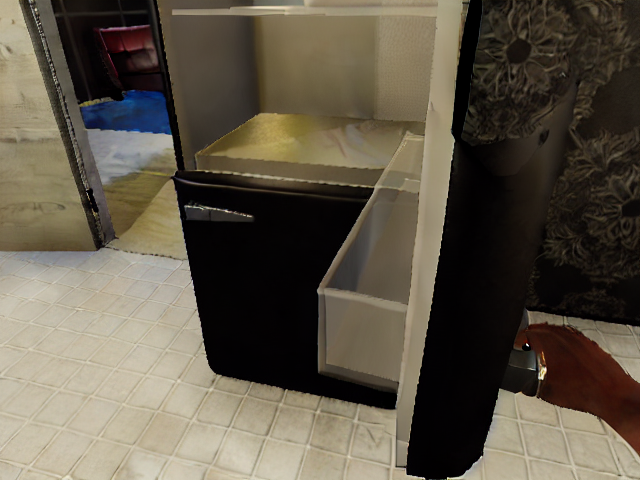}
    \label{subfig:04_Choose_weapon}
  \end{subfigure}
  \hspace{-0.16cm}
  \begin{subfigure}{\subfigwidth\textwidth}
    \centering
    \includegraphics[width=\linewidth]{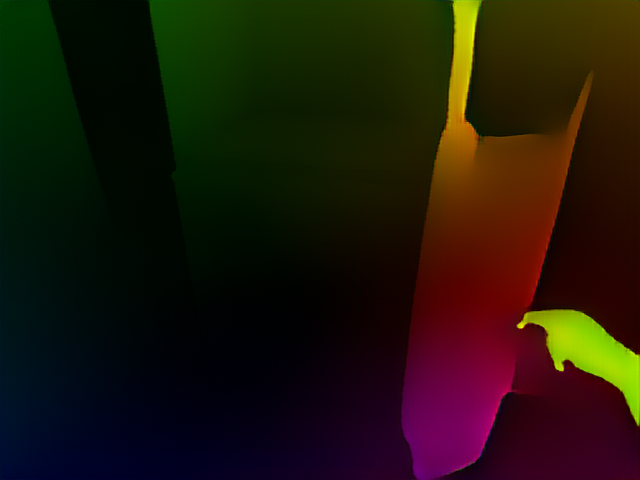}
    \label{subfig:05_Protect_Dutch}
  \end{subfigure}
  \hspace{-0.16cm}
  \begin{subfigure}{\subfigwidth\textwidth}
    \centering
    \includegraphics[width=\linewidth]{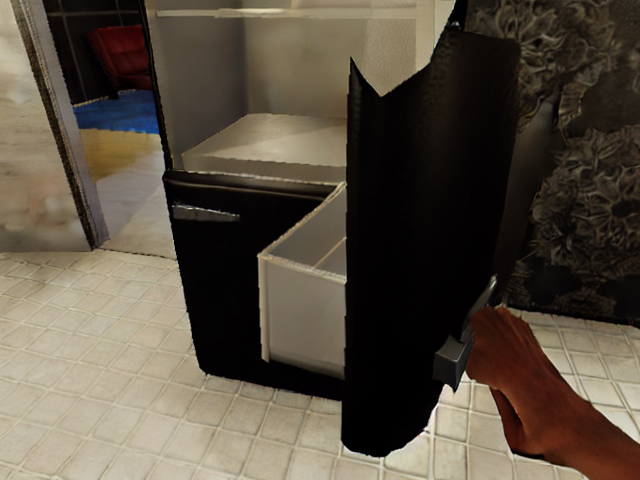}
    \label{subfig:06_Search_house}
  \end{subfigure}
  \vspace{-0.41cm}
  
  \begin{subfigure}{\subfigwidth\textwidth}
    \centering
    \includegraphics[width=\linewidth]{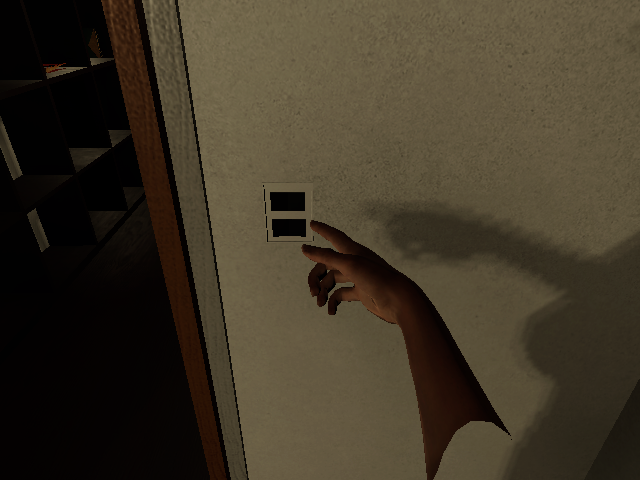}
    \label{subfig:01_Follow_Dutch}
  \end{subfigure}
  \hspace{-0.16cm}
  \begin{subfigure}{\subfigwidth\textwidth}
    \centering
    \includegraphics[width=\linewidth]{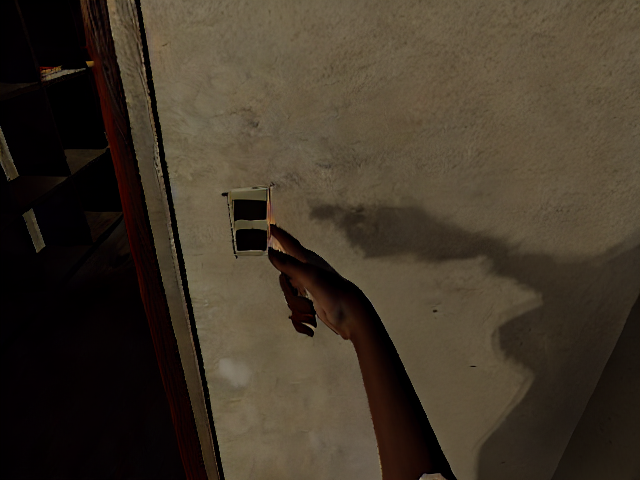} 
    \label{subfig:02_Hitch_horse}
  \end{subfigure}
  \hspace{-0.16cm}
  \begin{subfigure}{\subfigwidth\textwidth}
    \centering
    \includegraphics[width=\linewidth]{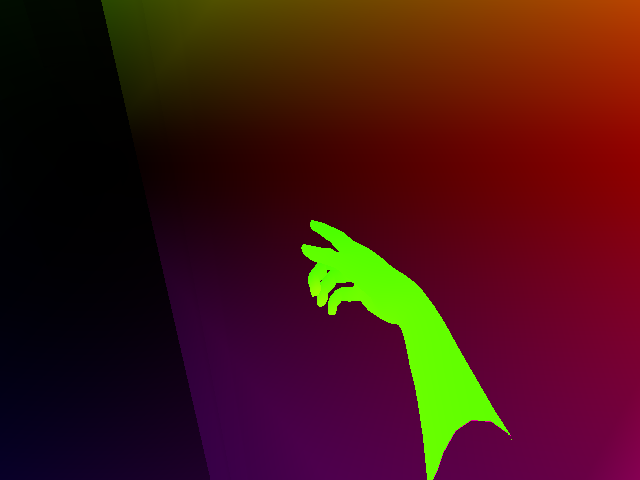}
    \label{subfig:03_Go_to_shed}
  \end{subfigure}
  \hspace{-0.16cm}
  \begin{subfigure}{\subfigwidth\textwidth}
    \centering
    \includegraphics[width=\linewidth]{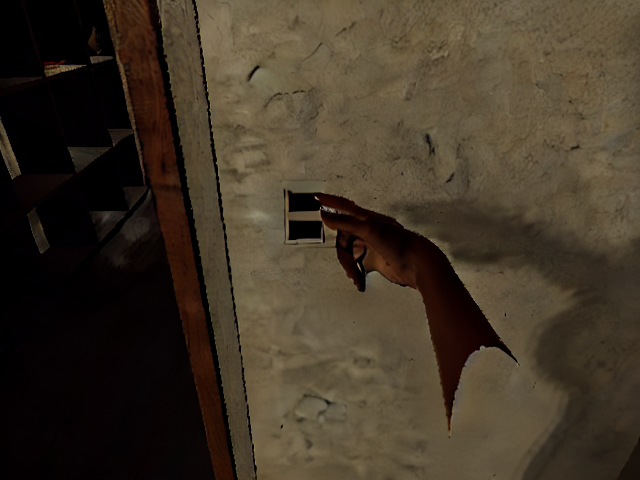}
    \label{subfig:04_Choose_weapon}
  \end{subfigure}
  \hspace{-0.16cm}
  \begin{subfigure}{\subfigwidth\textwidth}
    \centering
    \includegraphics[width=\linewidth]{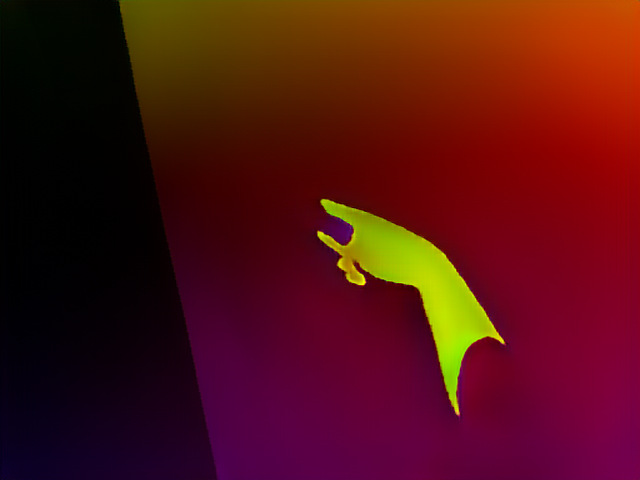}
    \label{subfig:05_Protect_Dutch}
  \end{subfigure}
  \hspace{-0.16cm}
  \begin{subfigure}{\subfigwidth\textwidth}
    \centering
    \includegraphics[width=\linewidth]{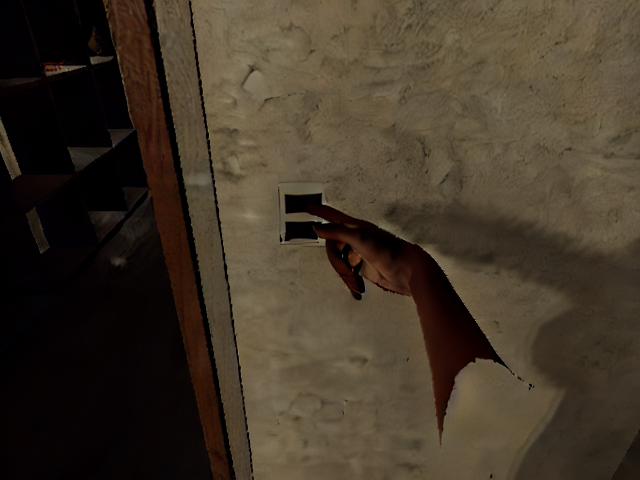}
    \label{subfig:06_Search_house}
  \end{subfigure}
  \vspace{-0.40cm}
  
  \begin{subfigure}{\subfigwidth\textwidth}
    \centering
    \includegraphics[width=\linewidth]{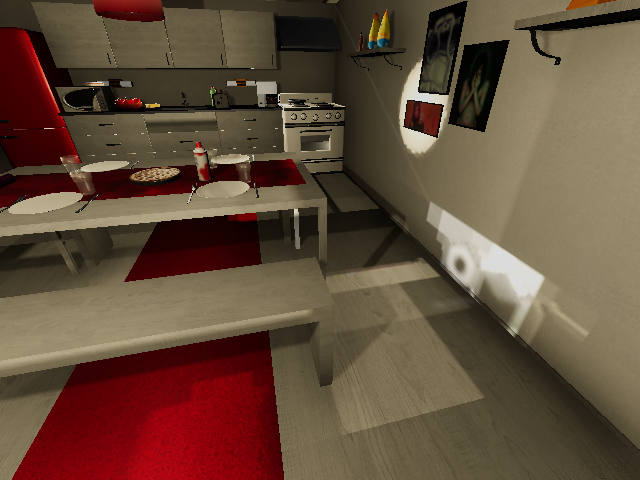}
    \label{subfig:01_Follow_Dutch}
  \end{subfigure}
  \hspace{-0.16cm}
  \begin{subfigure}{\subfigwidth\textwidth}
    \centering
    \includegraphics[width=\linewidth]{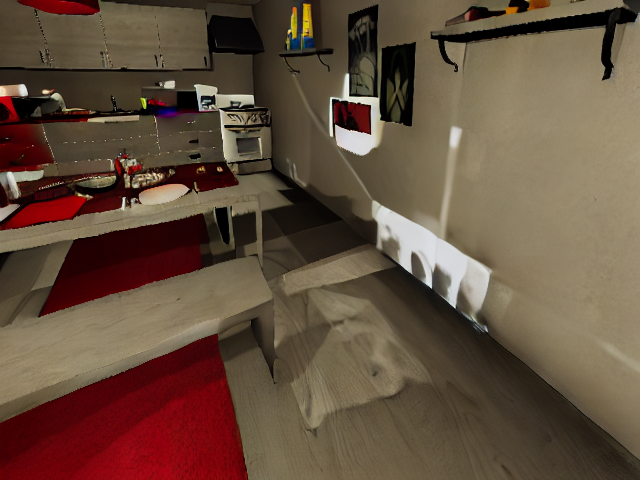} 
    \label{subfig:02_Hitch_horse}
  \end{subfigure}
  \hspace{-0.16cm}
  \begin{subfigure}{\subfigwidth\textwidth}
    \centering
    \includegraphics[width=\linewidth]{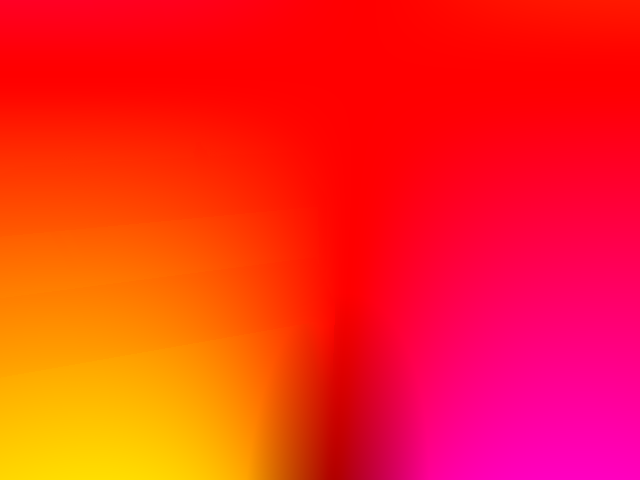}
    \label{subfig:03_Go_to_shed}
  \end{subfigure}
  \hspace{-0.16cm}
  \begin{subfigure}{\subfigwidth\textwidth}
    \centering
    \includegraphics[width=\linewidth]{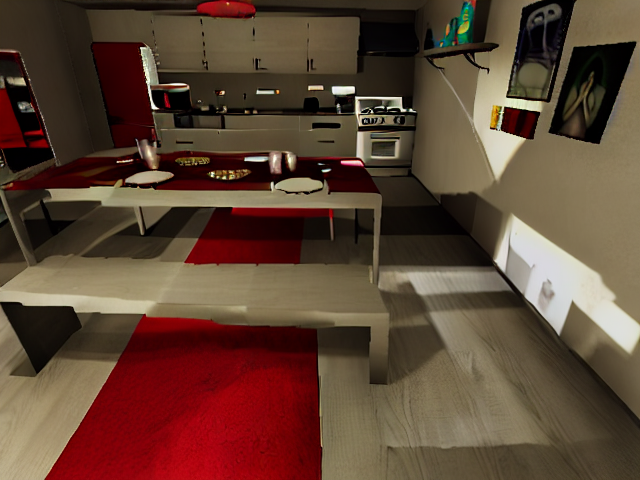}
    \label{subfig:04_Choose_weapon}
  \end{subfigure}
  \hspace{-0.16cm}
  \begin{subfigure}{\subfigwidth\textwidth}
    \centering
    \includegraphics[width=\linewidth]{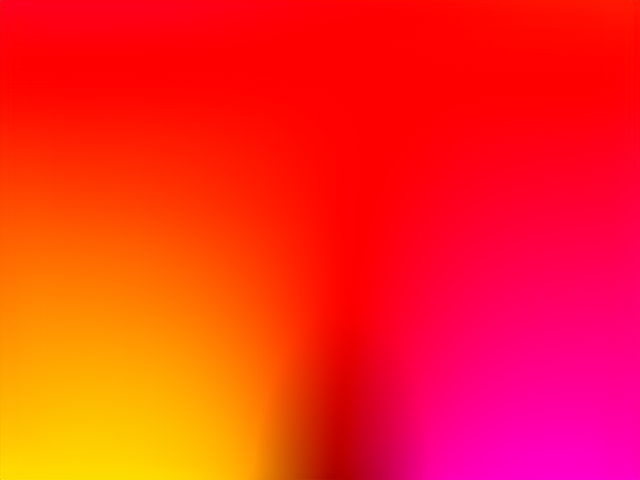}
    \label{subfig:05_Protect_Dutch}
  \end{subfigure}
  \hspace{-0.16cm}
  \begin{subfigure}{\subfigwidth\textwidth}
    \centering
    \includegraphics[width=\linewidth]{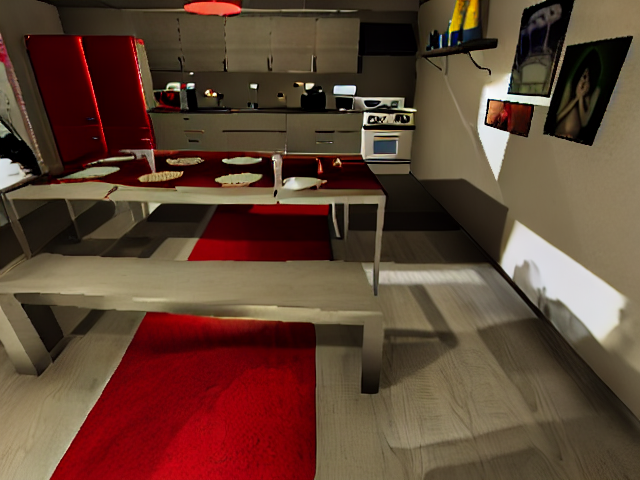}
    \label{subfig:06_Search_house}
  \end{subfigure}
  \vspace{-0.4cm}
  
    \begin{subfigure}{\subfigwidth\textwidth}
    \centering
    \includegraphics[width=\linewidth]{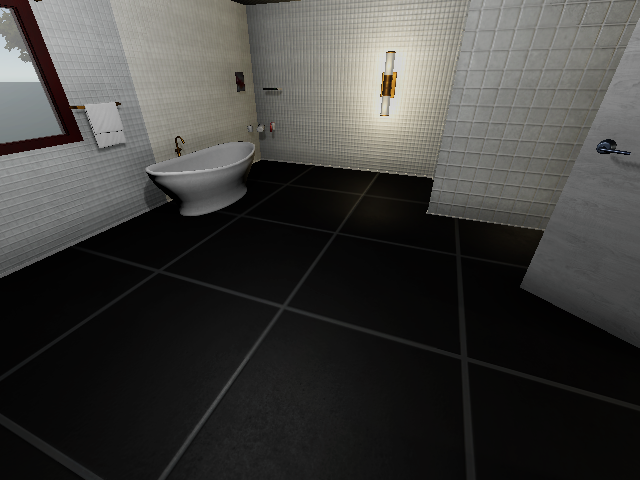}
    \captionsetup{font=tiny}
    \caption{Original} 
    \label{subfig:01_Follow_Dutch}
  \end{subfigure}
   \hspace{-0.16cm}
  \begin{subfigure}{\subfigwidth\textwidth}
    \centering
    \includegraphics[width=\linewidth]{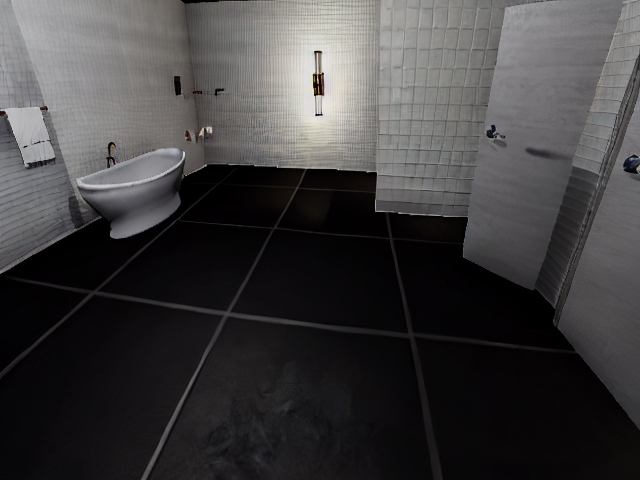} 
    \captionsetup{font=tiny}
    \caption{InstructP2P (finetuned)}
    \label{subfig:02_Hitch_horse}
  \end{subfigure}
   \hspace{-0.16cm}
  \begin{subfigure}{\subfigwidth\textwidth}
    \centering
    \includegraphics[width=\linewidth]{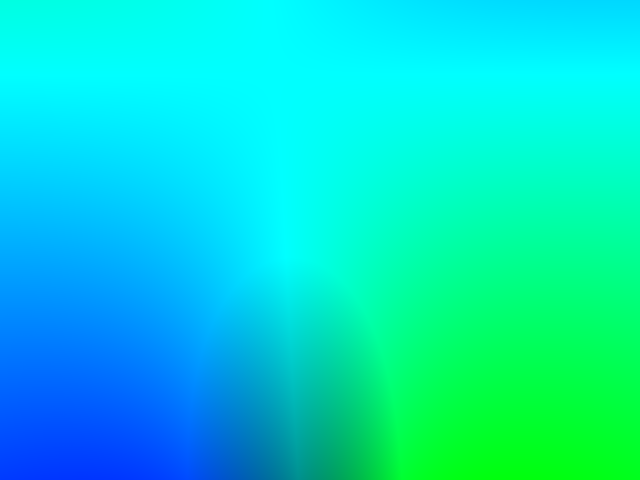}
    \captionsetup{font=tiny}
     \caption{Previous flow} 
    \label{subfig:03_Go_to_shed}
  \end{subfigure}
   \hspace{-0.16cm}
  \begin{subfigure}{\subfigwidth\textwidth}
    \centering
    \includegraphics[width=\linewidth]{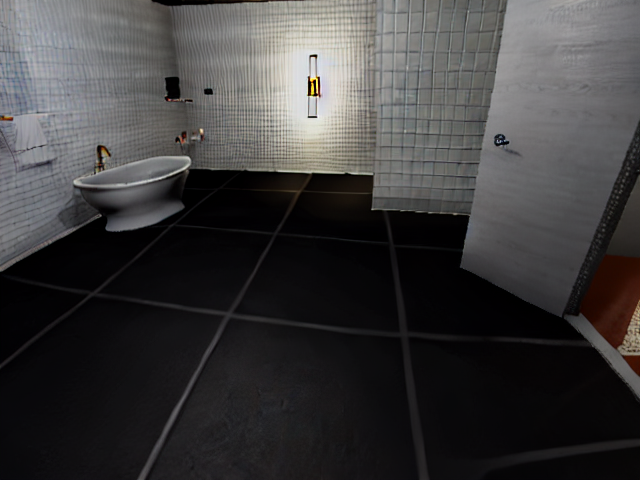}
    \captionsetup{font=tiny}
    \caption{Ours (previous flow)} 
    \label{subfig:04_Choose_weapon}
  \end{subfigure}
   \hspace{-0.16cm}
  \begin{subfigure}{\subfigwidth\textwidth}
    \centering
    \includegraphics[width=\linewidth]{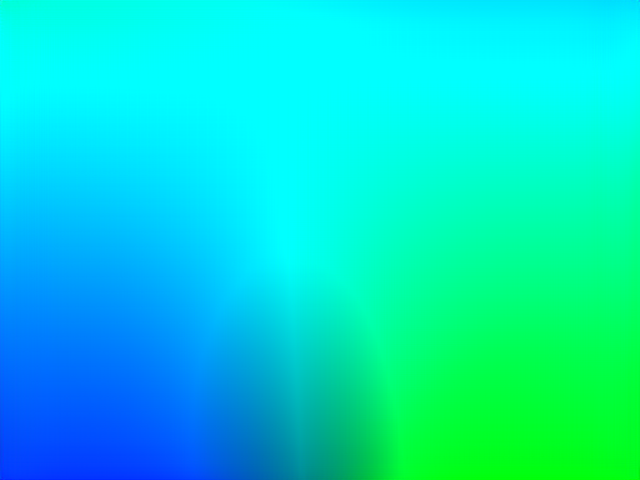}
    \captionsetup{font=tiny}
    \caption{Predicted flow} 
    \label{subfig:05_Protect_Dutch}
  \end{subfigure}
   \hspace{-0.16cm}
  \begin{subfigure}{\subfigwidth\textwidth}
    \centering
    \includegraphics[width=\linewidth]{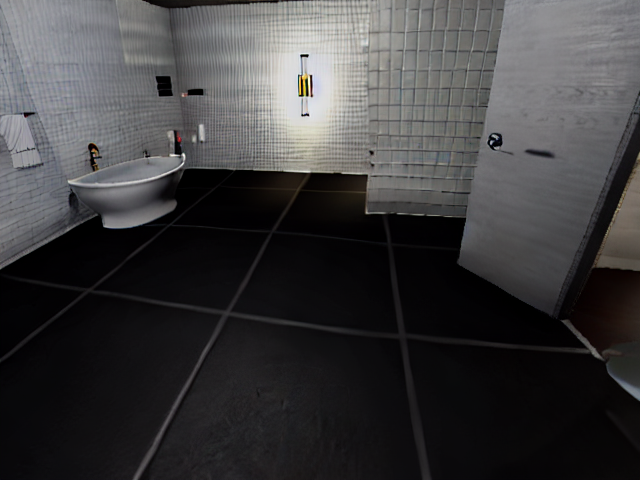}
    \captionsetup{font=tiny}
    \caption{Ours} 
    \label{subfig:06_Search_house}
  \end{subfigure}
  \hspace{-0.02cm}
  \caption{Examples of the generated image of the next observation in VirtualHome. The tasks from rows 1 to 4 are: close the fridge, switch off the light, turn left, and turn right.} 
  \label{fig:examples}
  \vspace{-0.2cm}
\end{figure}

\section{Experiment}
In this section, we comprehensively evaluate and analyze each module of the embodied agent. We first evaluate the quality of image generation using the world model and the quality of optical flow prediction. Secondly, we evaluate whether our world model can assist task planners in completing more complex tasks. Finally, we assess the generalization of our method. 




\subsection{Visual Quality}

We adopt two metrics, FID~\cite{heusel2018gans} and user score, to evaluate the visual quality of the generated image of the world model. For models, \textbf{InstructP2P (pre-trained)} is the default model of InstructP2P. \textbf{InstructP2P (fine-tuned)} is the model fine-tuned on our dataset. \textbf{Ours (previous flow)} is the world model that conditions on the previous optical flow map, while \textbf{Ours} is conditioned on the predicted optical flow map. Note that the validation set of VH-1.5M has around 5k samples.

 \begin{wraptable}{r}{0.50\textwidth}
        \small
        \vspace{-0.4cm}
        \caption{FID score comparison with other models on the validation set. It is calculated between the predicted observation and ground truth. The lower the number, the better the quality of the image.}
        \centering
        \begin{tabular}{ccc}
\toprule
Model & Mean  & Variance  \\ \midrule
InstructP2P (pre-trained) & 13.65 & 0.10 \\
InstructP2P (fine-tuned)  & 1.06 & 0.05  \\
Ours (previous flow)  & 0.83 & 0.03  \\
Ours & \textbf{0.82} & 0.03  \\
\bottomrule
\end{tabular}
\label{tab:1}
\vspace{-0.2cm}
\end{wraptable}

\textbf{FID Score.} FID is a standard metric measuring the distance of two image distributions using the inception model. The smaller the FID is, the more similar the two images are. Table \ref{tab:1} shows the FID score of our model and baselines. We can see that using existing diffusion models as world models is ineffective because their training data often lacks state transition-related data. Meanwhile, introducing an optical flow map, which serves as motion pattern information, significantly enhances the generation results. In addition, world models based on predicted optical flow are slightly better than those based on the optical flow of the previous frame. 

\begin{wraptable}{r}{0.50\textwidth}
        \vspace{-0.4cm}
        \small
        \setlength\tabcolsep{1pt}
        \caption{User score of the user study. The user score is the percentage of images that users consider to meet the criteria out of the total 1000 images. The higher the number, the better the quality of the image. The evaluated images are from the validation set. }
        \centering
        \begin{tabular}{ccc}
\toprule

 Model & Mean  & Variance  \\ \midrule
InstructP2P (fine-tuned) & 54.10$\%$ & 1.53$\%$\\
Ours (previous flow)  & 69.35$\%$& 1.34$\%$\\
Ours  & \textbf{74.93}$\%$& 2.57$\%$\\
\bottomrule
\end{tabular}
\vspace{-0.4cm}
\label{tab:user}
\end{wraptable}


\textbf{User Study.} We also conduct a user study on the accuracy of world models for image generation. For the criterion, users judge the correctness of the direction and amplitude of the executed action. Each user investigates a total of 1000 samples. There are 8 users participating in the survey in total. Our user study, shown in Table \ref{tab:user}, again verifies our predicted optical flow can help generate higher-quality images.

\textbf{Analysis.} As illustrated in Figure \ref{fig:examples}, InstructP2P (fine-tuned) generates the scene of steering in the wrong direction. However, this flaw can be greatly improved by incorporating optical flow information. Moreover, it is observed that the dynamics of closing the refrigerator can be more accurately predicted if the prediction of the motion pattern is considered.


\subsection{VirtualHome Tasks}\label{ex2}

\begin{figure}[t]
  \centering
  \begin{subfigure}{0.4\textwidth}
    \centering
    \includegraphics[width=1.15\linewidth]{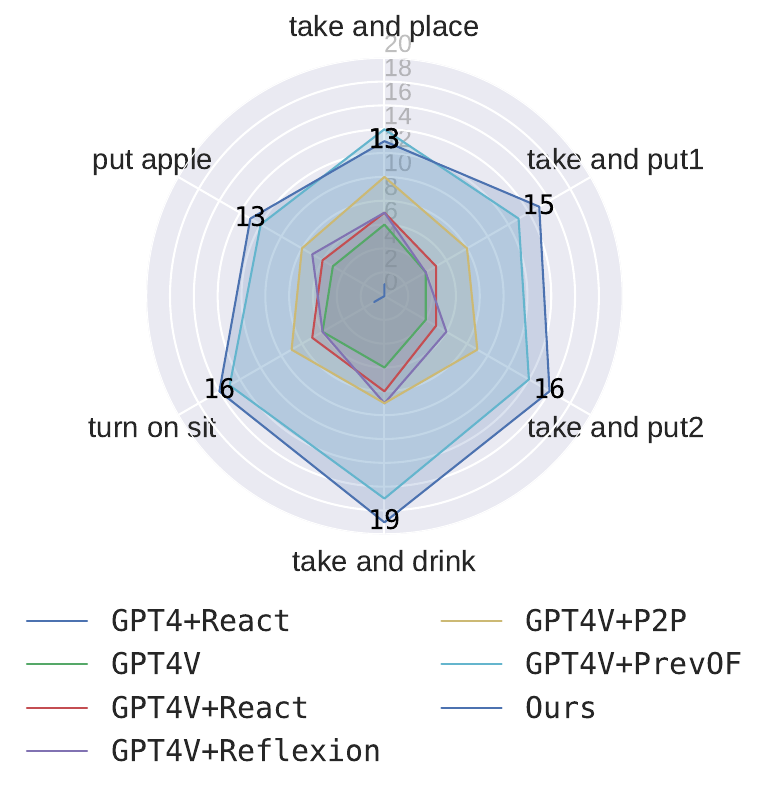}
    \captionsetup{font=scriptsize}
    \caption{Ours (text subgoal) and other baselines on task 1-6} 
    \label{subfig:01_Follow_Dutch}
  \end{subfigure}
  \hspace{1.4cm}
  \begin{subfigure}{0.4\textwidth}
    \centering
    \includegraphics[width=1.15\linewidth]{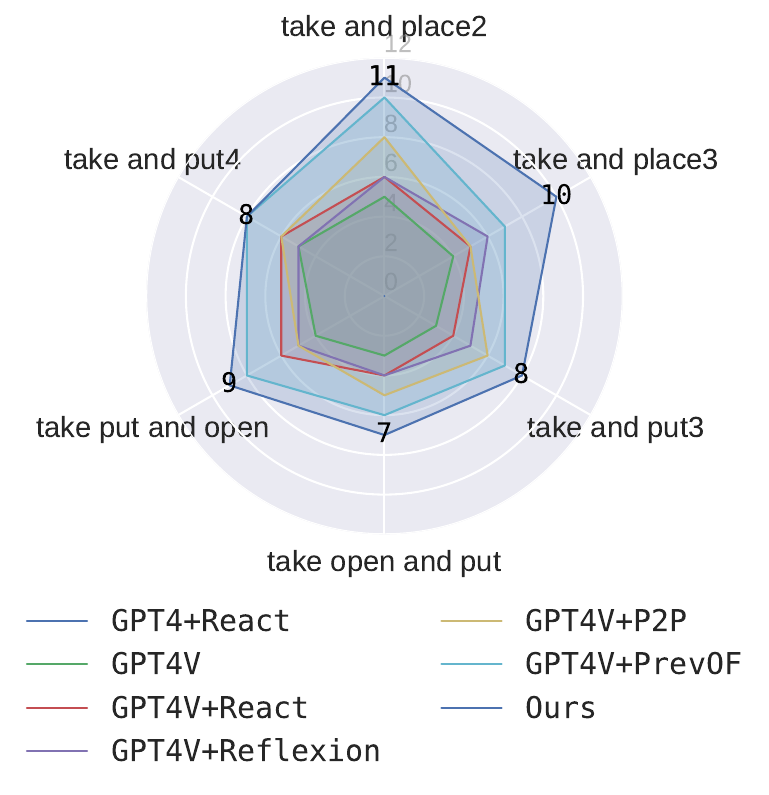} 
    \captionsetup{font=scriptsize}
    \caption{Ours (text subgoal) and other baselines on task 7-12} 
    \label{subfig:02_Hitch_horse}
  \end{subfigure}
  \vspace{2mm}
  \\
  \begin{subfigure}{0.4\textwidth}
    \centering
    \includegraphics[width=.95\linewidth]{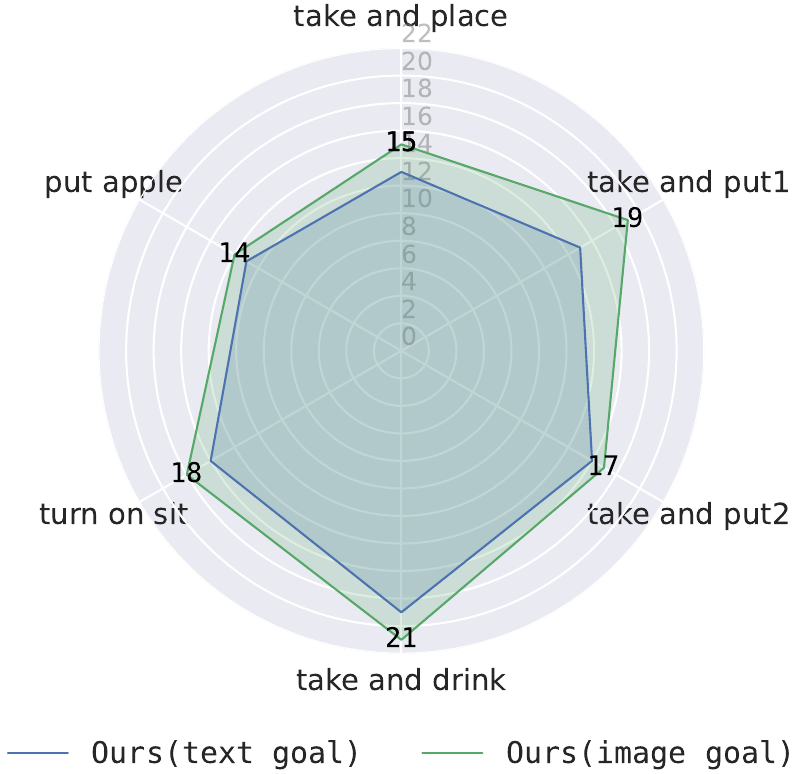}
    \captionsetup{font=scriptsize}
    \caption{Image subgoal and text subgoal on task 1-6} 
    \label{subfig:03_Go_to_shed}
  \end{subfigure}
  \hspace{1.2cm}
  \begin{subfigure}{0.4\textwidth}
    \centering
    \includegraphics[width=\linewidth]{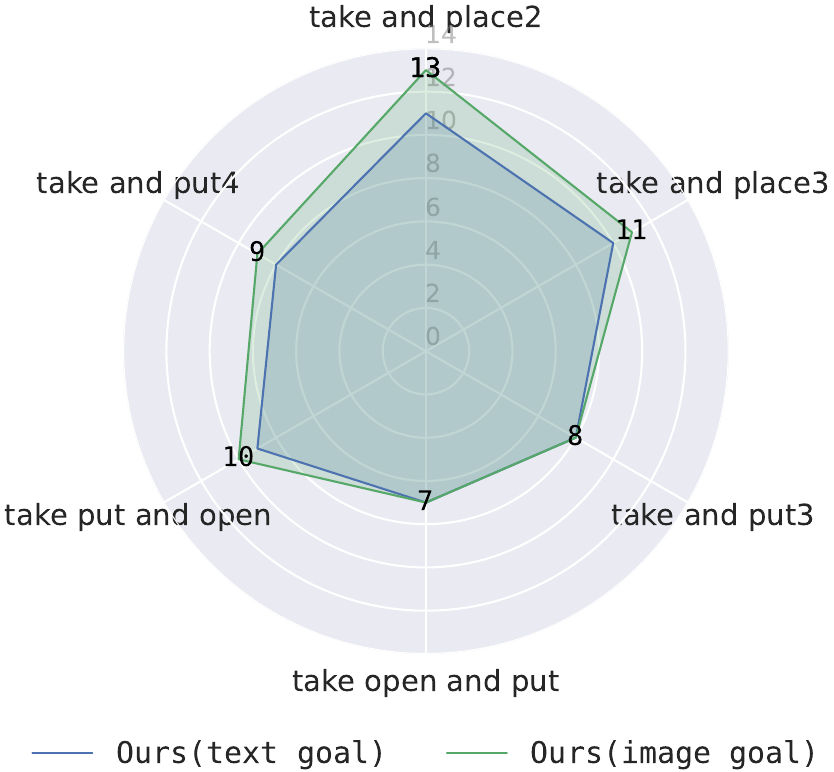}
    \captionsetup{font=scriptsize}
    \caption{Image subgoal and text subgoal on task 7-12} 
    \label{subfig:04_Choose_weapon}
  \end{subfigure}
  \caption{The success rate on 12 tasks for all the methods. Note that tasks 1-6 occur inside one room, while tasks 7-12 take place in two rooms.} 
  \label{fig:sr}
  \vspace{-0.2cm}
\end{figure}

\textbf{Results.}
\label{result}
To prove that our world model can well assist the LMM in task planning, we evaluate various methods on $12$ tasks, each task described by an instruction, in the VirtualHome environment. Each task is tested $100$ times, and the maximum step in one episode is $80$. For each of the $12$ tasks, we abbreviated the task names for convenience. For example, the instruction of task 1, "take the bread from the toaster and place it on the plate on the table," consists of four subtasks: a) walk to the toaster, b) grab the bread, c) walk to the plate, and d) place the bread on the plate. We use "take and place" to refer to task 1.


These 12 instructional tasks are comprised of multiple sequential sub-tasks. For baselines, we use GPT4 combined with React~\cite{yao2023react} as the task planner and policy, denoted as \textbf{GPT4+React}, and it takes input as the JSON format text environment description. We also directly use GPT-4V to make decisions, denoted as \textbf{GPT4V}, and we also combined GPT4V with React~\cite{yao2023react} and Reflexion~\cite{shinn2023reflexionlanguageagentsverbal} as the task planner and policy, denoted as \textbf{GPT4V+React} and \textbf{GPT4V+Reflexion}. For ablation baselines, we use the fine-tuned InstrctP2P as the world model, denoted as \textbf{GPT4V+P2P}. The world model that conditions on the previous optical flow map is denoted as \textbf{GPT4V+PrevOF}. 

As shown in Figure \ref{fig:sr}, the world model significantly improves the GPT-4V ability on various long-horizon tasks. Moreover, the inclusion of optical flow information enhances the accuracy of image generation and further improves task planning performance. The results also demonstrate the effectiveness of the predicted optical flow map.
 
\textbf{Image Subgoal \textit{vs.} Text Subgoal.} In this part, we analyze the impact of different types of subgoals on tasks. During the goal decomposition process, the text subgoal directly outputted by the LLM task planner represents a high-level, coarse-grained description. If our method can generate images of the scene at the completion time of the subgoal, a more detailed, fine-grained description can be obtained. This might enhance the action selection ability that relies on the quality of the subgoal. 

\begin{figure}[t]
  \centering
  \begin{subfigure}{0.135\textwidth}
    \centering
    \includegraphics[width=\linewidth]{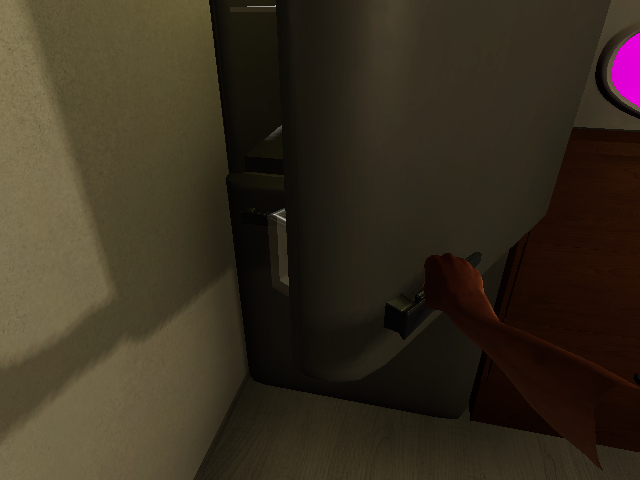}
    \label{subfig:01_Follow_Dutch}
  \end{subfigure}
  \hspace{-0.16cm}
  \begin{subfigure}{0.135\textwidth}
    \centering
    \includegraphics[width=\linewidth]{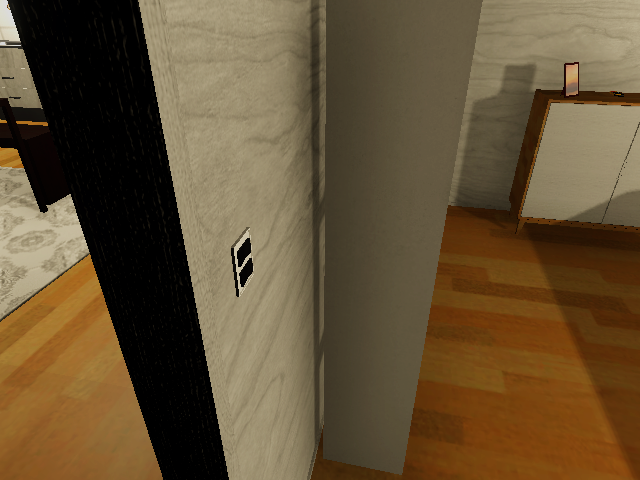} 
    \label{subfig:02_Hitch_horse}
  \end{subfigure}
  \hspace{-0.16cm}
  \begin{subfigure}{0.135\textwidth}
    \centering
    \includegraphics[width=\linewidth]{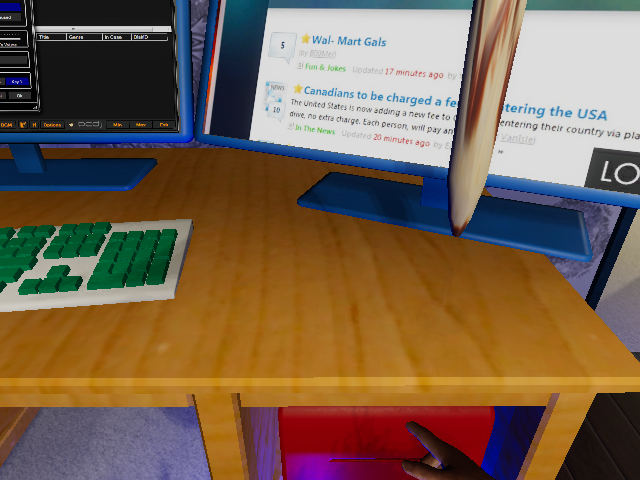}
    \label{subfig:03_Go_to_shed}
  \end{subfigure}
  \hspace{-0.16cm}
  \begin{subfigure}{0.135\textwidth}
    \centering
    \includegraphics[width=\linewidth]{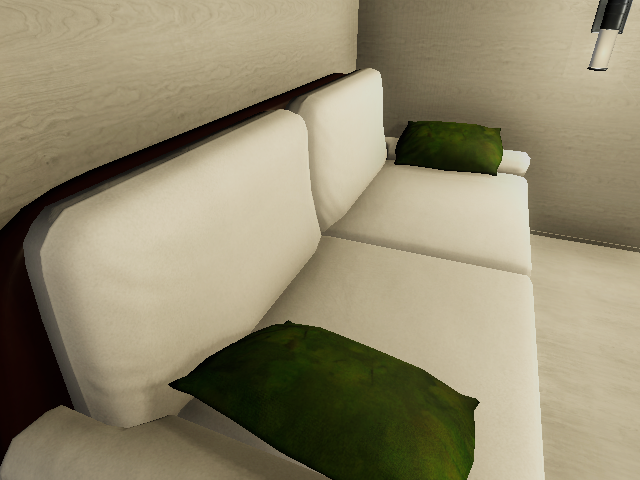}
    \label{subfig:06_Search_house}
  \end{subfigure}
  \hspace{-0.16cm}
    \begin{subfigure}{0.135\textwidth}
    \centering
    \includegraphics[width=\linewidth]{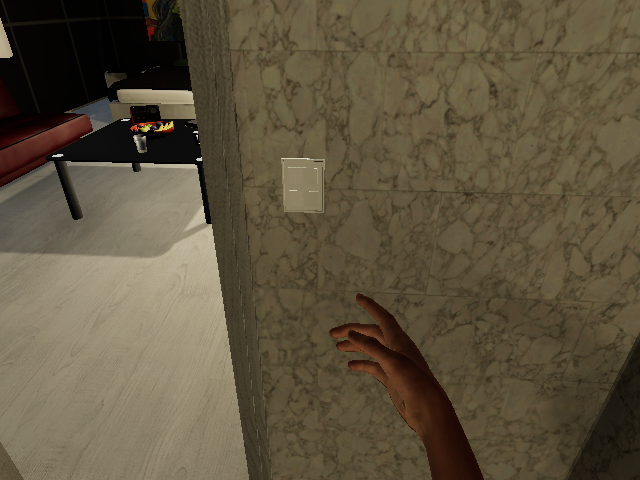}
    \label{subfig:06_Search_house}
  \end{subfigure}
    \hspace{-0.16cm}
   \begin{subfigure}{0.135\textwidth}
    \centering
    \includegraphics[width=\linewidth]{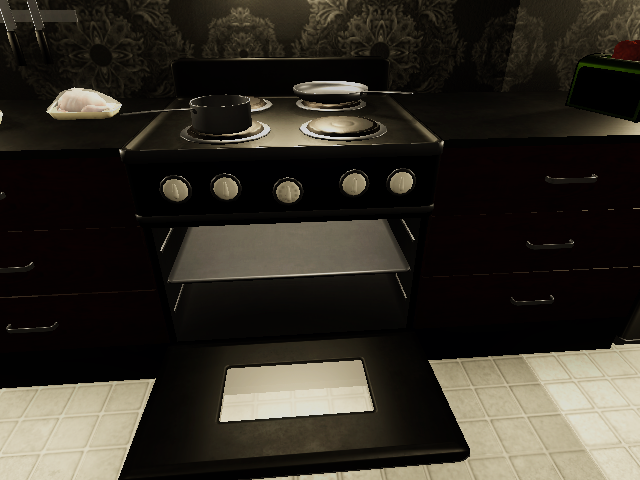}
    \label{subfig:06_Search_house}
  \end{subfigure}
  \hspace{-0.16cm}
  \begin{subfigure}{0.135\textwidth}
    \centering
    \includegraphics[width=\linewidth]{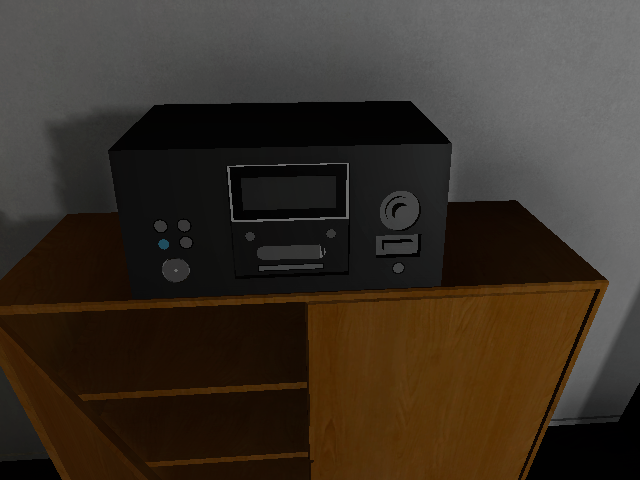}
    \label{subfig:06_Search_house}
  \end{subfigure}
    \vspace{-0.4cm}
  
  \begin{subfigure}{0.135\textwidth}
    \centering
    \includegraphics[width=\linewidth]{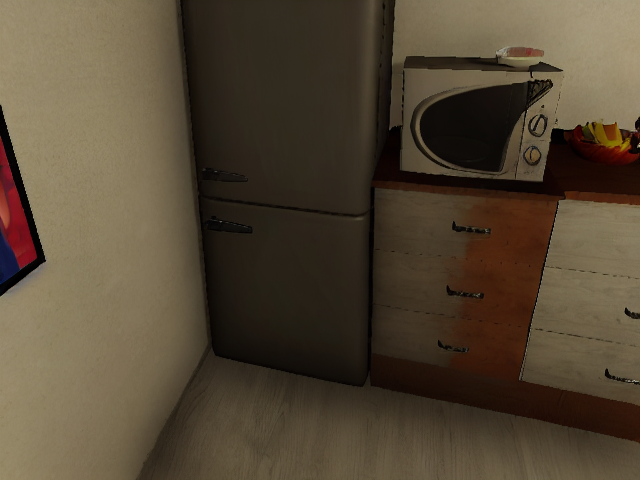}
    \captionsetup{font=tiny}
    \caption{Enclose the fridge} 
    \label{subfig:04_Choose_weapon}
  \end{subfigure}
  \hspace{-0.16cm}
  \begin{subfigure}{0.135\textwidth}
    \centering
    \includegraphics[width=\linewidth]{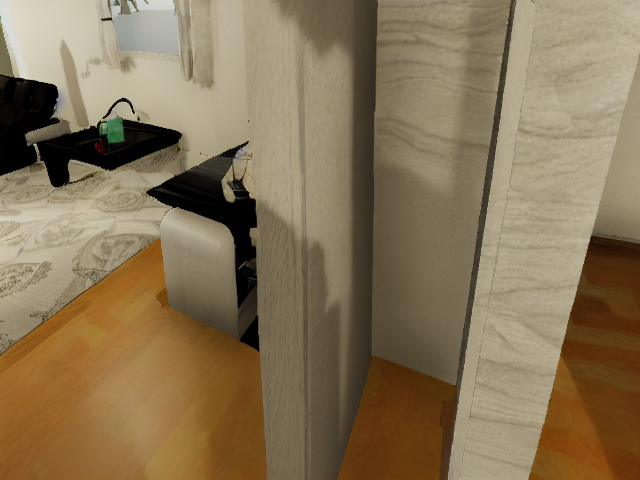}
    \captionsetup{font=tiny}
    \caption{Go through door}
    \label{subfig:05_Protect_Dutch}
  \end{subfigure}
  \hspace{-0.16cm}
  \begin{subfigure}{0.1353\textwidth}
    \centering
    \includegraphics[width=\linewidth]{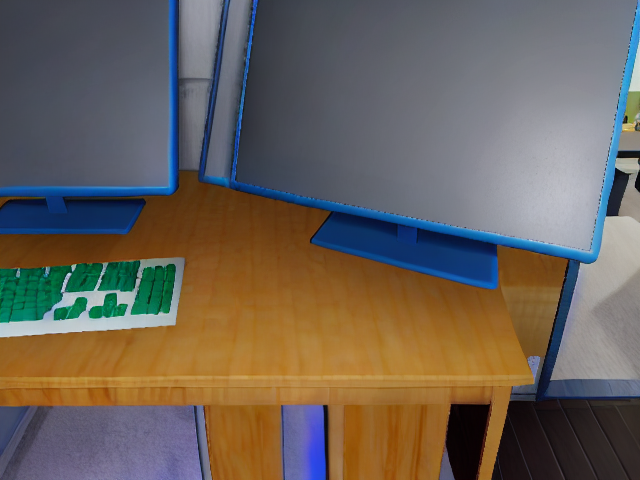}
    \captionsetup{font=tiny}
    \caption{Shut off the PC}
    \label{subfig:06_Search_house}
  \end{subfigure}
  \hspace{-0.16cm}
  \begin{subfigure}{0.135\textwidth}
    \centering
    \includegraphics[width=\linewidth]{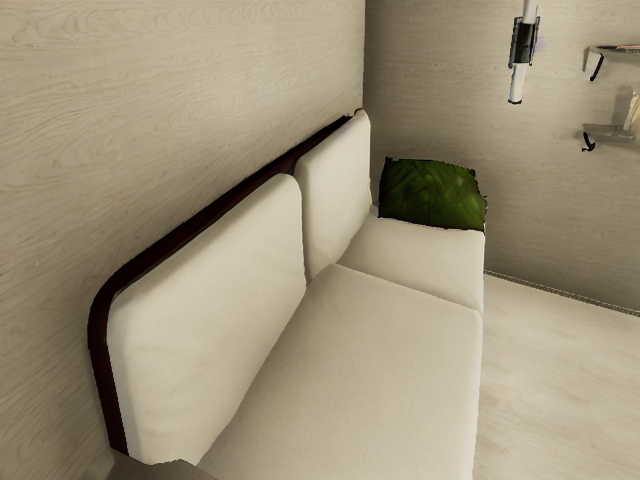}
    \captionsetup{font=tiny}
    \caption{Take hold of pillow}
    \label{subfig:06_Search_house}
  \end{subfigure}
  \hspace{-0.16cm}
  \begin{subfigure}{0.135\textwidth}
    \centering
    \includegraphics[width=\linewidth]{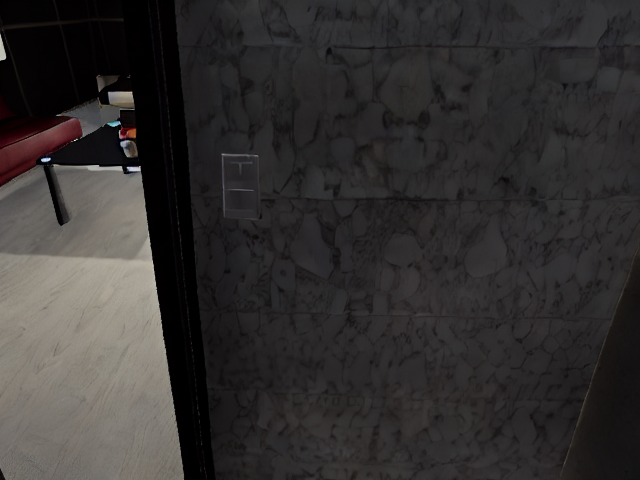}
    \captionsetup{font=tiny}
    \caption{Switch off the light}
    \label{subfig:06_Search_house}
  \end{subfigure}
  \hspace{-0.16cm}
   \begin{subfigure}{0.135\textwidth}
    \centering
    \includegraphics[width=\linewidth]{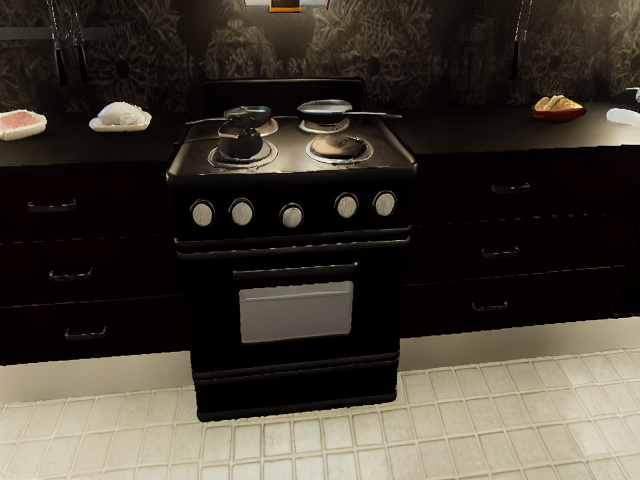}
    \captionsetup{font=tiny}
    \caption{Shut the stove}
    \label{subfig:06_Search_house}
  \end{subfigure}
  \hspace{-0.16cm}
  \begin{subfigure}{0.135\textwidth}
    \centering
    \includegraphics[width=\linewidth]{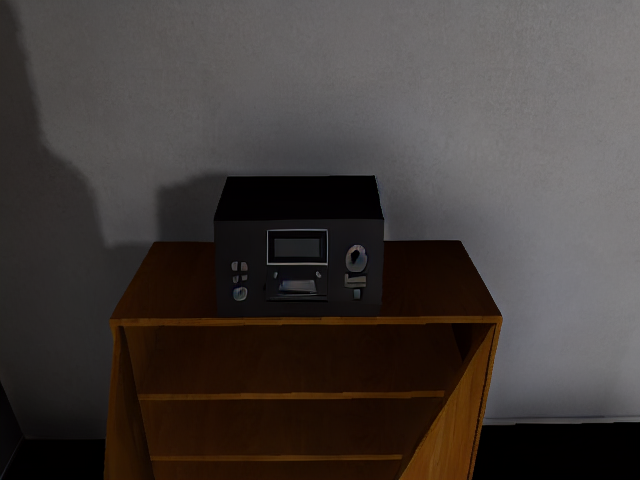}
    \captionsetup{font=tiny}
    \caption{Open the cabinet}
    \label{subfig:06_Search_house}
  \end{subfigure}
  \hspace{-0.01cm}
  \caption{Examples of the generated image subgoals. The first row is the original image, and the second row is the image subgoal generated based on the text subgoal.} 
  \label{fig:image_subtask}
  \vspace{-0.2cm}
\end{figure}

Specifically, we have trained an InstructP2P model based on VH-1.5M to generate the image when the subgoal is completed, with the generation results illustrated in Figure \ref{fig:image_subtask}. The decision-making results in Figure \ref{fig:sr} show that fine-grained subgoal description is better than coarse-grained description, even if the generated image is not so accurate. 

\begin{figure}[t]
    \centering
    \begin{minipage}[c]{0.1\textwidth}
        \captionsetup{justification=raggedright, singlelinecheck=false,font=scriptsize}
        \caption*{Previous flow}
        \vspace{1cm}
        \caption*{VQ-GAN prediction}
        \vspace{0.8cm}
        \caption*{Ground truth}
    \end{minipage}
    \begin{minipage}[c]{0.88\textwidth}
    \begin{subfigure}{0.195\textwidth}
        \centering
        \includegraphics[width=\textwidth]{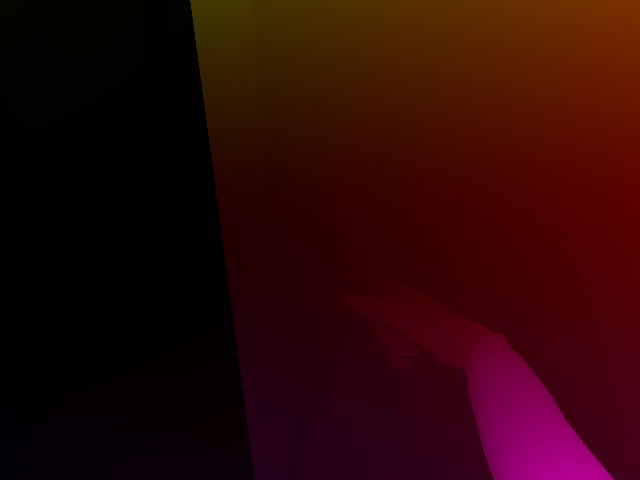}
        \label{fig:small2}
    \end{subfigure} 
    \hspace{-0.16cm}
    \begin{subfigure}{0.195\textwidth}
        \centering
        \includegraphics[width=\textwidth]{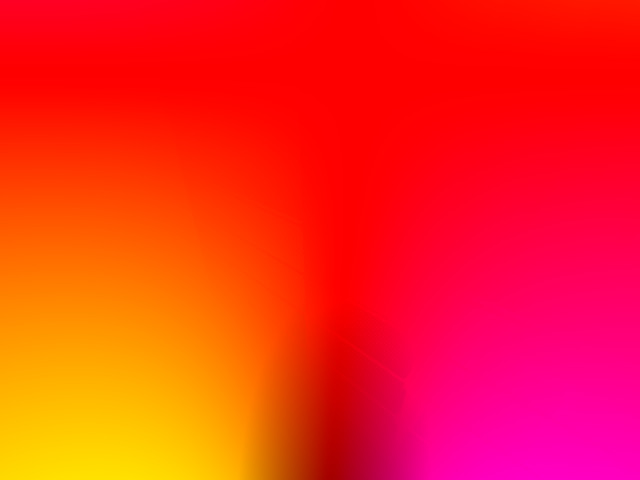}
        \label{fig:small3}
    \end{subfigure} 
    \hspace{-0.16cm}
    \begin{subfigure}{0.195\textwidth}
        \centering
        \includegraphics[width=\textwidth]{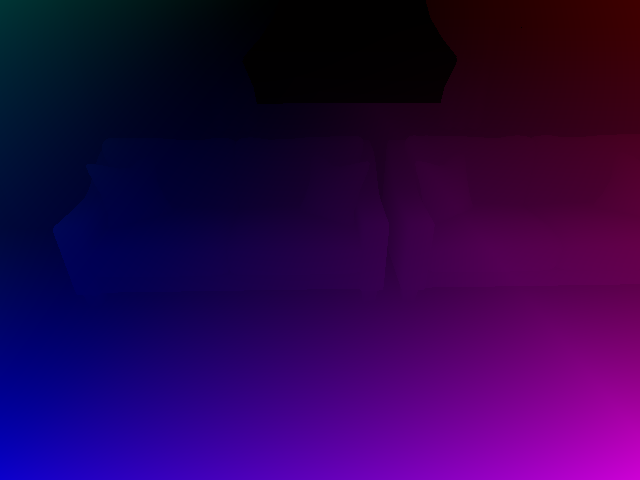}
        \label{fig:small4}
    \end{subfigure} 
    \hspace{-0.16cm}
    \begin{subfigure}{0.195\textwidth}
        \centering
        \includegraphics[width=\textwidth]{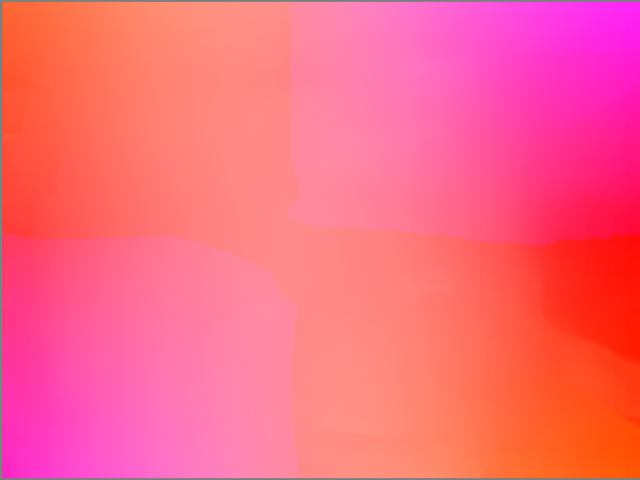}
        \label{fig:small3}
    \end{subfigure} 
    \hspace{-0.16cm}
    \begin{subfigure}{0.195\textwidth}
        \centering
        \includegraphics[width=\textwidth]{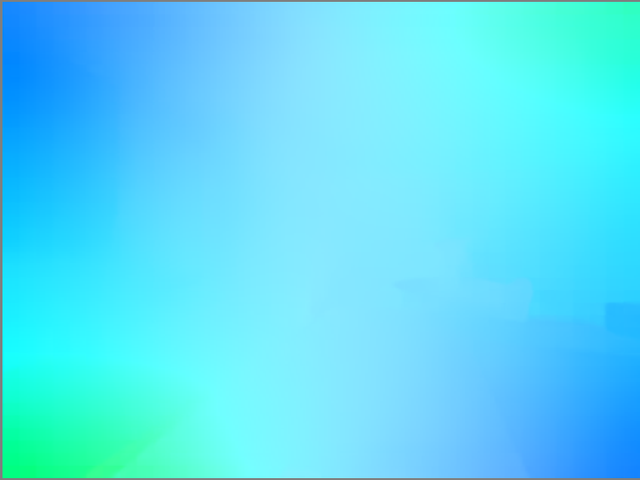}
        \label{fig:small4}
    \end{subfigure} 
    \vspace{-0.4cm}
    
    \begin{subfigure}{0.195\textwidth}
        \centering
        \includegraphics[width=\textwidth]{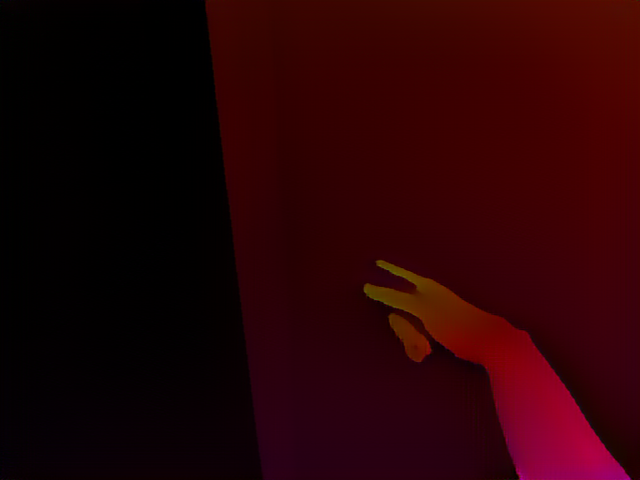}
        \label{fig:small6}
    \end{subfigure} 
    \hspace{-0.16cm}
    \begin{subfigure}{0.195\textwidth}
        \centering
        \includegraphics[width=\textwidth]{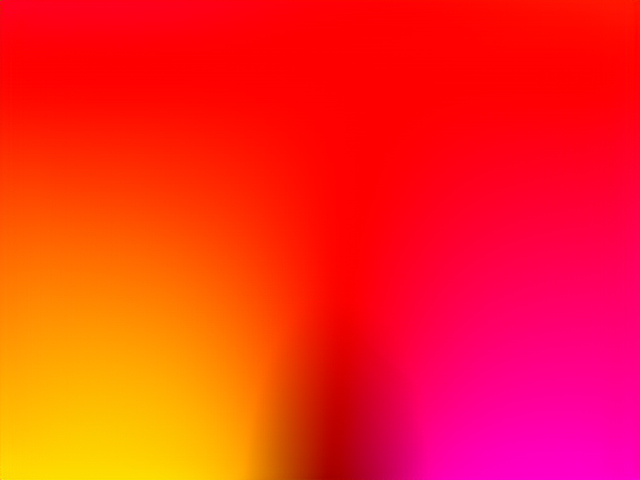}
        \label{fig:small7}
    \end{subfigure} 
    \hspace{-0.16cm}
    \begin{subfigure}{0.195\textwidth}
        \centering
        \includegraphics[width=\textwidth]{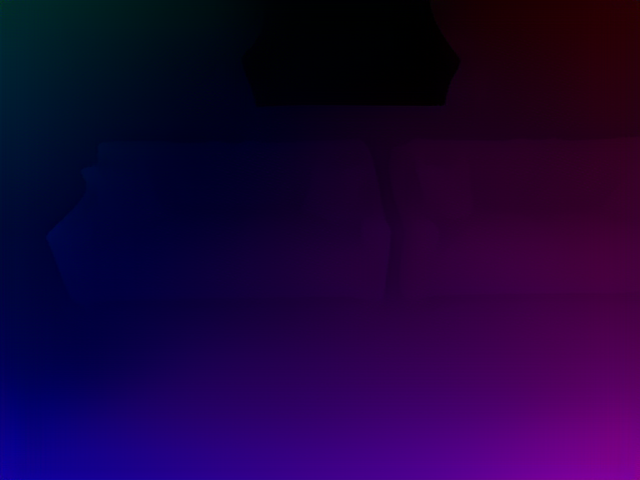}
        \label{fig:small4}
    \end{subfigure}
    \hspace{-0.16cm}
    \begin{subfigure}{0.195\textwidth}
        \centering
        \includegraphics[width=\textwidth]{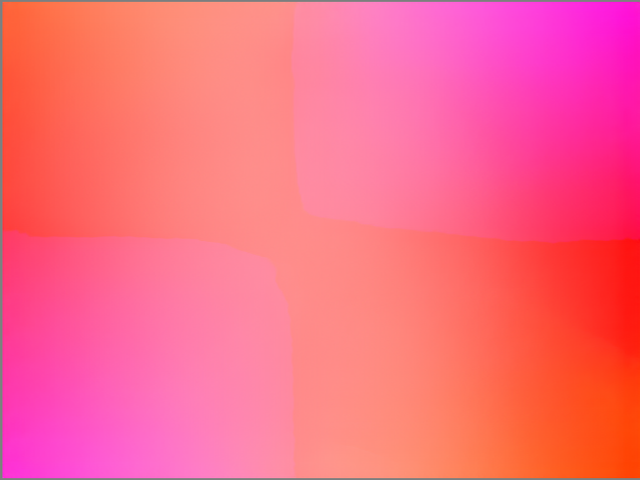}
        \label{fig:small7}
    \end{subfigure} 
    \hspace{-0.16cm}
    \begin{subfigure}{0.195\textwidth}
        \centering
        \includegraphics[width=\textwidth]{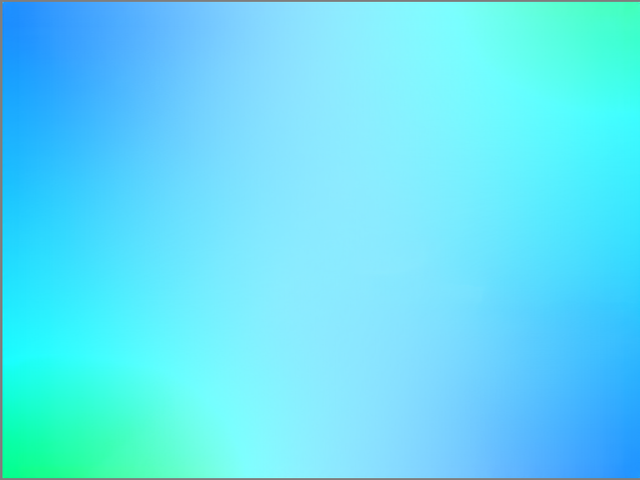}
        \label{fig:small8}
    \end{subfigure}
    \vspace{-0.4cm}
    
    \begin{subfigure}{0.195\textwidth}
        \centering
        \includegraphics[width=\textwidth]{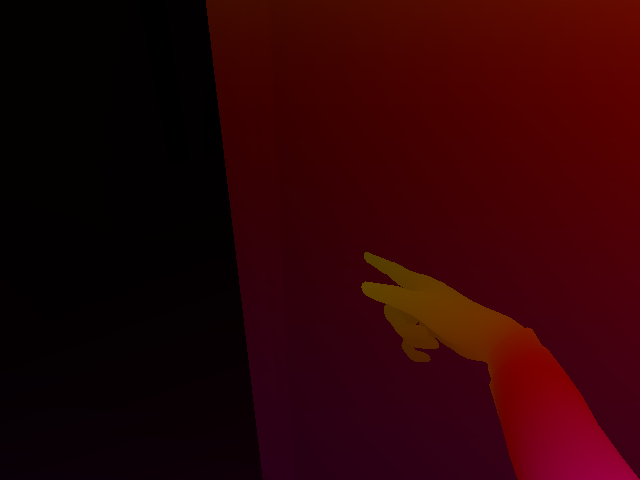}
        \captionsetup{font=scriptsize}
        \caption{Switch off lights}
        \label{fig:small01}
    \end{subfigure} 
    \hspace{-0.16cm}
    \begin{subfigure}{0.195\textwidth}
        \centering
        \includegraphics[width=\textwidth]{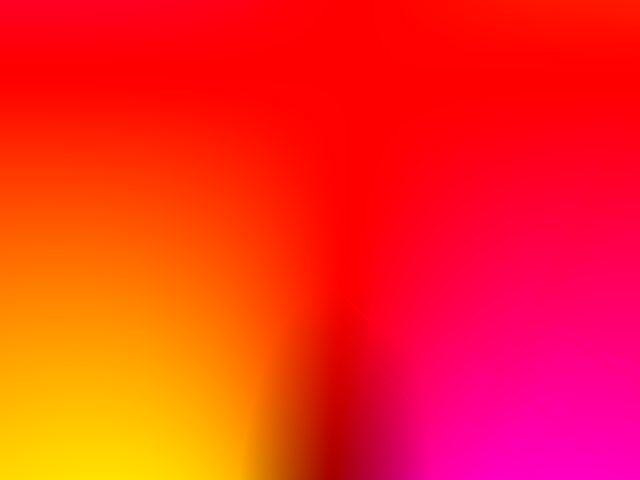}
        \captionsetup{font=scriptsize}
        \caption{Turn to the left}
        \label{fig:small02}
    \end{subfigure} 
    \hspace{-0.16cm}
    \begin{subfigure}{0.195\textwidth}
        \centering
        \includegraphics[width=\textwidth]{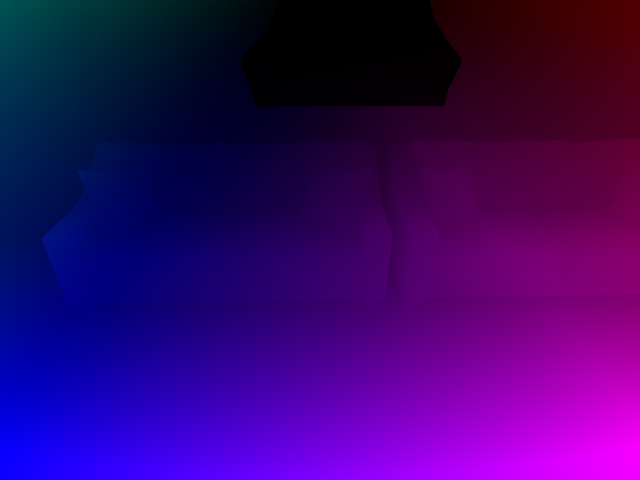}
        \captionsetup{font=scriptsize}
        \caption{Walk straight ahead}
        \label{fig:small03}
    \end{subfigure}
    \hspace{-0.16cm}
    \begin{subfigure}{0.195\textwidth}
        \centering
        \includegraphics[width=\textwidth]{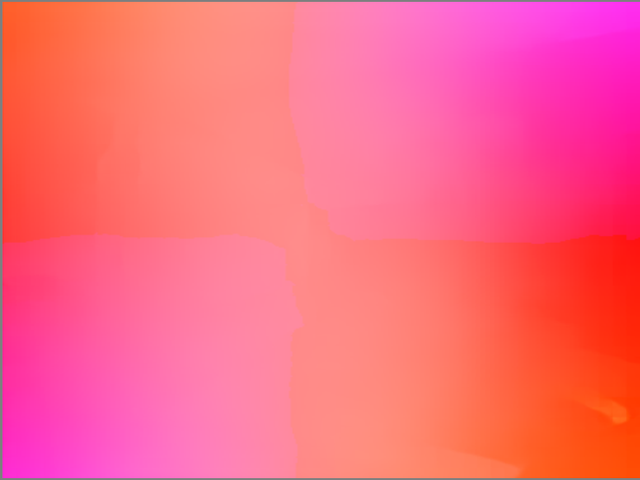}
        \captionsetup{font=scriptsize}
        \caption{Turn to the left}
        \label{fig:small10}
    \end{subfigure} 
    \hspace{-0.16cm}
    \begin{subfigure}{0.195\textwidth}
        \centering
        \includegraphics[width=\textwidth]{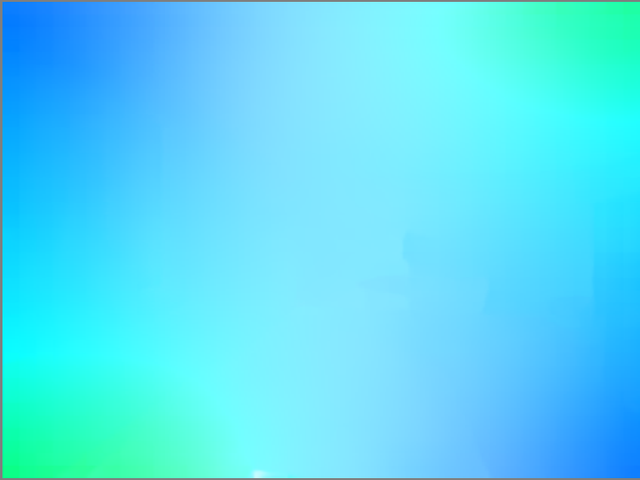}
        \captionsetup{font=scriptsize}
        \caption{Turn to the right }
        \label{fig:small11}
    \end{subfigure} 
    \end{minipage}
    \caption{Examples of optical flow prediction by VQ-GAN. The first 3 columns are optical flow from the VirthualHome environment. The last $2$ columns are optical flow from the Habitat2.0 environment.}\label{fig4gl}
    \vspace{-0.2cm}
\end{figure}

We also conduct a user study to evaluate the visual quality of the generated image-based subgoals. More details can be found in the appendix. 


\subsection{Motion Pattern}
\label{sec:mp}
As mentioned before, we cannot obtain the optical flow from the current timestep to the next timestep. Therefore, we adopt the VQ-GAN model to predict the current optical flow map. As illustrated in Figure \ref{fig:small01} and \ref{fig:small03}, the quality of prediction for details is promising. Furthermore, as demonstrated in Figure \ref{fig:small10} and \ref{fig:small11}, the VQ-GAN trained on the VH-1.5M dataset can easily generalize to other environments. This is because the optical flow map is a universal feature and does not require the prediction of complex textures.


\begin{wraptable}{r}{0.50\textwidth}
\small
\setlength\tabcolsep{3pt}
\vspace{-0.4cm}
\centering
\caption{Average endpoint error (AEE) results. The lower the number, the closer the image is to the ground truth.}
\label{epeaee}

\begin{tabular}{ccc}
\toprule
   & Previous flow  &  Prediction flow \\ \midrule
   Habitat2.0 & 3.30 & \textbf{3.09}      \\
AI2-THOR & 5.00 & \textbf{4.08}      \\
VirtualHome  & 21.22 & \textbf{15.71}  \\
\bottomrule
\end{tabular}
\vspace{-.2cm}
\end{wraptable}


The average endpoint error (AEE) specifically measures the average distance between two motion vectors at the pixel level. As illustrated in Table \ref{epeaee}, the gap between the predicted optical flow map and ground truth is narrower than that between the previous flow map and ground truth (current optical flow map). In addition, the model trained on VirtualHome can still predict optical flow maps in Habitat2.0 and AI2-THOR~\cite{ai2thor}. This confirms the effectiveness and generalization of the VQ-GAN model.


\begin{figure*}[t]
    \begin{minipage}[c]{.5\linewidth}
        \centering
        \includegraphics[width=1\linewidth]{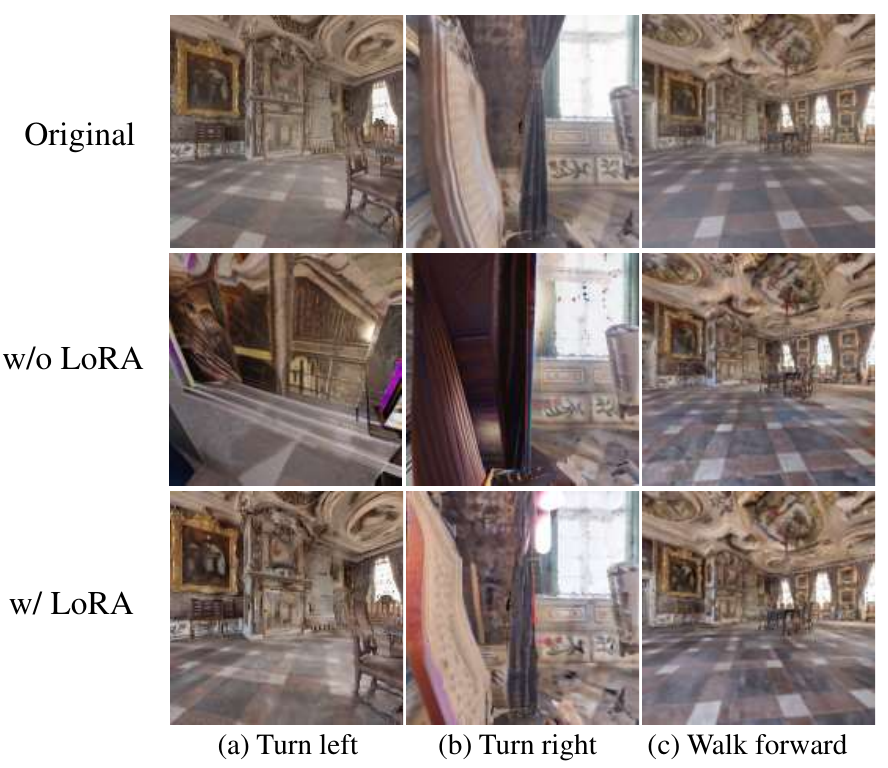}
        \caption{Examples of the generated images of the next observation in Habitat2.0. }
    \label{fig:lora_results}
    \end{minipage}
    \hspace{0.4cm}
    \begin{minipage}[c]{.45\linewidth}
        \centering
        \includegraphics[width=.9\linewidth]{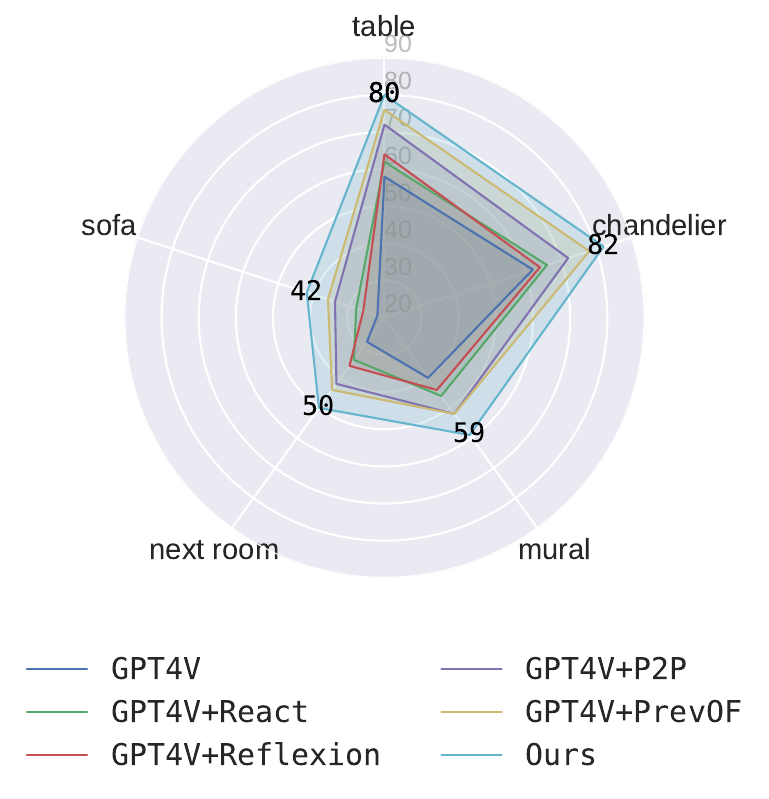}
        \caption{The success rate on 5 navigation tasks for all the methods in Habitat2.0. GPT4+React is omitted due to its poor performance.}
        \label{fig:gen_results}
    \end{minipage}
\vspace{-2mm}
\end{figure*}

\subsection{Generalization}

To assess the generalization of our method, we also evaluate its performance in a new household environment. In more detail, we choose Habitat2.0 due to its high-fidelity scenes compared with other simulators, such as AI2-THOR. However, Habitat2.0 does not provide any inter-frame regarding manipulation skills, which is unrealistic. Therefore, we only carry out navigation tasks.



To enhance usability, we use the pre-trained optical flow model, RAFT~\cite{teed2020raft}, to calculate the optical flow for the previous step since the optical flow cannot be directly obtained. The RAFT results are shown in the last 2 columns of Figure \ref{fig4gl}. Since VQ-GAN has demonstrated some degree of generalization ability to Habitat2.0 in Section \ref{sec:mp}, we can predict the motion pattern of the new environment. The remaining task is to transfer the visual style to a new environment, and we adopt LoRA to fine-tune the world model. As shown in Figure \ref{fig:lora_results}, we successfully perform style transfer with a small amount of data (tens of samples), and the results with LoRA are closer to real scene images compared to those without LoRA visually.  


Figure \ref{fig:gen_results} shows the success rate of all methods on navigation tasks in Habitat2.0. We can draw the same conclusion as in the VirtualHome environment: incorporating predicted optical flow into the world model enhances the agent's decision-making capabilities. Additionally, our method achieved a high success rate, which further demonstrates its strong generalization ability.


\section{Conclusion and Limitations}

This paper introduces EgoPlan, an embodied agent, using the LMM as the one-step planner and the text2image model as the world model for long-horizon tasks. We demonstrate its high-quality image generation, precise optical flow prediction, and promising decision-making ability. More importantly, we have confirmed its generalization capabilities across different environments. It is also important to acknowledge the limitations of EgoPlan. Currently, the agent uses encapsulated skills as actions. It cannot perform low-level control, \eg joint position. How to directly control low-level actions is left as future work. 


\bibliographystyle{plain}
\bibliography{reference}

{
\small

\clearpage
\appendix

\section{Appendix}
\subsection{Details of Virtualhome tasks}
We conducted experiments to evaluate the decision-making ability of all methods in the VirtualHome environment. In total, we investigated 12 complex tasks, with detailed instructions for each task as follows:
\begin{lstlisting}[breaklines=true,caption={Instructions and subtasks.}]
<$one-house instructions$>

1. take and place: take the bread from the toaster and place it on the plate on the table
steps: (a). walk to the toaster
	 (b). grab the bread
	 (c). walk to the table
	 (d). place the bread on the plate
2. take and put1: take the apple from the table and put it in the microwave
steps: (a). walk to the table
	 (b). grab the apple
	 (c). walk to the microwave
	 (d). open the microwave (if the microwave is closed)
	 (e). put the apple in the microwave
3. take and put2: take the book from the table and put it on the bookshelf
steps: (a). walk to the table
	 (b). take the book
	 (c). grab the book
	 (d). walk to the bookshelf
	 (e). put the book on the bookshelf
4. take and drink: take the water glass from the table and drink from it
steps: (a). walk to the table
	 (b). take the water glass
	 (c). drink the water glass
5. turn on sit: turn on the TV and sit down
steps: (a). walk to the TV
	 (b). turn on the TV
	 (c). walk to the chair
	 (d). sit down
6. put apple: Put an apple that is on the table into the bookshelf
steps: (a). walk to the table
	 (b). grab the apple
	 (c). walk to the bookshelf
	 (d). put the apple on the bookshelf

<$two-houses instructions$>

7. take and place2: take the frying pan from the counter and place it in the sink
steps: (a). walk to the counter
	 (b). grab the frying pan
	 (c). walk through the door
	 (d). walk to the sink
	 (e). place frying pan in the sink
8. take and place3: take the condiment shaker from the bookshelf and place it on the table
steps: (a). walk to the bookshelf
	 (b). grab the condiment shaker
	 (c). walk through the door
	 (d). walk to the table
	 (e). place condiment shaker on the table
9. take and put3: take the salmon on top of the microwave and put it in the fridge
steps: (a). walk to the microwave
	 (b). grab the salmon
	 (c). walk through the door
	 (d). walk to the fridge
	 (e). open the fridge (if the fridge is closed)
	 (f). put salmon in the fridge
10. take open and put: take the pie on the table and warm it using the stove
steps: (a). walk to the table
	 (b). grab the pie
	 (c). walk through the door
	 (d). walk to the stove
	 (e). put pie on the stove
	 (f). switch on the stove
11. take put and open: put the sponge in the sink and wet it by switching on the faucet
steps: (a). walk to the sponge
	 (b). grab the sponge
	 (c). walk through the door
	 (d). walk to the sink
	 (e). put sponge in the sink
	 (f). switching on the faucet
12. take and put4: take the condiment bottle from the kitchen table and put it on the plate
steps: (a). walk through the door
	 (b). walk to the kitchen table
	 (c). grab the condiment bottle
	 (d). walk to the plate
	 (e). put pie on the stove
	 (f). switch on the stove
\end{lstlisting}
\label{prompt:appendix_instruction.txt}
\subsection{Details of VH-1.5M's text actions}
The dataset includes a wide range of action sequences, each meticulously annotated with corresponding text actions. These text actions are crucial for providing contextual information that aligns visual actions with natural language descriptions. Below, we detail the process and structure used to generate the text actions for each action sequence in the dataset.

The generation of text actions for VH-1.5M involves a systematic and automated process. This process ensures consistency and variety in the text actions, which are essential for robust training and evaluation in vision-and-language tasks. The key steps in this process are as follows:

\textbf{Verb Selection:} A list of verbs related to various actions (e.g., "walk through," "close," "drink") is predefined. For each identified action sequence directory, a verb is randomly selected from the relevant list. This selection ensures a diverse representation of actions.

\textbf{Object Name Extraction:} Each directory represents the object acted upon, which signifies the object affected by the action. However, if the action does not involve an object, such as "walk through" or "turn left," no extraction is necessary.

\textbf{Phrase Construction:} Two types of phrases are constructed for each action sequence:

Next Timestep Phrase: Describes the immediate next action in the sequence. For example, "next timestep: redeposit the plate".

Goal State Phrase: Describes the intended final action or goal of the sequence. For example, "the goal state: redeposit plate".

\textbf{Prompt File Creation:} The constructed phrases are saved in a prompt json file within the respective action sequence directory. This JSON file contains two keys: "next" and "goal," corresponding to the next timestep phrase and goal state phrase, respectively.

\subsubsection{More examples of the samples}
We give some samples in the sequence of the task, which are shown in Figure~\ref{1111}. Note that samples in one sequence are arranged in chronological order, with the timestep increasing from top to bottom.

\begin{figure}[!t]
  \centering
 
   \vspace{-0.7cm}
  \includegraphics[width=0.9\textwidth]{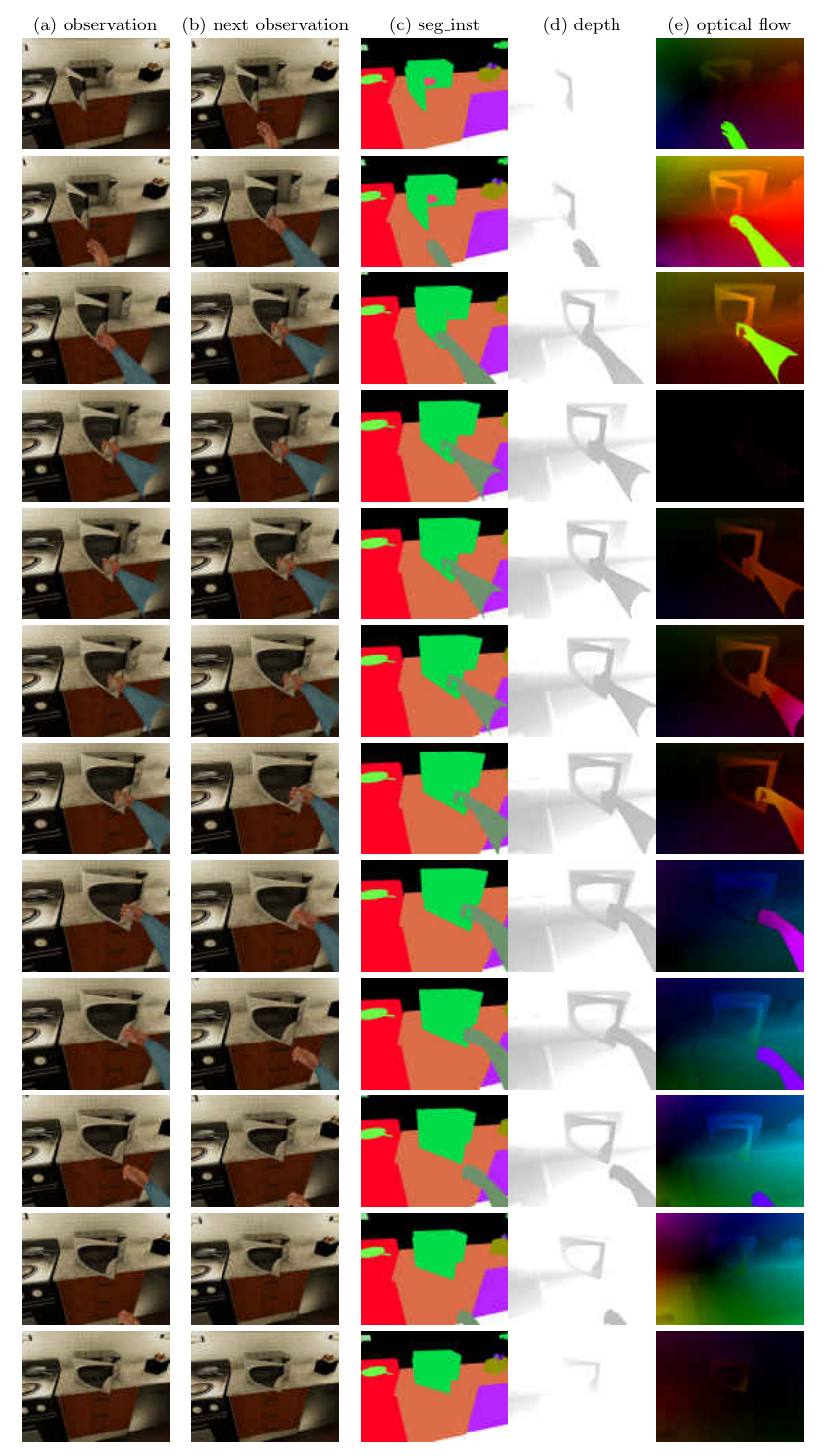}
  \caption{Samples in the sequence of closing the microwave.}
  \label{1111}
\vspace{-0.4cm}
\end{figure}

\begin{figure}[!t]
  \centering
 
   \vspace{-0.7cm}
  \includegraphics[width=0.9\textwidth]{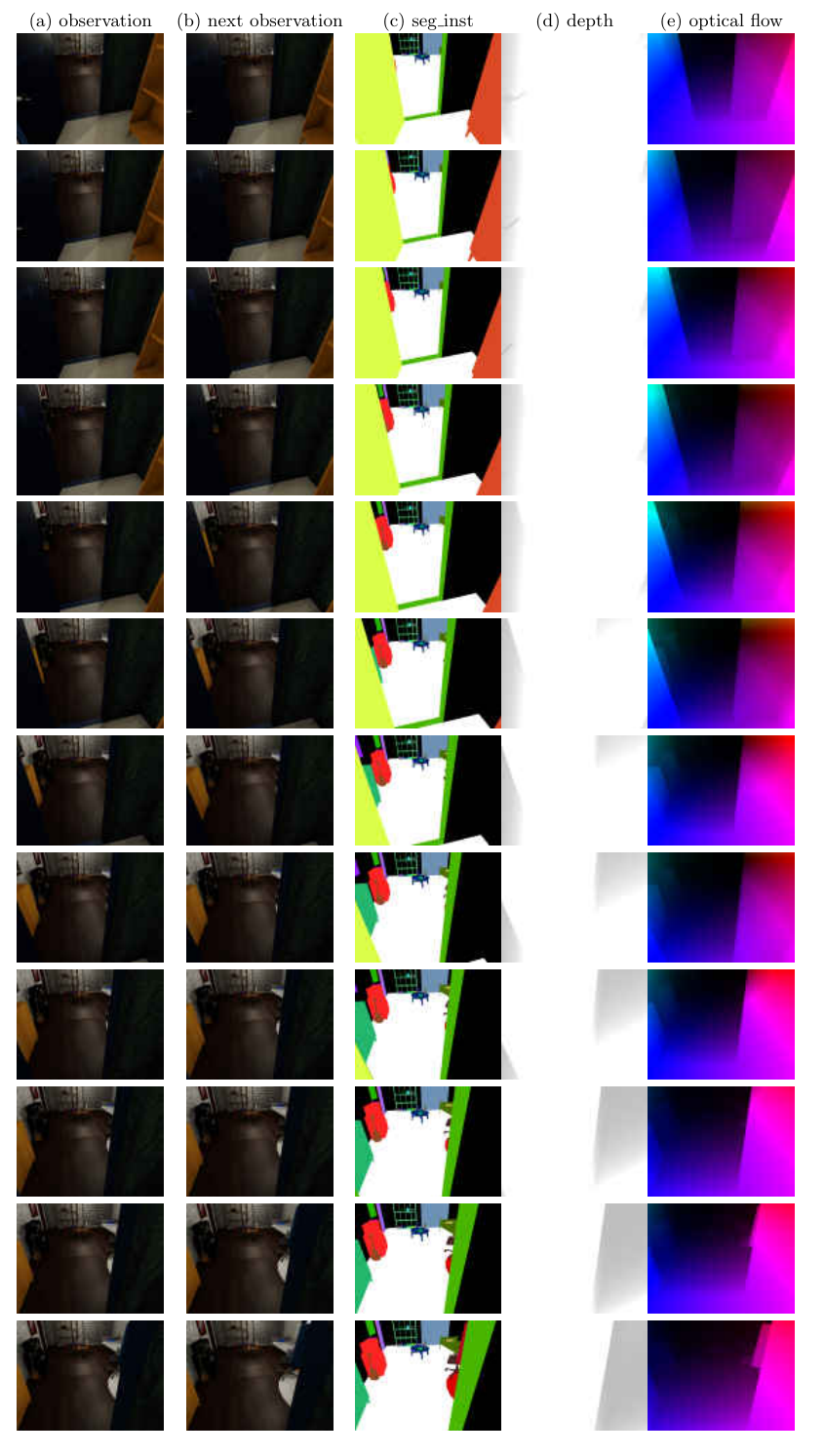}
  \caption{Samples in the sequence of walking through the door.}
 \label{2222}
\vspace{-0.4cm}
\end{figure}

\begin{figure}[!t]
  \centering
 
  \includegraphics[width=0.95\textwidth]{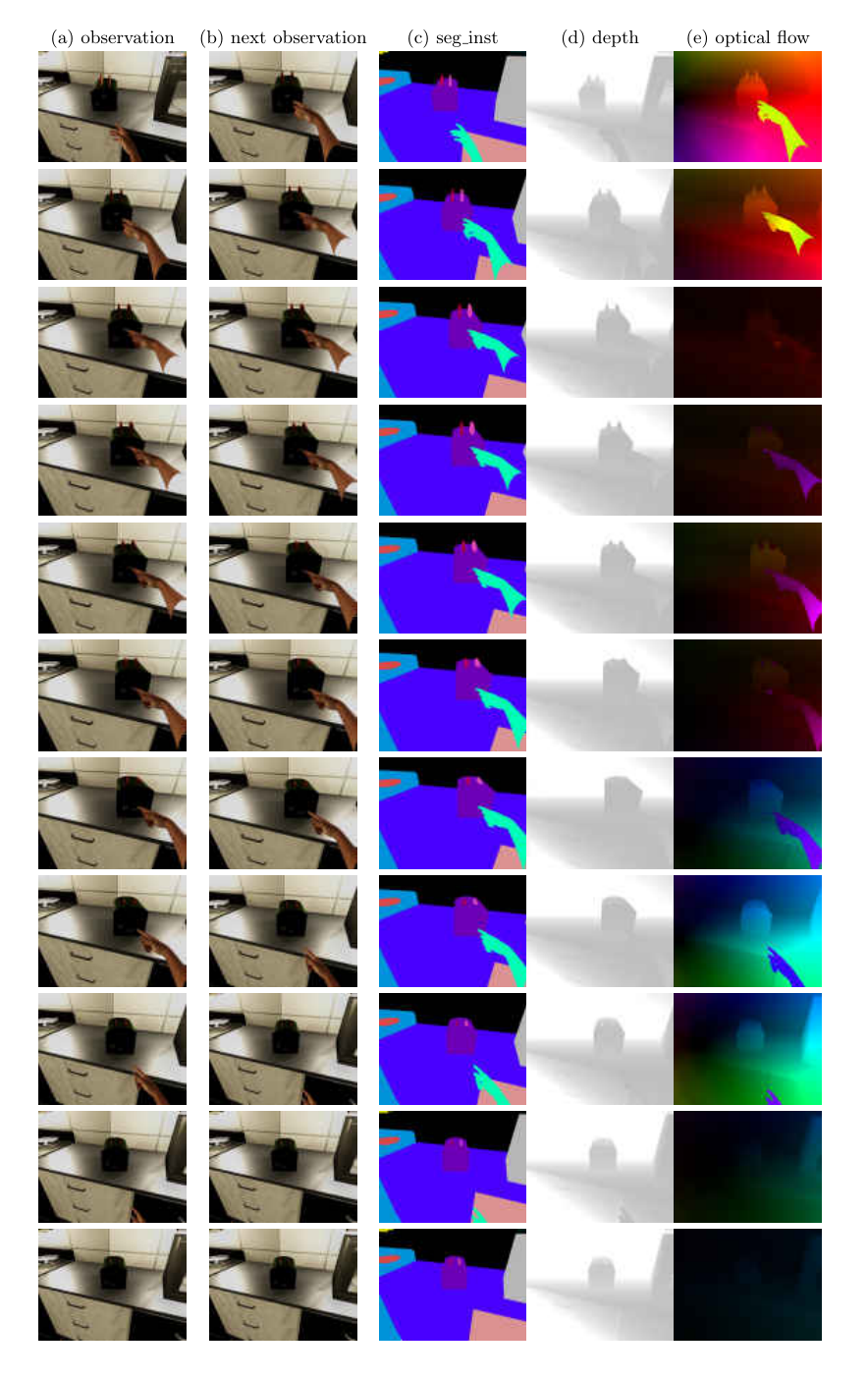}
  \caption{Samples in the sequence of switching on the toaster.}
  \label{3333}
\vspace{-0.4cm}
\end{figure}

\subsection{More examples of generating images}
More examples of generated images from EgoPlan can be seen in Figure \ref{fig:examples_appendix}. Each line represents a task, and the task prompts are, in order: "capture the chicken", "grasp juice", "grasp the hairproduct", "open the cabinet", "open the microwave", "go left", "make a left", "make a left-hand turn", "make a right", "turn right", "turn to the right", "walk straight ahead".

\begin{figure}
  \centering
  \begin{subfigure}{\subfigwidth\textwidth}
    \centering
    \includegraphics[width=\linewidth]{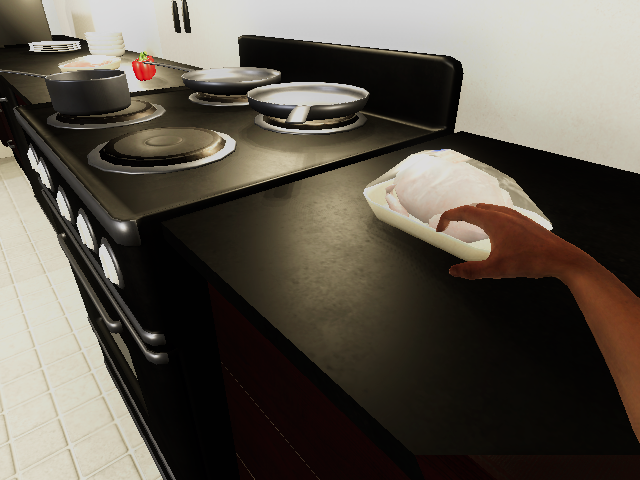}
    \label{subfig:01_Follow_Dutch}
  \end{subfigure}
    \hspace{-0.16cm}
  \begin{subfigure}{\subfigwidth\textwidth}
    \centering
    \includegraphics[width=\linewidth]{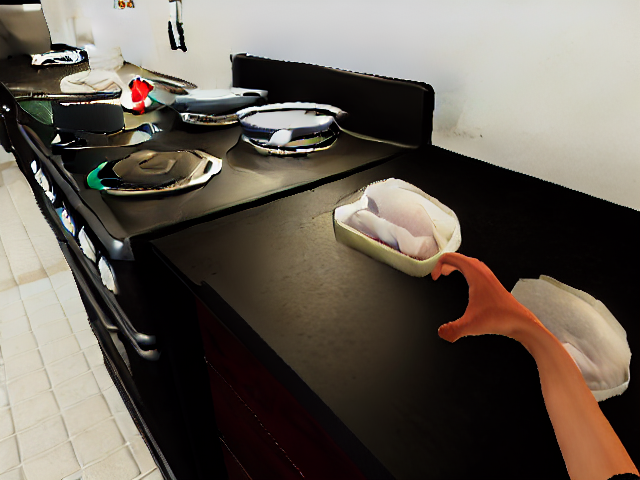} 
    \label{subfig:02_Hitch_horse}
  \end{subfigure}
    \hspace{-0.16cm}
  \begin{subfigure}{\subfigwidth\textwidth}
    \centering
    \includegraphics[width=\linewidth]{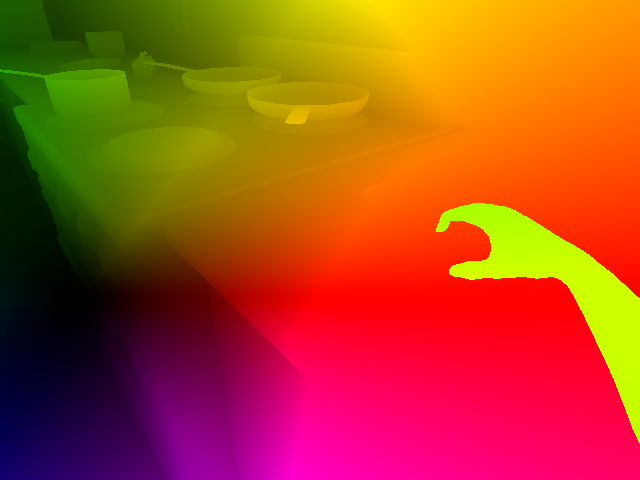}
    \label{subfig:03_Go_to_shed}
  \end{subfigure}
  \hspace{-0.16cm}
  \begin{subfigure}{\subfigwidth\textwidth}
    \centering
    \includegraphics[width=\linewidth]{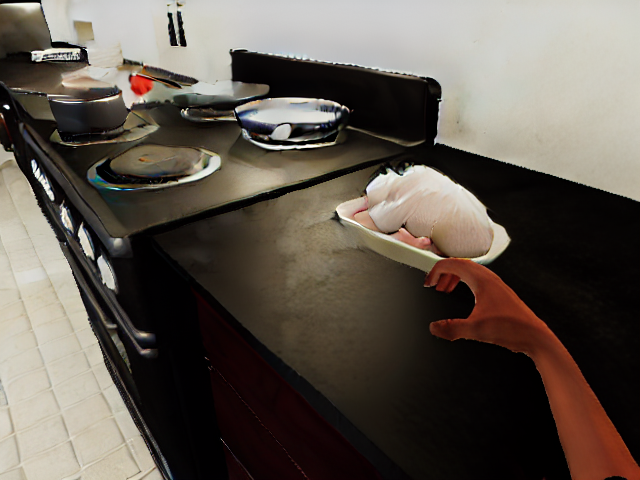}
    \label{subfig:04_Choose_weapon}
  \end{subfigure}
  \hspace{-0.16cm}
  \begin{subfigure}{\subfigwidth\textwidth}
    \centering
    \includegraphics[width=\linewidth]{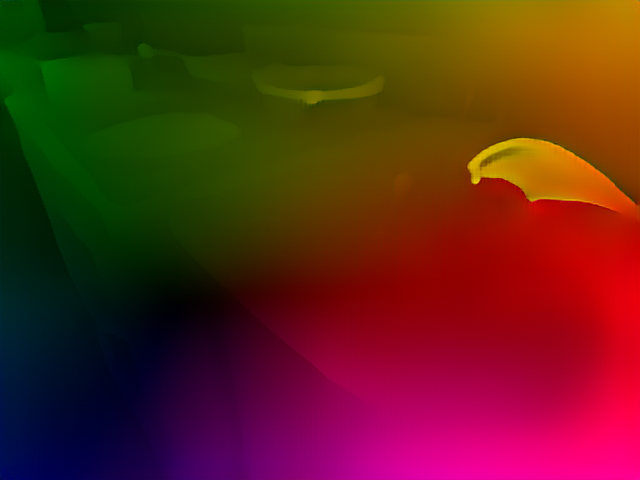}
    \label{subfig:05_Protect_Dutch}
  \end{subfigure}
  \hspace{-0.16cm}
  \begin{subfigure}{\subfigwidth\textwidth}
    \centering
    \includegraphics[width=\linewidth]{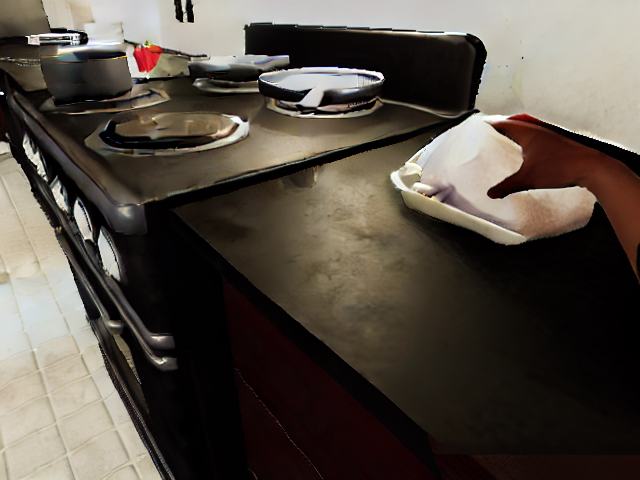}
    \label{subfig:06_Search_house}
  \end{subfigure}
  \vspace{-0.41cm}
  
    \centering
  \begin{subfigure}{\subfigwidth\textwidth}
    \centering
    \includegraphics[width=\linewidth]{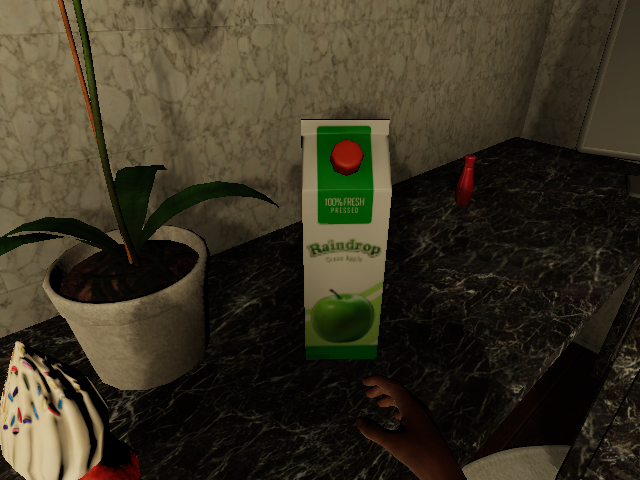}
    \label{subfig:01_Follow_Dutch}
  \end{subfigure}
    \hspace{-0.16cm}
  \begin{subfigure}{\subfigwidth\textwidth}
    \centering
    \includegraphics[width=\linewidth]{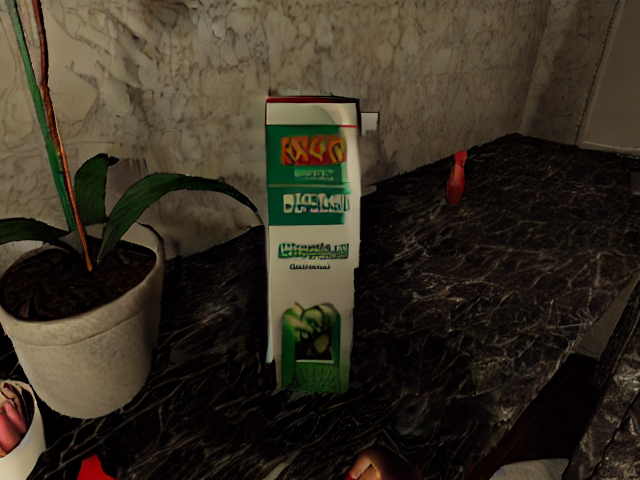} 
    \label{subfig:02_Hitch_horse}
  \end{subfigure}
    \hspace{-0.16cm}
  \begin{subfigure}{\subfigwidth\textwidth}
    \centering
    \includegraphics[width=\linewidth]{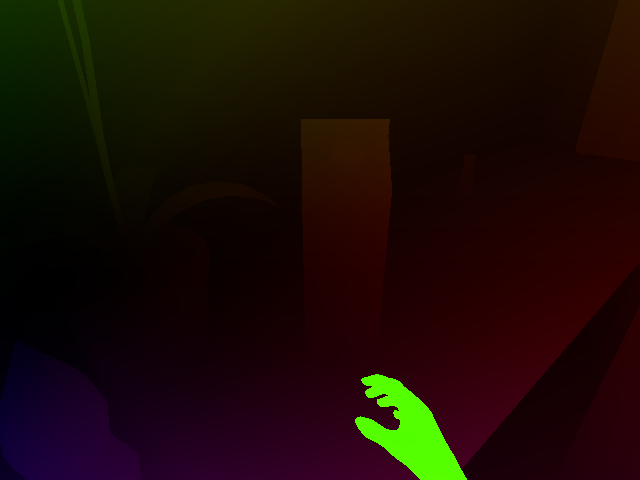}
    \label{subfig:03_Go_to_shed}
  \end{subfigure}
  \hspace{-0.16cm}
  \begin{subfigure}{\subfigwidth\textwidth}
    \centering
    \includegraphics[width=\linewidth]{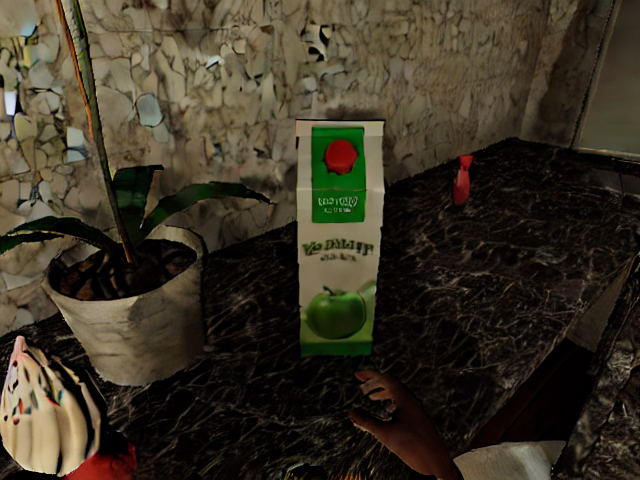}
    \label{subfig:04_Choose_weapon}
  \end{subfigure}
  \hspace{-0.16cm}
  \begin{subfigure}{\subfigwidth\textwidth}
    \centering
    \includegraphics[width=\linewidth]{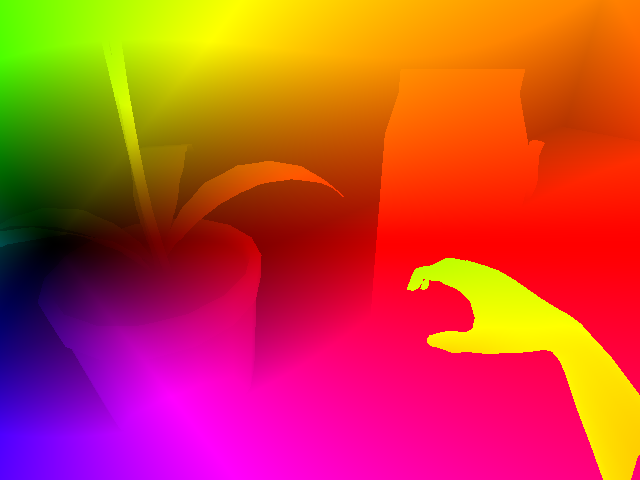}
    \label{subfig:05_Protect_Dutch}
  \end{subfigure}
  \hspace{-0.16cm}
  \begin{subfigure}{\subfigwidth\textwidth}
    \centering
    \includegraphics[width=\linewidth]{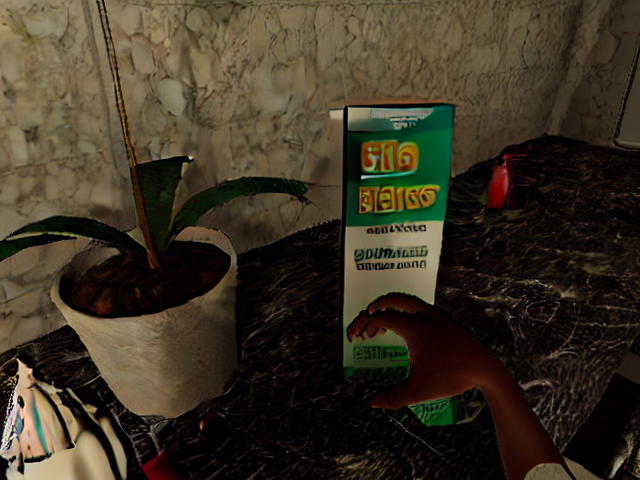}
    \label{subfig:06_Search_house}
  \end{subfigure}
  \vspace{-0.41cm}

      \centering
  \begin{subfigure}{\subfigwidth\textwidth}
    \centering
    \includegraphics[width=\linewidth]{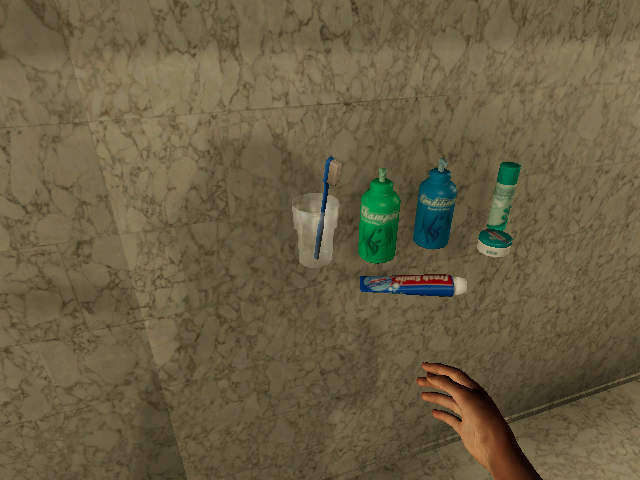}
    \label{subfig:01_Follow_Dutch}
  \end{subfigure}
    \hspace{-0.16cm}
  \begin{subfigure}{\subfigwidth\textwidth}
    \centering
    \includegraphics[width=\linewidth]{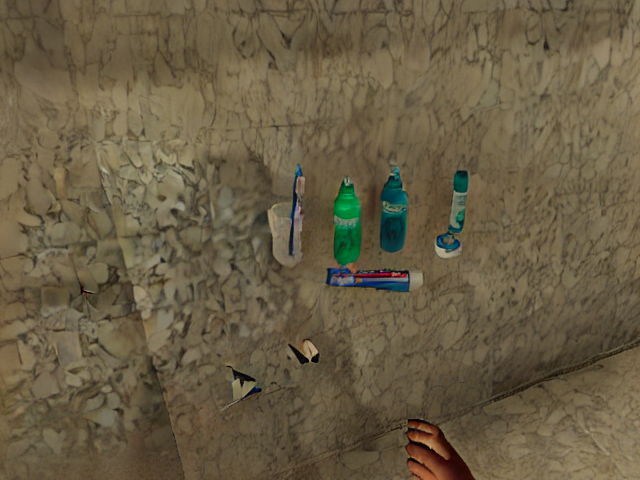} 
    \label{subfig:02_Hitch_horse}
  \end{subfigure}
    \hspace{-0.16cm}
  \begin{subfigure}{\subfigwidth\textwidth}
    \centering
    \includegraphics[width=\linewidth]{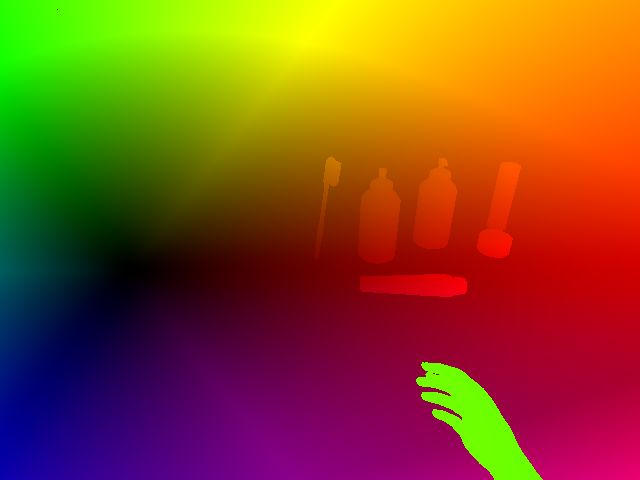}
    \label{subfig:03_Go_to_shed}
  \end{subfigure}
  \hspace{-0.16cm}
  \begin{subfigure}{\subfigwidth\textwidth}
    \centering
    \includegraphics[width=\linewidth]{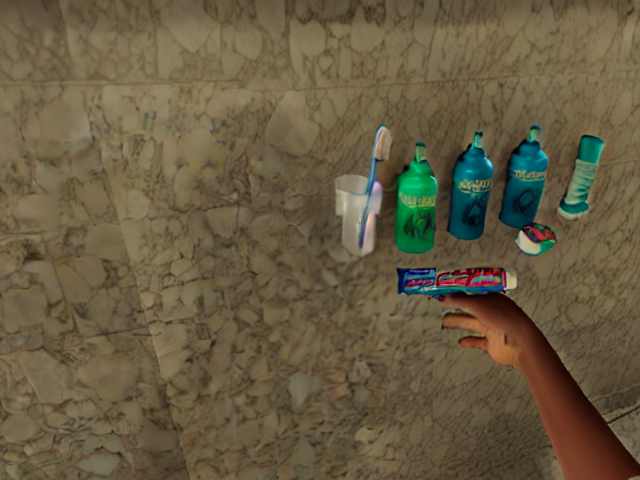}
    \label{subfig:04_Choose_weapon}
  \end{subfigure}
  \hspace{-0.16cm}
  \begin{subfigure}{\subfigwidth\textwidth}
    \centering
    \includegraphics[width=\linewidth]{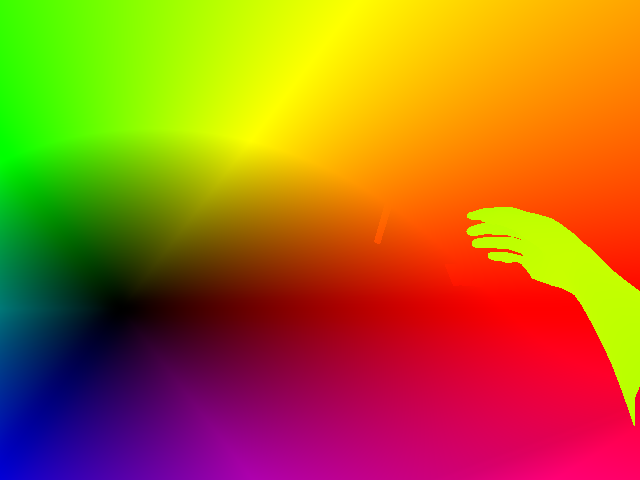}
    \label{subfig:05_Protect_Dutch}
  \end{subfigure}
  \hspace{-0.16cm}
  \begin{subfigure}{\subfigwidth\textwidth}
    \centering
    \includegraphics[width=\linewidth]{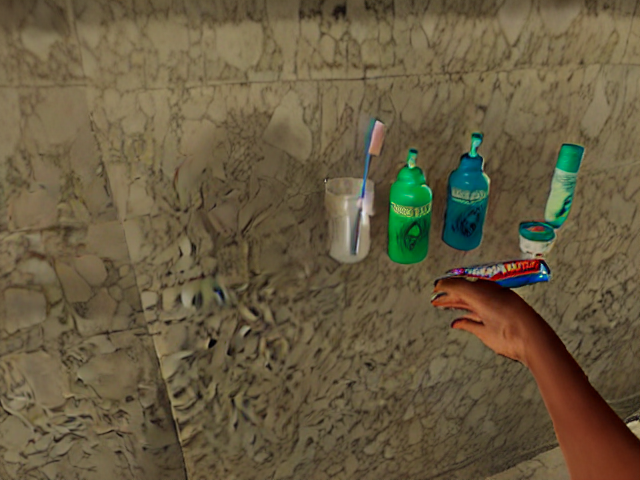}
    \label{subfig:06_Search_house}
  \end{subfigure}
  \vspace{-0.41cm}

    \centering
  \begin{subfigure}{\subfigwidth\textwidth}
    \centering
    \includegraphics[width=\linewidth]{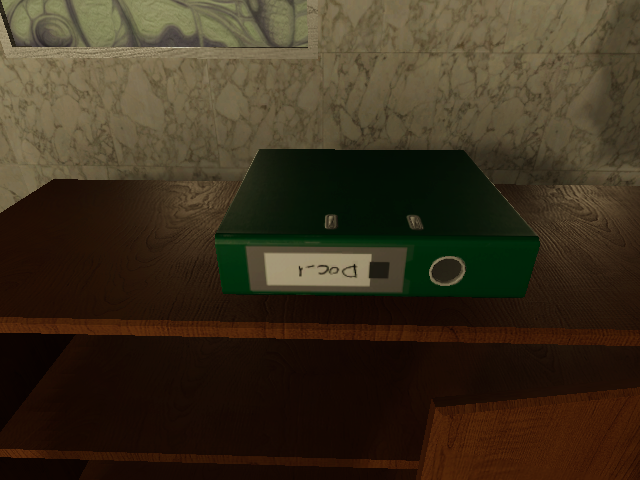}
    \label{subfig:01_Follow_Dutch}
  \end{subfigure}
    \hspace{-0.16cm}
  \begin{subfigure}{\subfigwidth\textwidth}
    \centering
    \includegraphics[width=\linewidth]{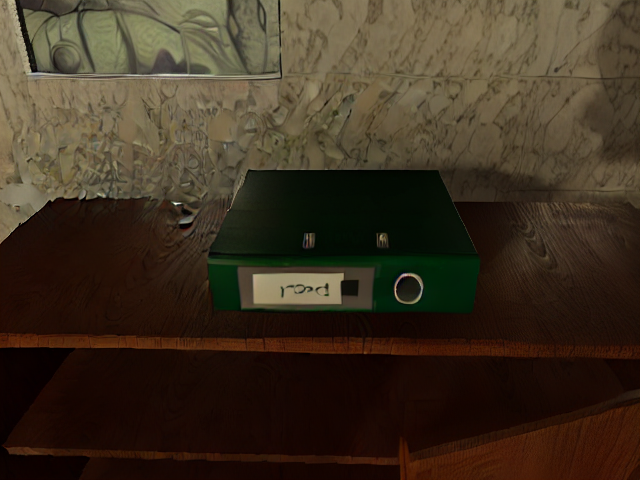} 
    \label{subfig:02_Hitch_horse}
  \end{subfigure}
    \hspace{-0.16cm}
  \begin{subfigure}{\subfigwidth\textwidth}
    \centering
    \includegraphics[width=\linewidth]{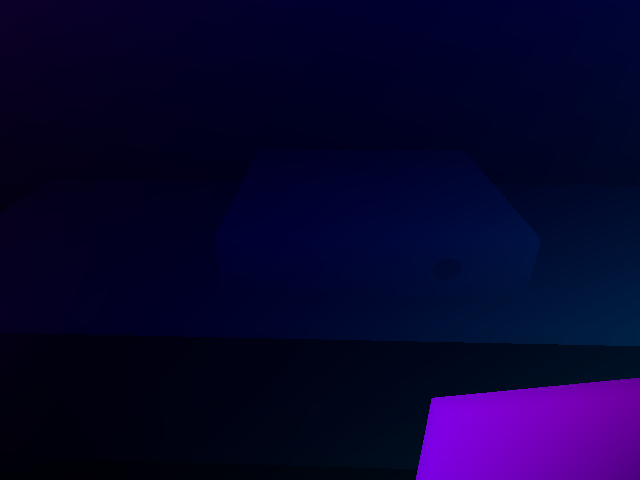}
    \label{subfig:03_Go_to_shed}
  \end{subfigure}
  \hspace{-0.16cm}
  \begin{subfigure}{\subfigwidth\textwidth}
    \centering
    \includegraphics[width=\linewidth]{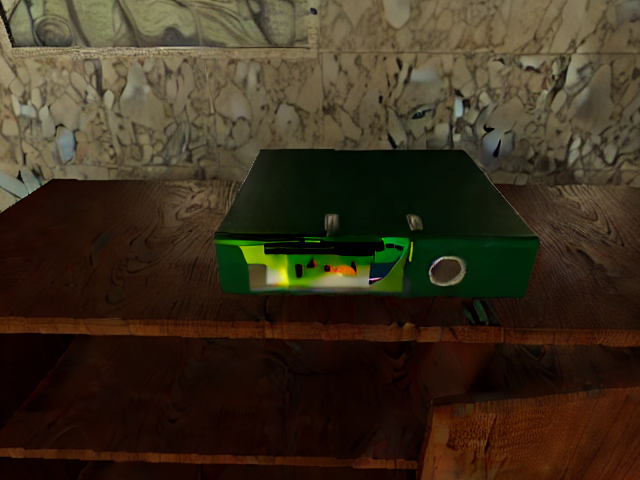}
    \label{subfig:04_Choose_weapon}
  \end{subfigure}
  \hspace{-0.16cm}
  \begin{subfigure}{\subfigwidth\textwidth}
    \centering
    \includegraphics[width=\linewidth]{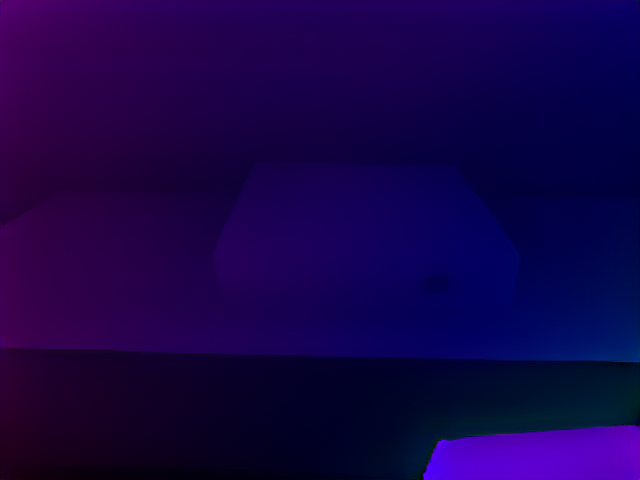}
    \label{subfig:05_Protect_Dutch}
  \end{subfigure}
  \hspace{-0.16cm}
  \begin{subfigure}{\subfigwidth\textwidth}
    \centering
    \includegraphics[width=\linewidth]{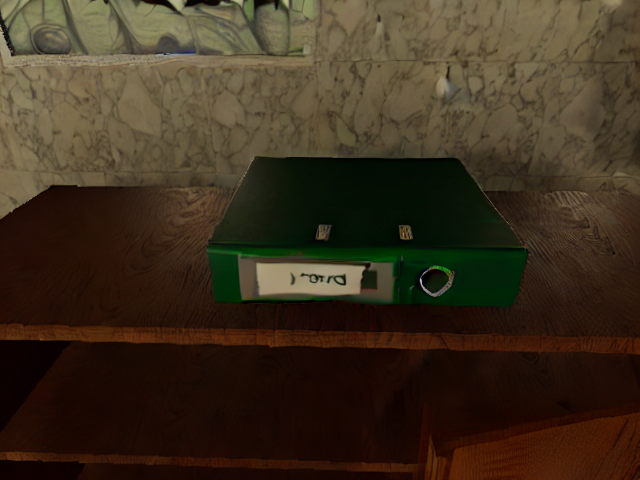}
    \label{subfig:06_Search_house}
  \end{subfigure}
  \vspace{-0.41cm}

    \centering
  \begin{subfigure}{\subfigwidth\textwidth}
    \centering
    \includegraphics[width=\linewidth]{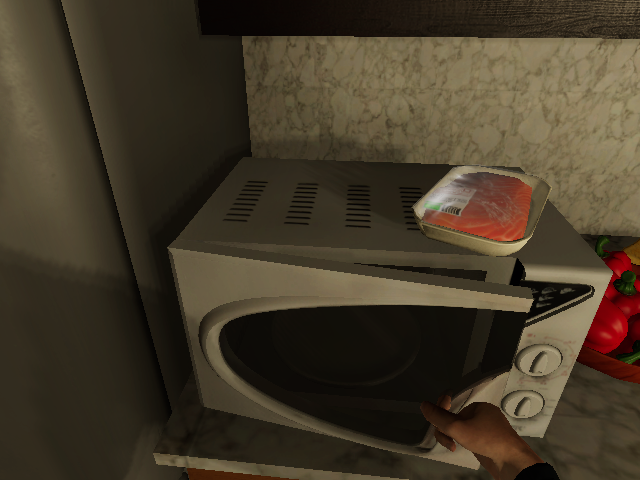}
    \label{subfig:01_Follow_Dutch}
  \end{subfigure}
    \hspace{-0.16cm}
  \begin{subfigure}{\subfigwidth\textwidth}
    \centering
    \includegraphics[width=\linewidth]{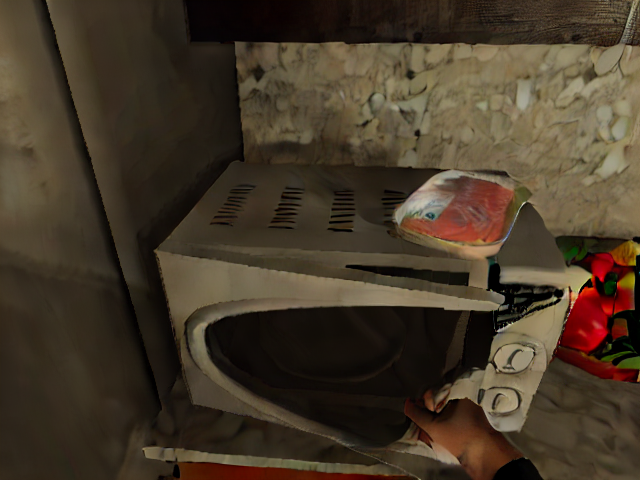} 
    \label{subfig:02_Hitch_horse}
  \end{subfigure}
    \hspace{-0.16cm}
  \begin{subfigure}{\subfigwidth\textwidth}
    \centering
    \includegraphics[width=\linewidth]{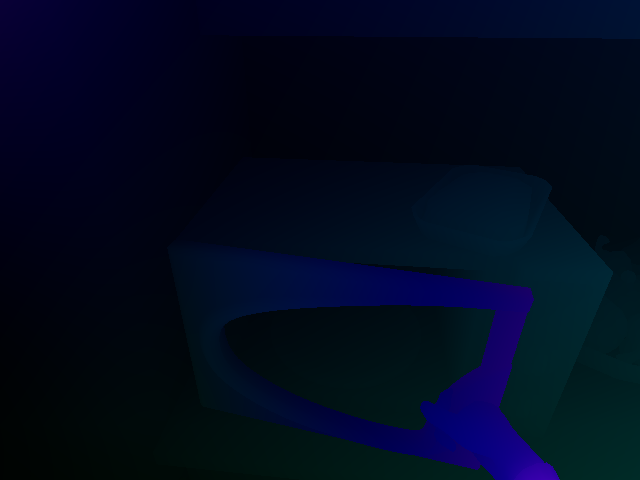}
    \label{subfig:03_Go_to_shed}
  \end{subfigure}
  \hspace{-0.16cm}
  \begin{subfigure}{\subfigwidth\textwidth}
    \centering
    \includegraphics[width=\linewidth]{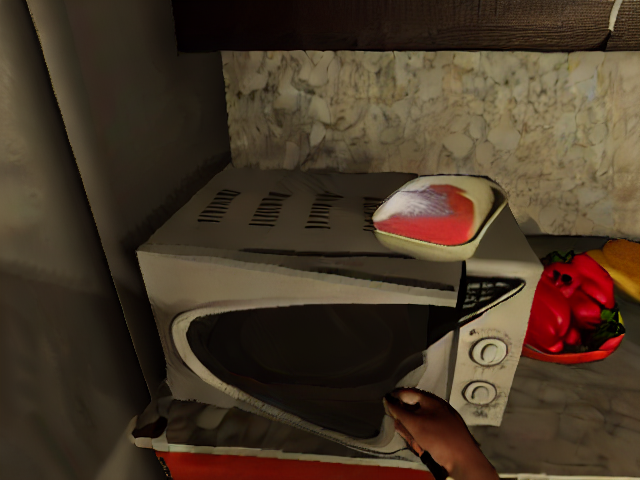}
    \label{subfig:04_Choose_weapon}
  \end{subfigure}
  \hspace{-0.16cm}
  \begin{subfigure}{\subfigwidth\textwidth}
    \centering
    \includegraphics[width=\linewidth]{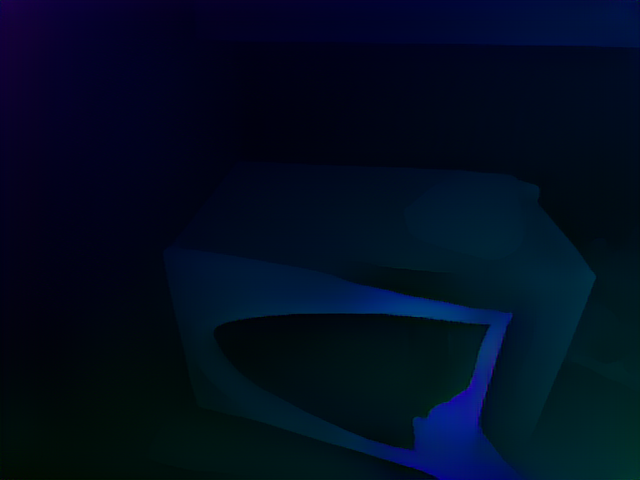}
    \label{subfig:05_Protect_Dutch}
  \end{subfigure}
  \hspace{-0.16cm}
  \begin{subfigure}{\subfigwidth\textwidth}
    \centering
    \includegraphics[width=\linewidth]{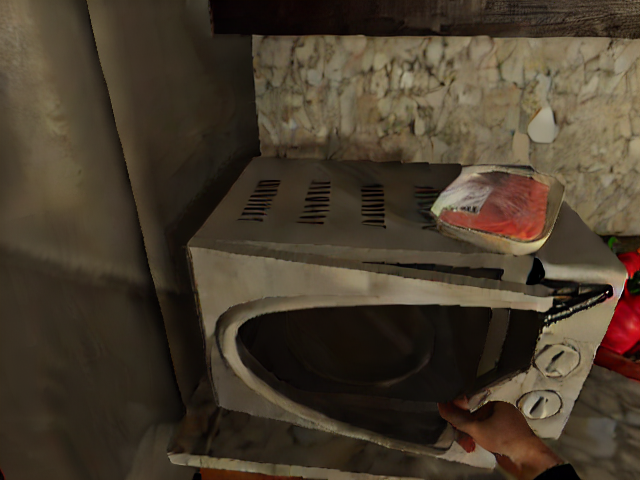}
    \label{subfig:06_Search_house}
  \end{subfigure}
  \vspace{-0.41cm}

  \centering
  \begin{subfigure}{\subfigwidth\textwidth}
    \centering
    \includegraphics[width=\linewidth]{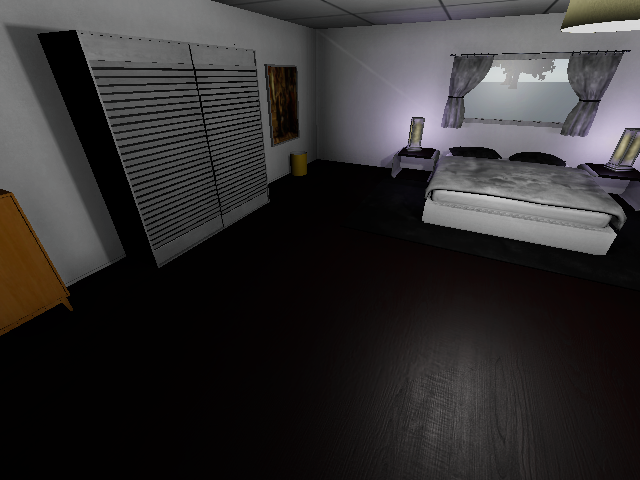}
    \label{subfig:01_Follow_Dutch}
  \end{subfigure}
    \hspace{-0.16cm}
  \begin{subfigure}{\subfigwidth\textwidth}
    \centering
    \includegraphics[width=\linewidth]{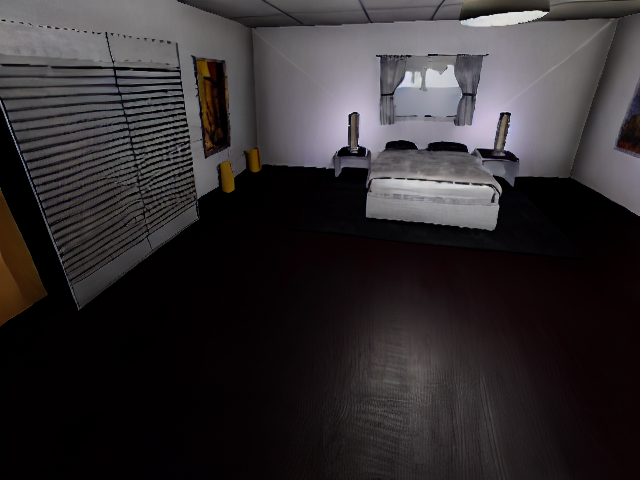} 
    \label{subfig:02_Hitch_horse}
  \end{subfigure}
    \hspace{-0.16cm}
  \begin{subfigure}{\subfigwidth\textwidth}
    \centering
    \includegraphics[width=\linewidth]{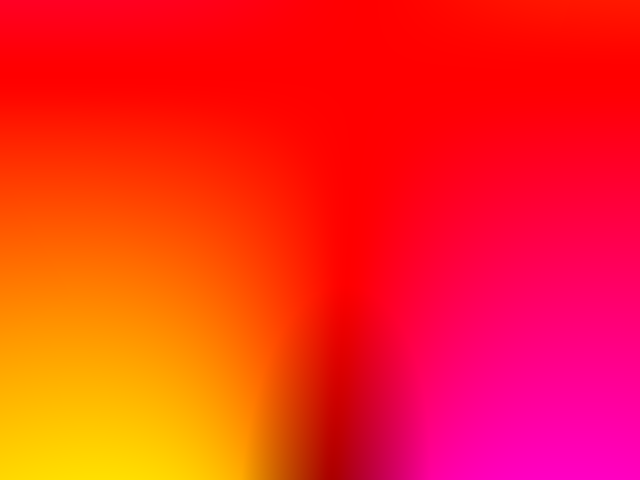}
    \label{subfig:03_Go_to_shed}
  \end{subfigure}
  \hspace{-0.16cm}
  \begin{subfigure}{\subfigwidth\textwidth}
    \centering
    \includegraphics[width=\linewidth]{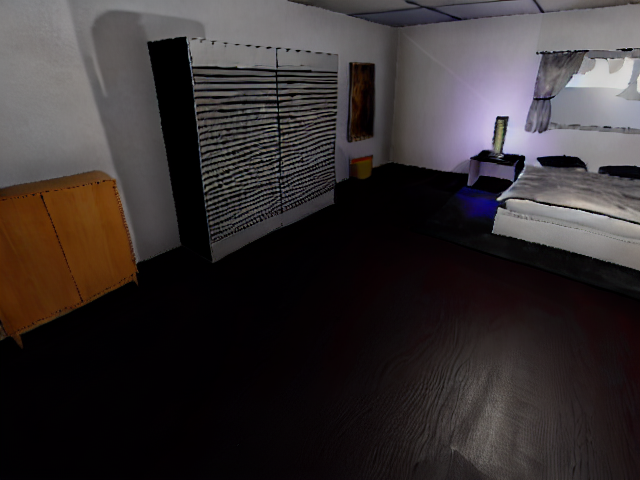}
    \label{subfig:04_Choose_weapon}
  \end{subfigure}
  \hspace{-0.16cm}
  \begin{subfigure}{\subfigwidth\textwidth}
    \centering
    \includegraphics[width=\linewidth]{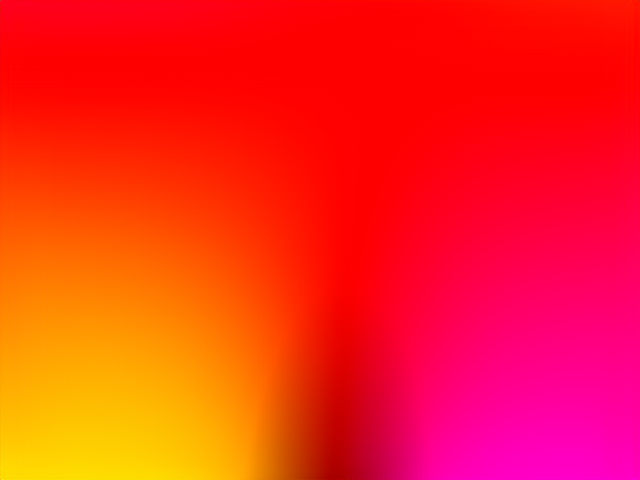}
    \label{subfig:05_Protect_Dutch}
  \end{subfigure}
  \hspace{-0.16cm}
  \begin{subfigure}{\subfigwidth\textwidth}
    \centering
    \includegraphics[width=\linewidth]{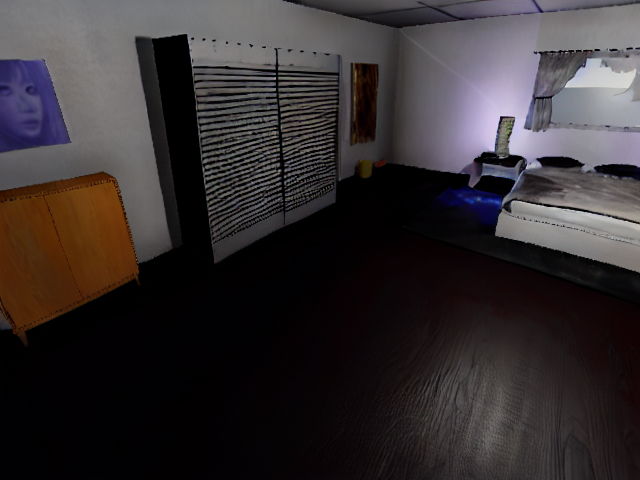}
    \label{subfig:06_Search_house}
  \end{subfigure}
  \vspace{-0.41cm}

  \centering
  \begin{subfigure}{\subfigwidth\textwidth}
    \centering
    \includegraphics[width=\linewidth]{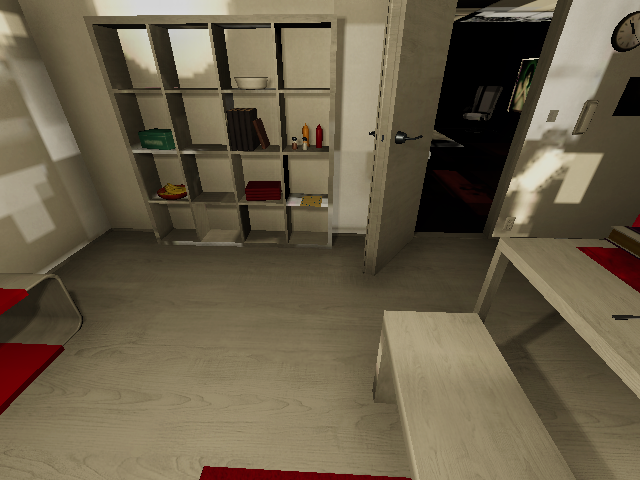}
    \label{subfig:01_Follow_Dutch}
  \end{subfigure}
    \hspace{-0.16cm}
  \begin{subfigure}{\subfigwidth\textwidth}
    \centering
    \includegraphics[width=\linewidth]{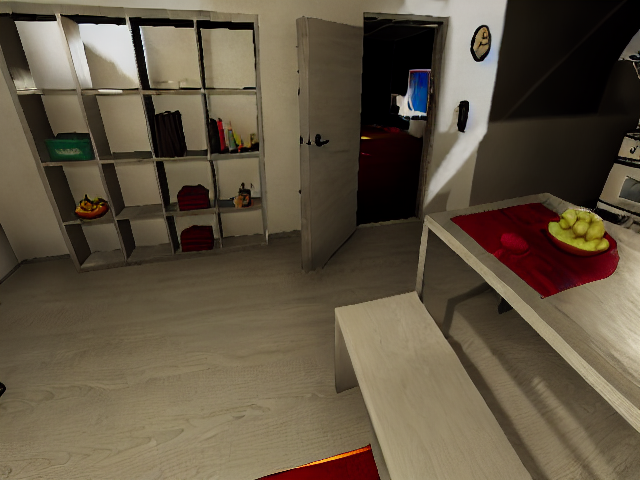} 
    \label{subfig:02_Hitch_horse}
  \end{subfigure}
    \hspace{-0.16cm}
  \begin{subfigure}{\subfigwidth\textwidth}
    \centering
    \includegraphics[width=\linewidth]{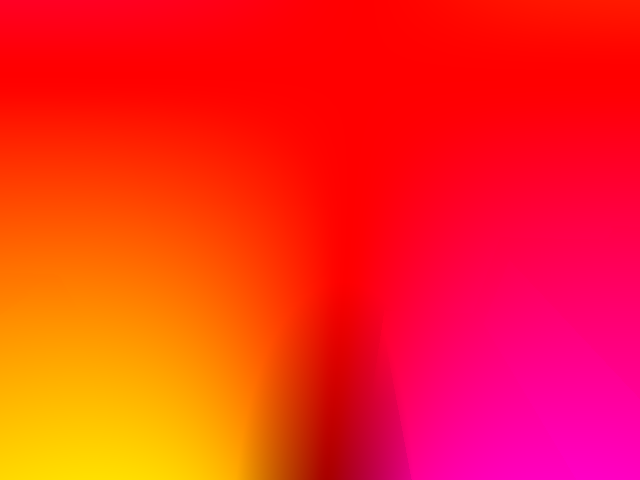}
    \label{subfig:03_Go_to_shed}
  \end{subfigure}
  \hspace{-0.16cm}
  \begin{subfigure}{\subfigwidth\textwidth}
    \centering
    \includegraphics[width=\linewidth]{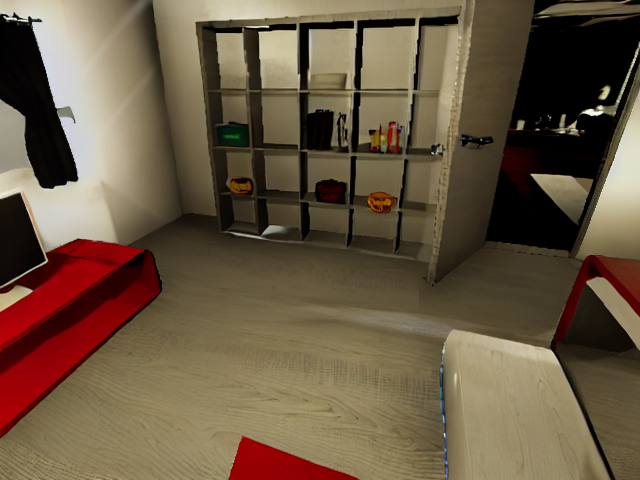}
    \label{subfig:04_Choose_weapon}
  \end{subfigure}
  \hspace{-0.16cm}
  \begin{subfigure}{\subfigwidth\textwidth}
    \centering
    \includegraphics[width=\linewidth]{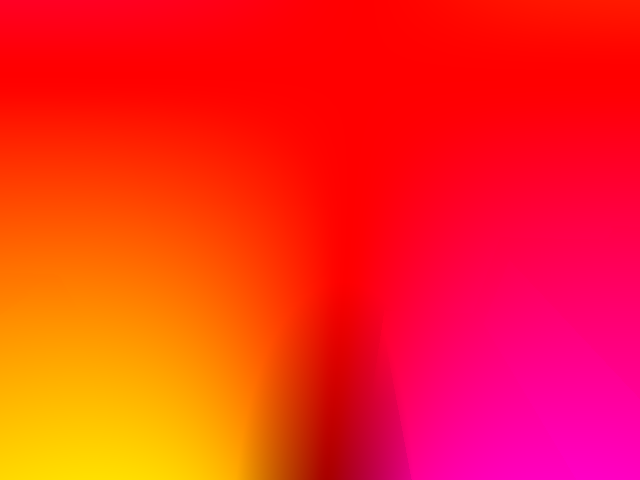}
    \label{subfig:05_Protect_Dutch}
  \end{subfigure}
  \hspace{-0.16cm}
  \begin{subfigure}{\subfigwidth\textwidth}
    \centering
    \includegraphics[width=\linewidth]{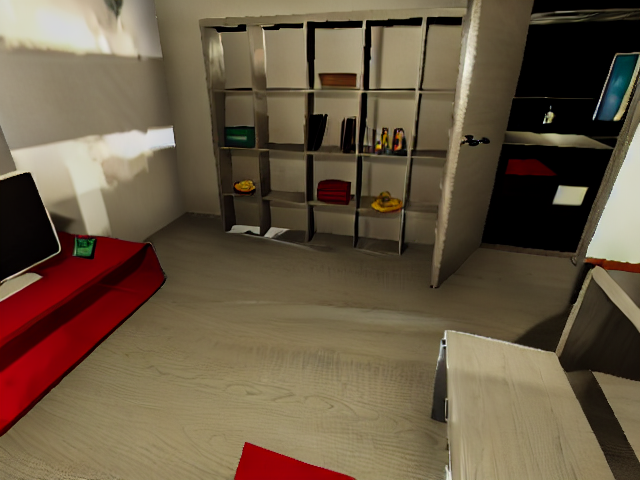}
    \label{subfig:06_Search_house}
  \end{subfigure}
  \vspace{-0.41cm}

  \centering
  \begin{subfigure}{\subfigwidth\textwidth}
    \centering
    \includegraphics[width=\linewidth]{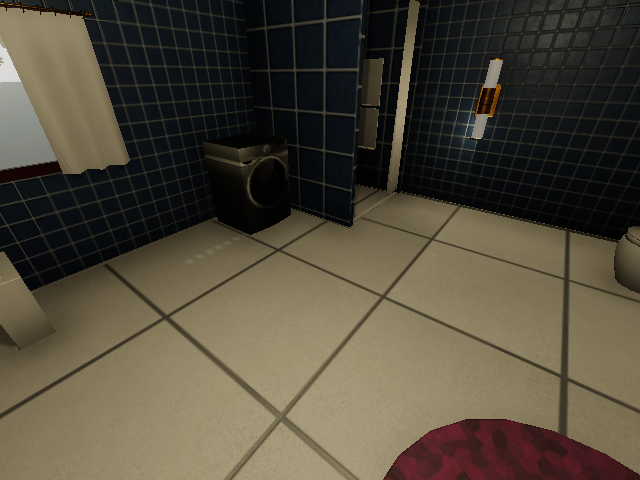}
    \label{subfig:01_Follow_Dutch}
  \end{subfigure}
    \hspace{-0.16cm}
  \begin{subfigure}{\subfigwidth\textwidth}
    \centering
    \includegraphics[width=\linewidth]{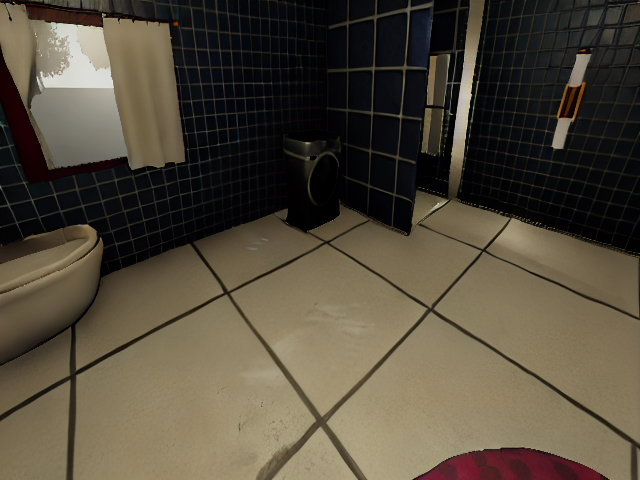} 
    \label{subfig:02_Hitch_horse}
  \end{subfigure}
    \hspace{-0.16cm}
  \begin{subfigure}{\subfigwidth\textwidth}
    \centering
    \includegraphics[width=\linewidth]{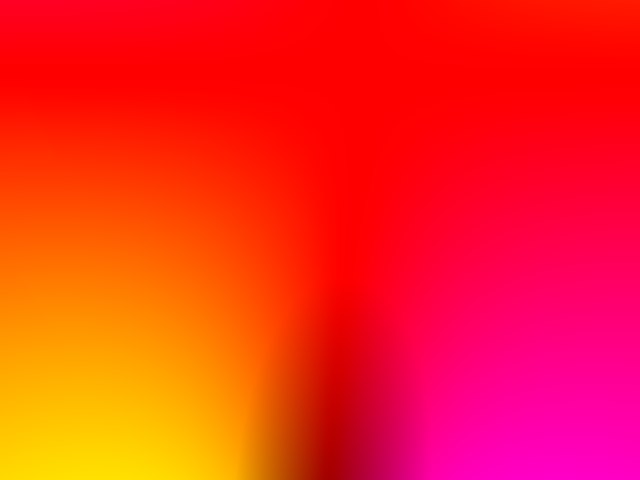}
    \label{subfig:03_Go_to_shed}
  \end{subfigure}
  \hspace{-0.16cm}
  \begin{subfigure}{\subfigwidth\textwidth}
    \centering
    \includegraphics[width=\linewidth]{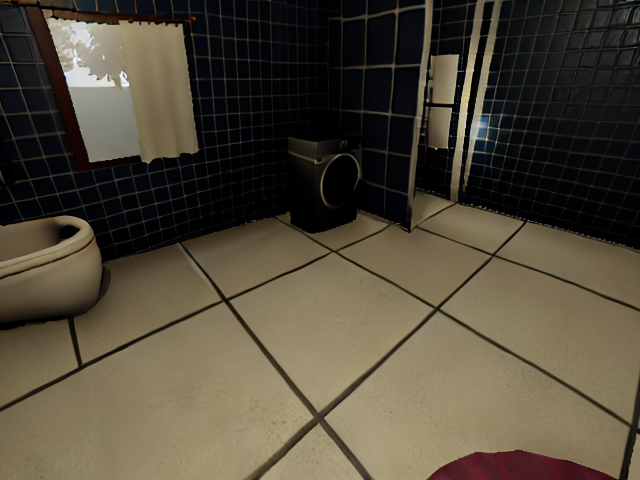}
    \label{subfig:04_Choose_weapon}
  \end{subfigure}
  \hspace{-0.16cm}
  \begin{subfigure}{\subfigwidth\textwidth}
    \centering
    \includegraphics[width=\linewidth]{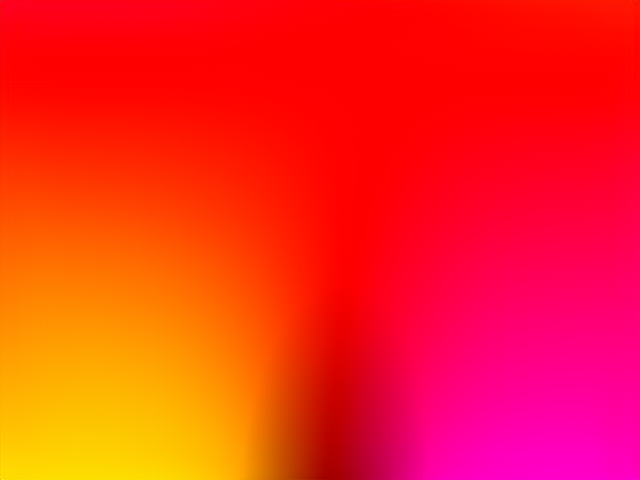}
    \label{subfig:05_Protect_Dutch}
  \end{subfigure}
  \hspace{-0.16cm}
  \begin{subfigure}{\subfigwidth\textwidth}
    \centering
    \includegraphics[width=\linewidth]{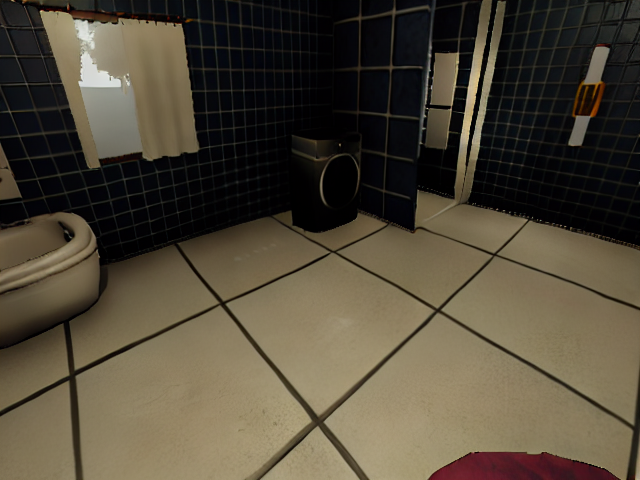}
    \label{subfig:06_Search_house}
  \end{subfigure}
  \vspace{-0.41cm}
  
  \begin{subfigure}{\subfigwidth\textwidth}
    \centering
    \includegraphics[width=\linewidth]{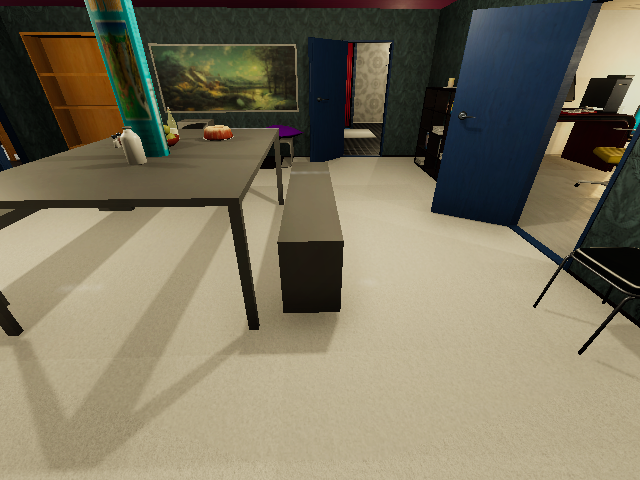}
    \label{subfig:01_Follow_Dutch}
  \end{subfigure}
  \hspace{-0.16cm}
  \begin{subfigure}{\subfigwidth\textwidth}
    \centering
    \includegraphics[width=\linewidth]{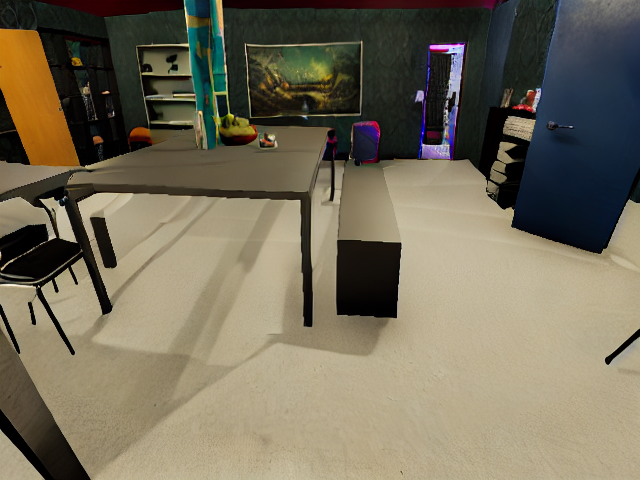} 
    \label{subfig:02_Hitch_horse}
  \end{subfigure}
  \hspace{-0.16cm}
  \begin{subfigure}{\subfigwidth\textwidth}
    \centering
    \includegraphics[width=\linewidth]{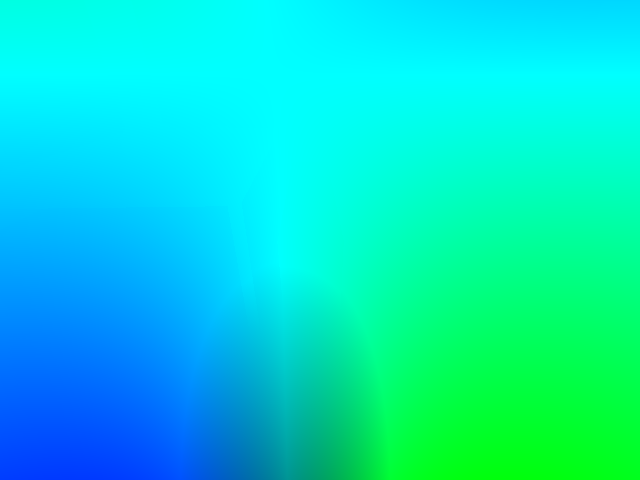}
    \label{subfig:03_Go_to_shed}
  \end{subfigure}
  \hspace{-0.16cm}
  \begin{subfigure}{\subfigwidth\textwidth}
    \centering
    \includegraphics[width=\linewidth]{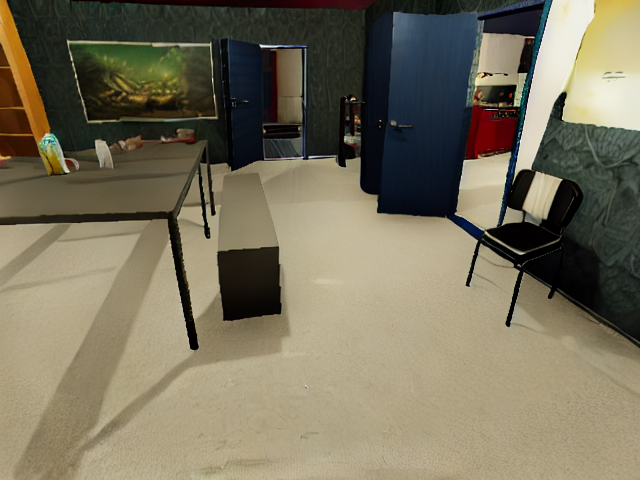}
    \label{subfig:04_Choose_weapon}
  \end{subfigure}
  \hspace{-0.16cm}
  \begin{subfigure}{\subfigwidth\textwidth}
    \centering
    \includegraphics[width=\linewidth]{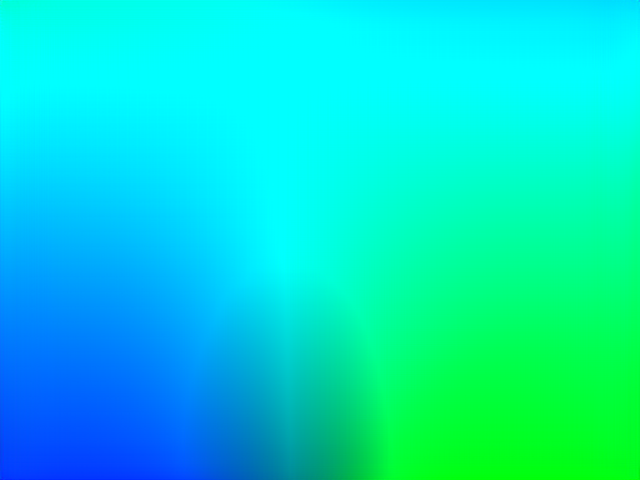}
    \label{subfig:05_Protect_Dutch}
  \end{subfigure}
  \hspace{-0.16cm}
  \begin{subfigure}{\subfigwidth\textwidth}
    \centering
    \includegraphics[width=\linewidth]{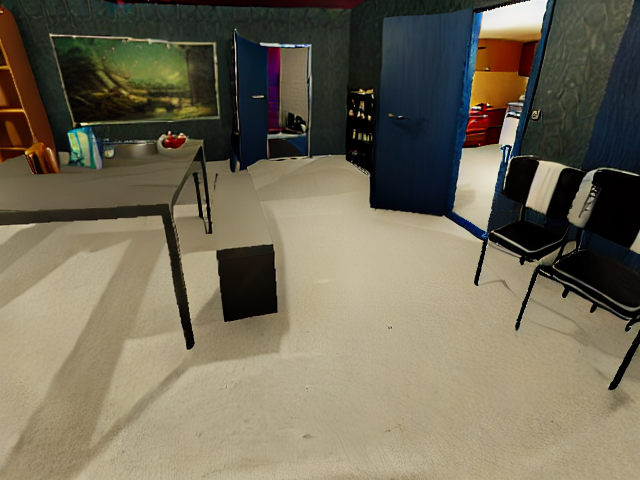}
    \label{subfig:06_Search_house}
  \end{subfigure}
  \vspace{-0.40cm}
  
  \begin{subfigure}{\subfigwidth\textwidth}
    \centering
    \includegraphics[width=\linewidth]{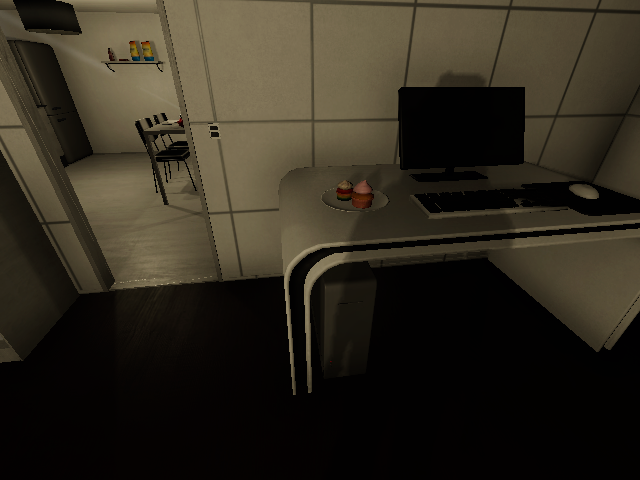}
    \label{subfig:01_Follow_Dutch}
  \end{subfigure}
  \hspace{-0.16cm}
  \begin{subfigure}{\subfigwidth\textwidth}
    \centering
    \includegraphics[width=\linewidth]{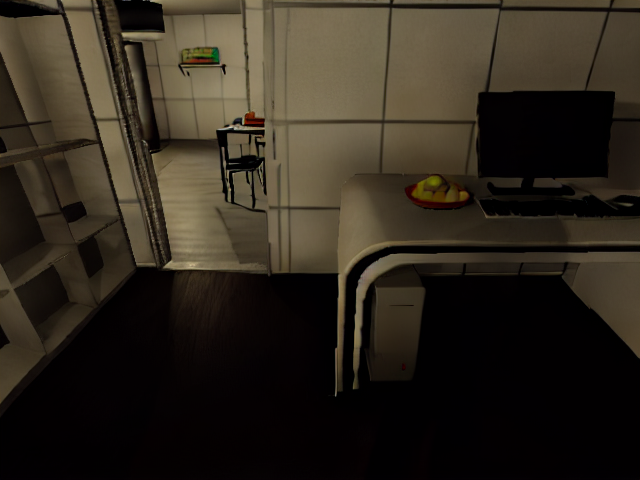} 
    \label{subfig:02_Hitch_horse}
  \end{subfigure}
  \hspace{-0.16cm}
  \begin{subfigure}{\subfigwidth\textwidth}
    \centering
    \includegraphics[width=\linewidth]{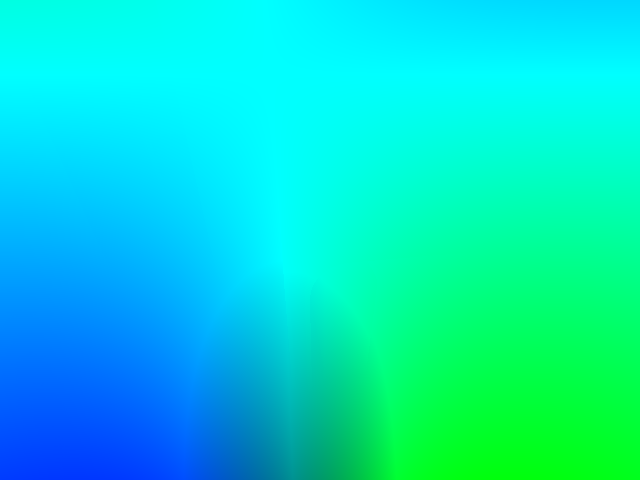}
    \label{subfig:03_Go_to_shed}
  \end{subfigure}
  \hspace{-0.16cm}
  \begin{subfigure}{\subfigwidth\textwidth}
    \centering
    \includegraphics[width=\linewidth]{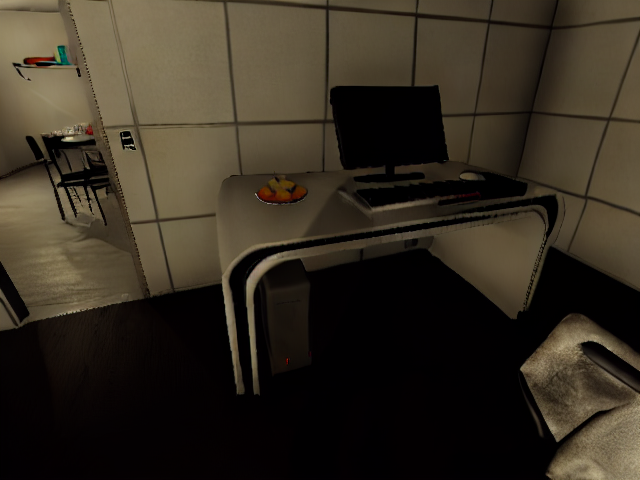}
    \label{subfig:04_Choose_weapon}
  \end{subfigure}
  \hspace{-0.16cm}
  \begin{subfigure}{\subfigwidth\textwidth}
    \centering
    \includegraphics[width=\linewidth]{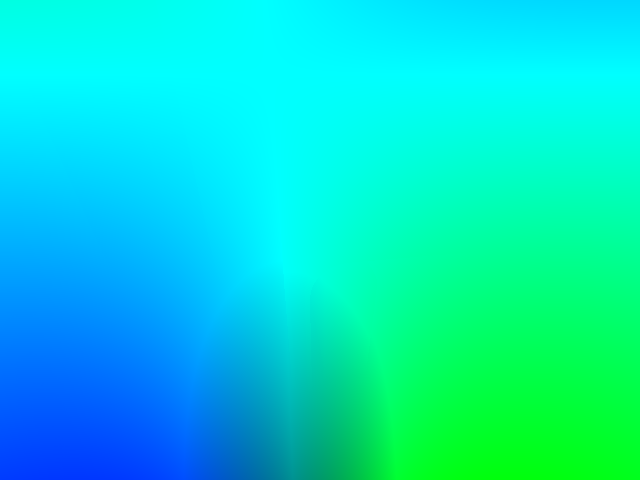}
    \label{subfig:05_Protect_Dutch}
  \end{subfigure}
  \hspace{-0.16cm}
  \begin{subfigure}{\subfigwidth\textwidth}
    \centering
    \includegraphics[width=\linewidth]{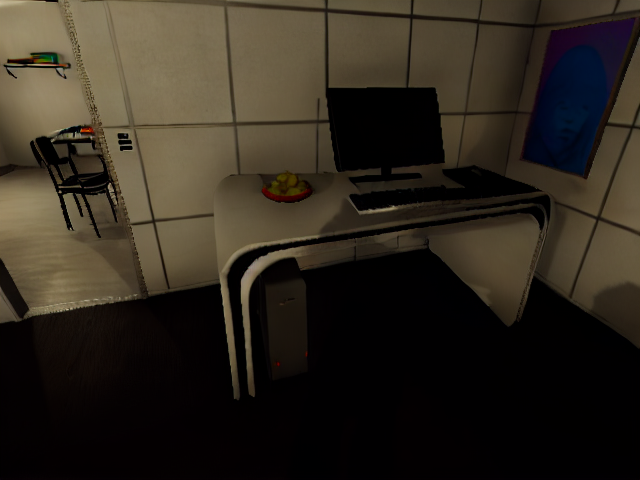}
    \label{subfig:06_Search_house}
  \end{subfigure}
  \vspace{-0.4cm}

   \begin{subfigure}{\subfigwidth\textwidth}
    \centering
    \includegraphics[width=\linewidth]{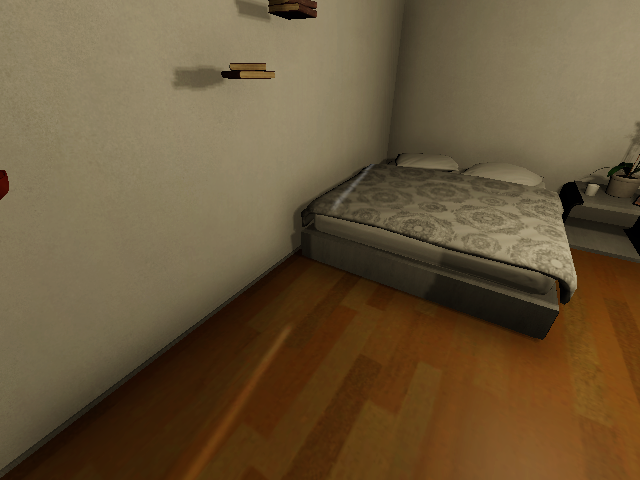} 
    \label{subfig:01_Follow_Dutch}
  \end{subfigure}
   \hspace{-0.16cm}
  \begin{subfigure}{\subfigwidth\textwidth}
    \centering
    \includegraphics[width=\linewidth]{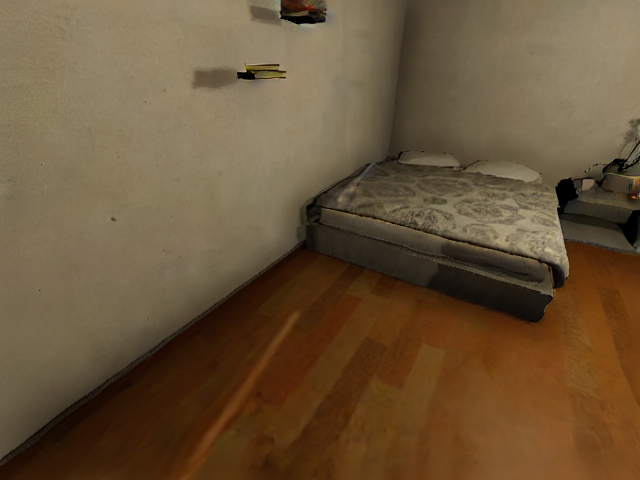} 
    \label{subfig:02_Hitch_horse}
  \end{subfigure}
   \hspace{-0.16cm}
  \begin{subfigure}{\subfigwidth\textwidth}
    \centering
    \includegraphics[width=\linewidth]{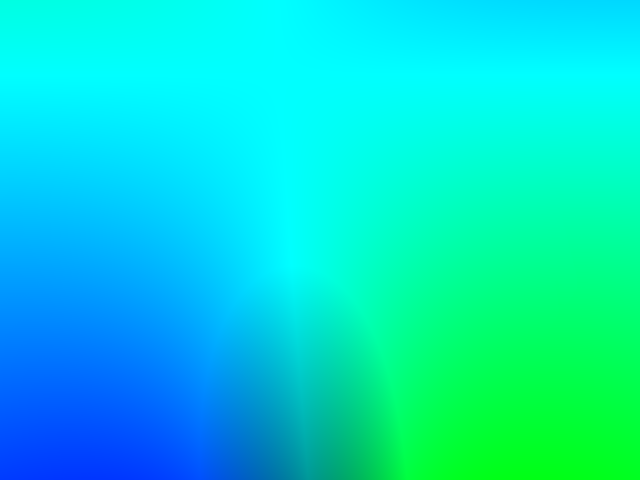}
    \label{subfig:03_Go_to_shed}
  \end{subfigure}
   \hspace{-0.16cm}
  \begin{subfigure}{\subfigwidth\textwidth}
    \centering
    \includegraphics[width=\linewidth]{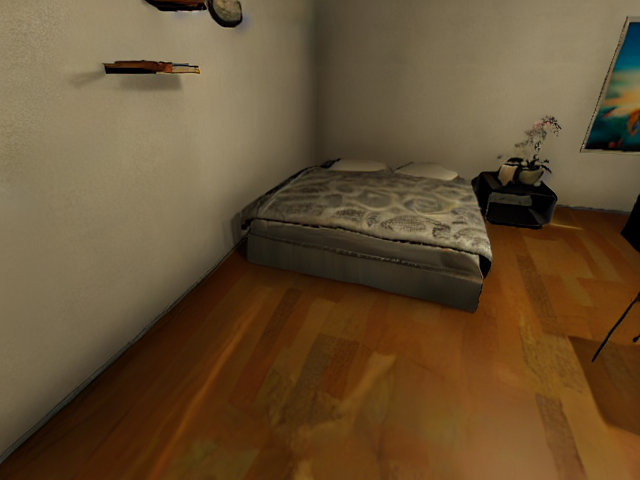}
    \label{subfig:04_Choose_weapon}
  \end{subfigure}
   \hspace{-0.16cm}
  \begin{subfigure}{\subfigwidth\textwidth}
    \centering
    \includegraphics[width=\linewidth]{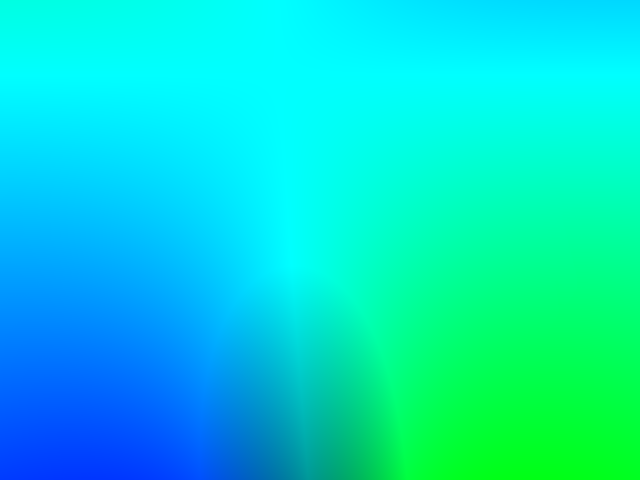}
    \label{subfig:05_Protect_Dutch}
  \end{subfigure}
   \hspace{-0.16cm}
  \begin{subfigure}{\subfigwidth\textwidth}
    \centering
    \includegraphics[width=\linewidth]{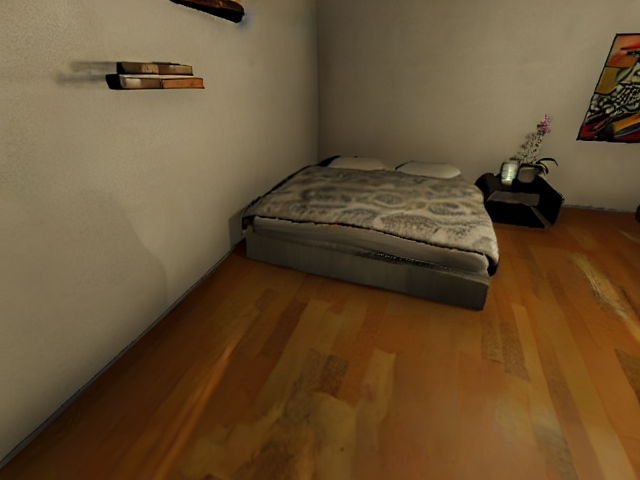}
    \label{subfig:06_Search_house}
  \end{subfigure}
  \vspace{-0.4cm}

  \begin{subfigure}{\subfigwidth\textwidth}
    \centering
    \includegraphics[width=\linewidth]{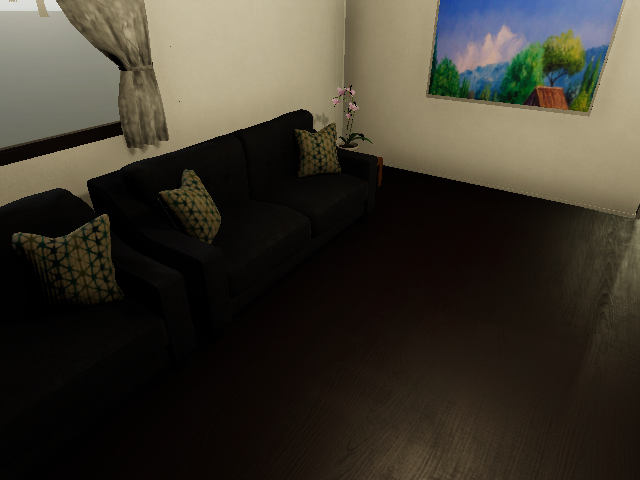}
    \captionsetup{font=tiny}
    \caption{Original}
    \label{subfig:01_Follow_Dutch}
  \end{subfigure}
  \hspace{-0.16cm}
  \begin{subfigure}{\subfigwidth\textwidth}
    \centering
    \includegraphics[width=\linewidth]{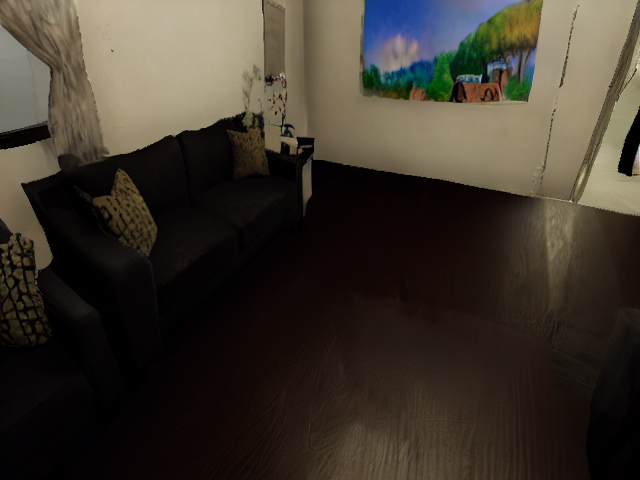} 
    \captionsetup{font=tiny}
    \caption{InstructP2P (finetuned)}
    \label{subfig:02_Hitch_horse}
  \end{subfigure}
  \hspace{-0.16cm}
  \begin{subfigure}{\subfigwidth\textwidth}
    \centering
    \includegraphics[width=\linewidth]{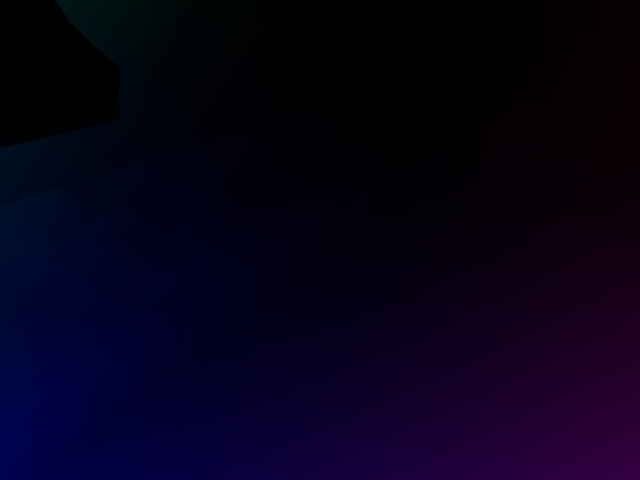}
    \captionsetup{font=tiny}
     \caption{Previous flow} 
    \label{subfig:03_Go_to_shed}
  \end{subfigure}
  \hspace{-0.16cm}
  \begin{subfigure}{\subfigwidth\textwidth}
    \centering
    \includegraphics[width=\linewidth]{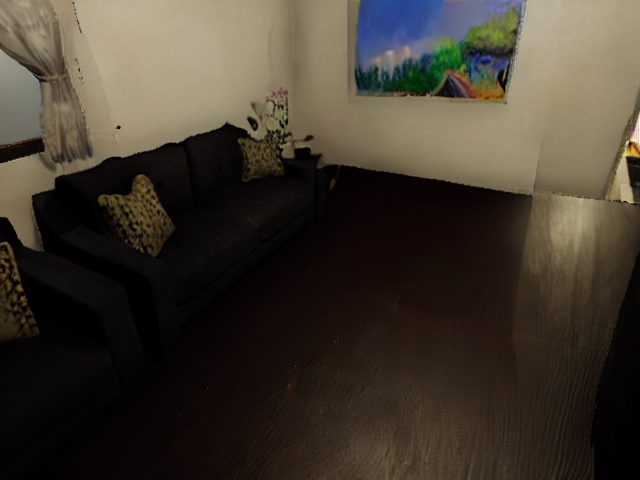}
    \captionsetup{font=tiny}
    \caption{Ours (previous flow)}
    \label{subfig:04_Choose_weapon}
  \end{subfigure}
  \hspace{-0.16cm}
  \begin{subfigure}{\subfigwidth\textwidth}
    \centering
    \includegraphics[width=\linewidth]{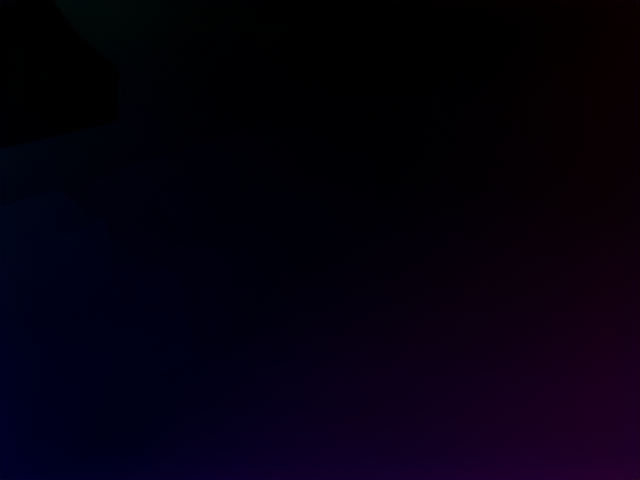}
    \captionsetup{font=tiny}
    \caption{Predicted flow}
    \label{subfig:05_Protect_Dutch}
  \end{subfigure}
  \hspace{-0.16cm}
  \begin{subfigure}{\subfigwidth\textwidth}
    \centering
    \includegraphics[width=\linewidth]{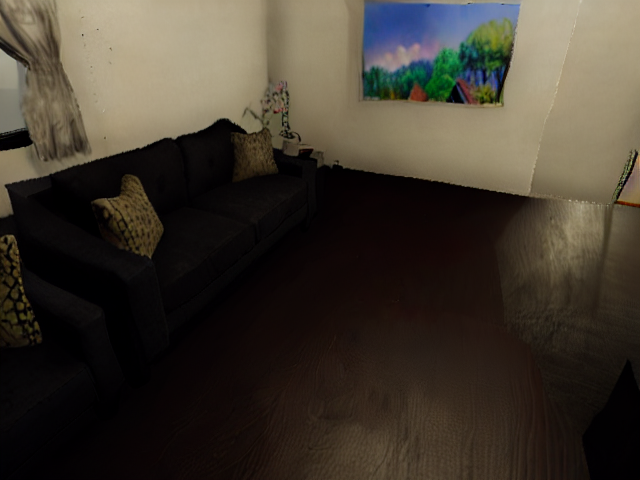}
   \captionsetup{font=tiny}
    \caption{Ours} 
    \label{subfig:06_Search_house}
  \end{subfigure}
  \vspace{0.2cm}
  \caption{Examples of the generated image of the EgoPlan in VirtualHome. We can find that in some hand reconstruction and direction understanding scenes, the model without introducing optical flow prior information often performs poorly.} 
  \label{fig:examples_appendix}
\end{figure}

\begin{figure}[t]
  \centering
  \begin{subfigure}{0.30\textwidth}
    \centering
    \includegraphics[width=\linewidth]{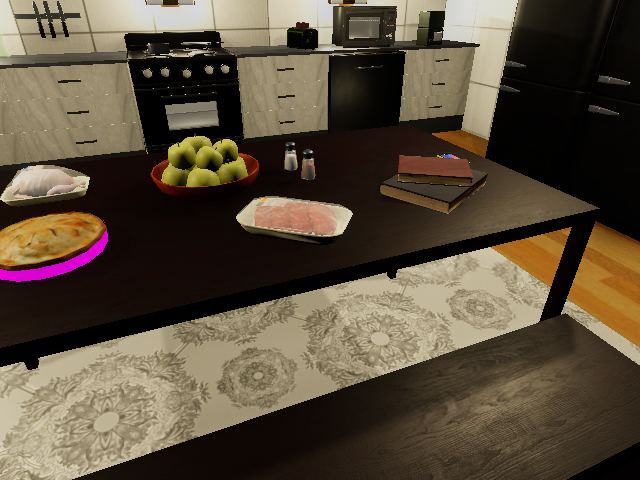}
    \label{subfig:01_Follow_Dutch}
  \end{subfigure}
  \hspace{-0.16cm}
  \begin{subfigure}{0.30\textwidth}
    \centering
    \includegraphics[width=\linewidth]{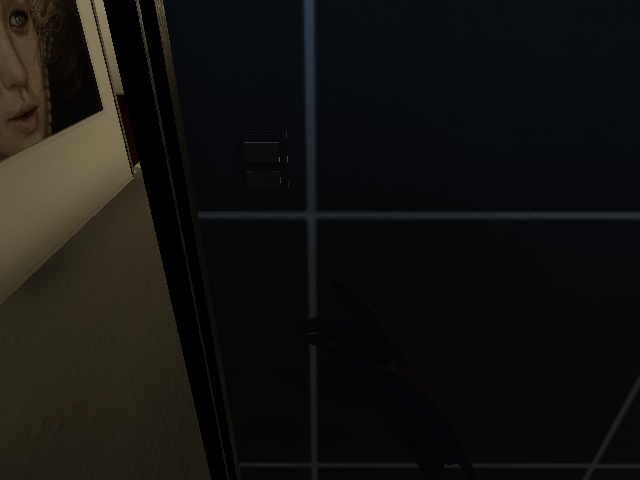} 
    \label{subfig:02_Hitch_horse}
  \end{subfigure}
  \hspace{-0.16cm}
  \begin{subfigure}{0.30\textwidth}
    \centering
    \includegraphics[width=\linewidth]{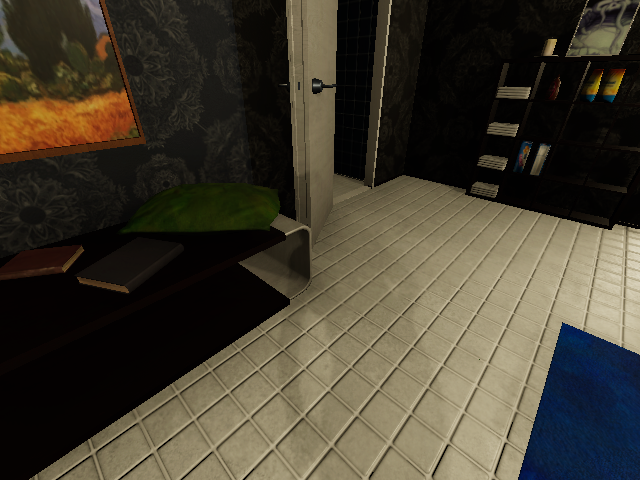}
    \label{subfig:03_Go_to_shed}
  \end{subfigure}
    \begin{subfigure}{0.30\textwidth}
    \centering
    \includegraphics[width=\linewidth]{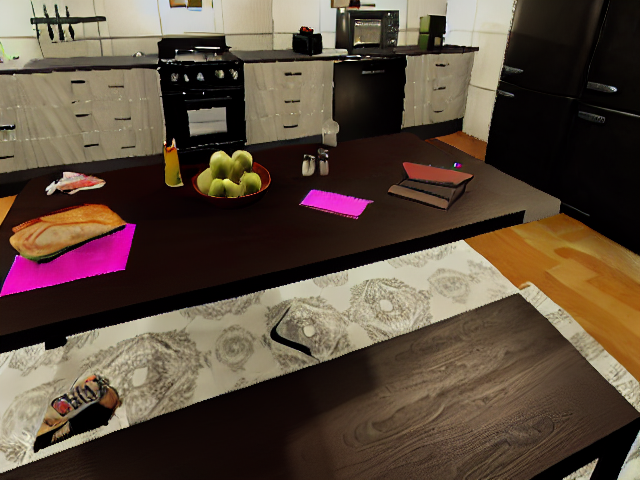}
    \captionsetup{font=tiny}
    \caption{grasp the cutlets}
    \label{subfig:04_Choose_weapon}
  \end{subfigure}
  \hspace{-0.16cm}
  \begin{subfigure}{0.30\textwidth}
    \centering
    \includegraphics[width=\linewidth]{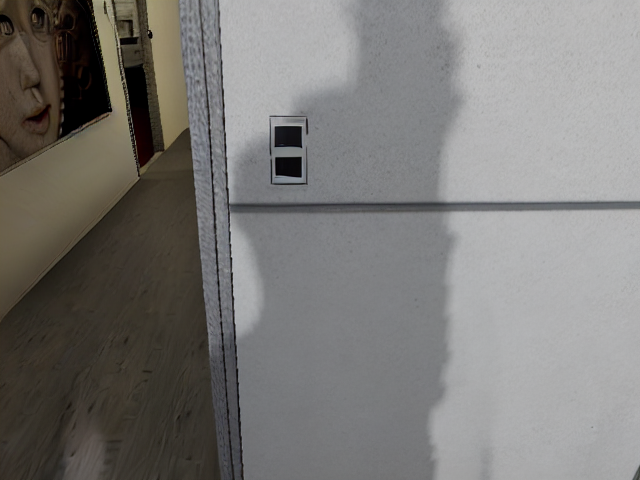}
    \captionsetup{font=tiny}
    \caption{power up the lightswitch}
    \label{subfig:05_Protect_Dutch}
  \end{subfigure}
  \hspace{-0.16cm}
  \begin{subfigure}{0.30\textwidth}
    \centering
    \includegraphics[width=\linewidth]{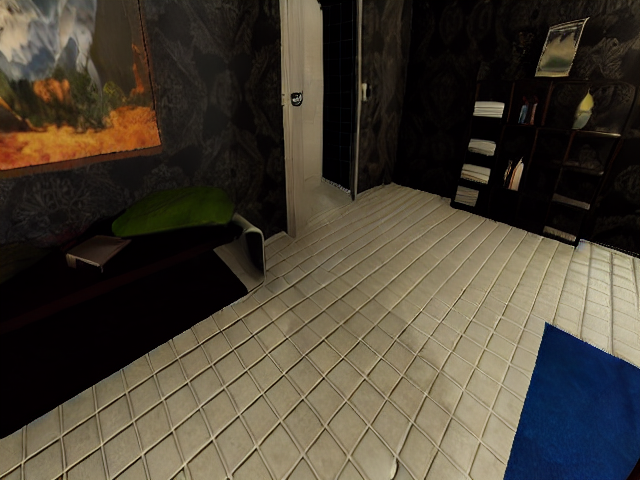}
    \captionsetup{font=tiny}
    \caption{sit on the bench}
    \label{subfig:06_Search_house}
  \end{subfigure}
  
  \begin{subfigure}{0.30\textwidth}
    \centering
    \includegraphics[width=\linewidth]{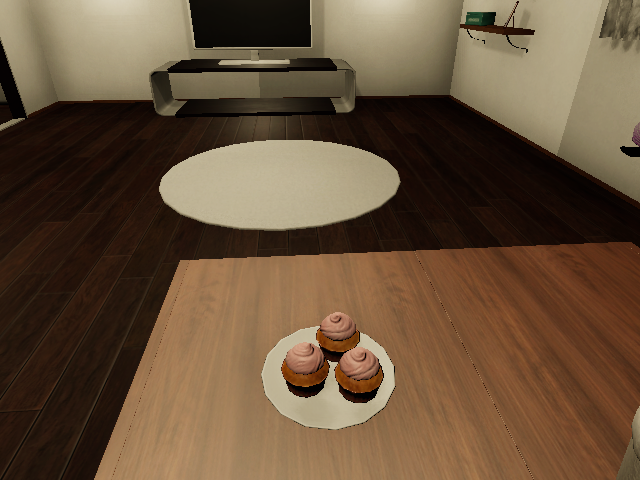}
    \label{subfig:06_Search_house}
  \end{subfigure}
  \hspace{-0.16cm}
    \begin{subfigure}{0.30\textwidth}
    \centering
    \includegraphics[width=\linewidth]{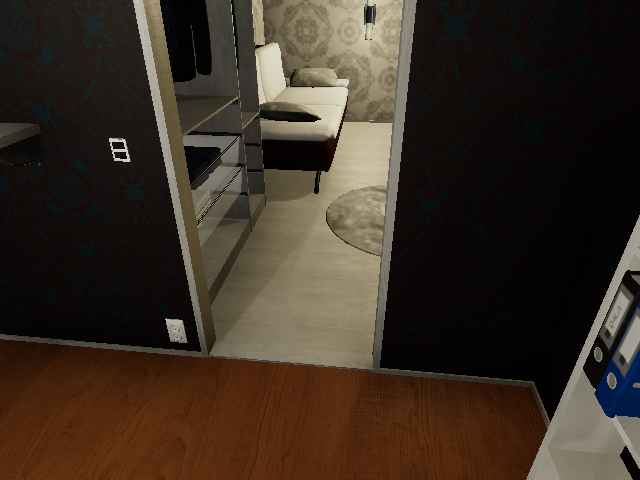}
    \label{subfig:06_Search_house}
  \end{subfigure}
    \hspace{-0.16cm}
   \begin{subfigure}{0.30\textwidth}
    \centering
    \includegraphics[width=\linewidth]{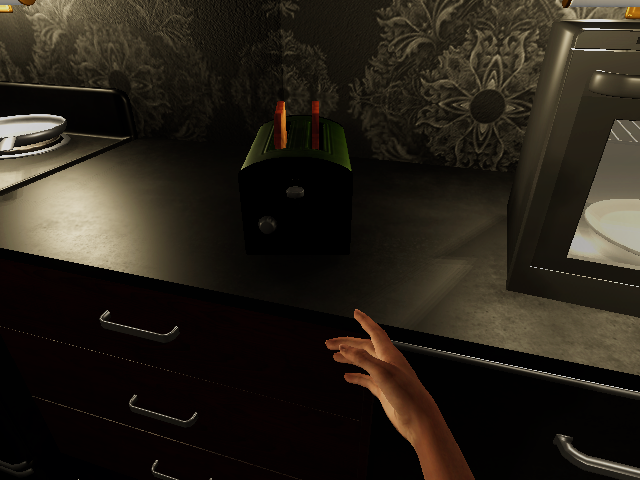}
    \label{subfig:06_Search_house}
  \end{subfigure}
  \begin{subfigure}{0.30\textwidth}
    \centering
    \includegraphics[width=\linewidth]{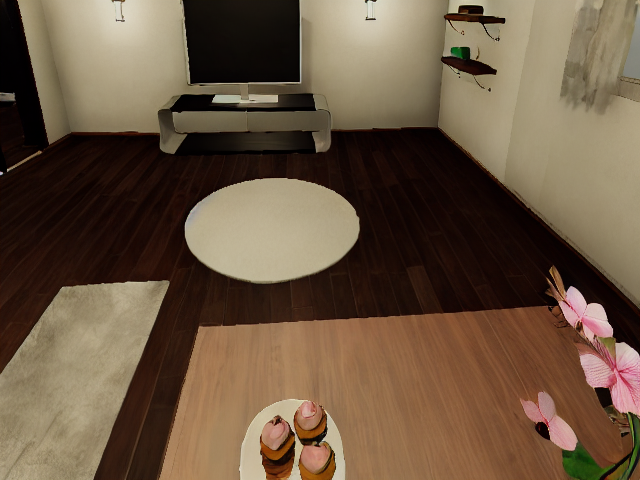}
    \captionsetup{font=tiny}
    \caption{stand from sofa}
    \label{subfig:06_Search_house}
  \end{subfigure}
  \hspace{-0.16cm}
  \begin{subfigure}{0.30\textwidth}
    \centering
    \includegraphics[width=\linewidth]{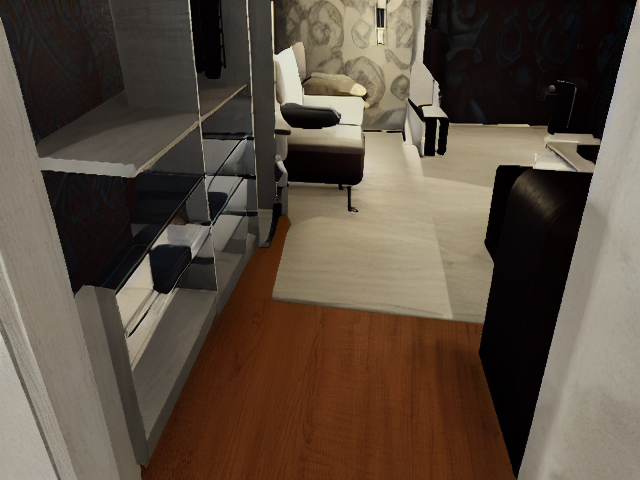}
    \captionsetup{font=tiny}
    \caption{walk through}
    \label{subfig:06_Search_house}
  \end{subfigure}
  \hspace{-0.16cm}
   \begin{subfigure}{0.30\textwidth}
    \centering
    \includegraphics[width=\linewidth]{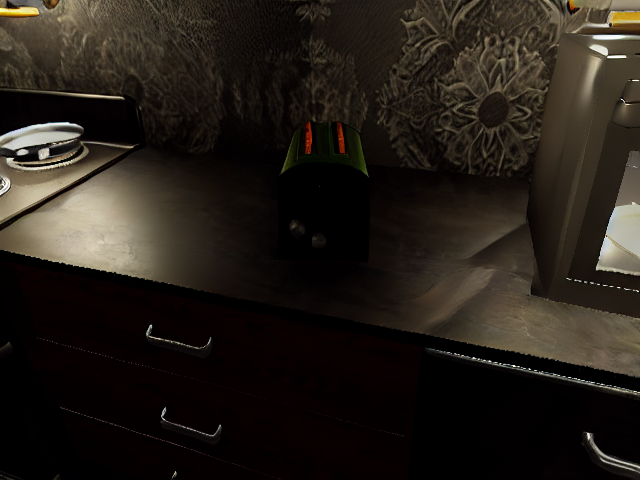}
    \captionsetup{font=tiny}
    \caption{power up toaster}
    \label{subfig:06_Search_house}
  \end{subfigure}
  \vspace{0.2cm}
  \caption{Examples of the generated image subgoals. The first and third rows is the original image, and the second and forth rows is the image subgoal generated based on the text subgoal.} 
  \label{fig:image_subtask_appendix2}
\end{figure}

\subsection{User study of subgoal image generation}
We also conduct a user study on the image generation of the subgoal. A total of $8$ users evaluated whether the generated image met the criteria of the subgoal described in the text. Each user evaluates $100$ generated images for each action, and the evaluation results are shown in Table \ref{goal_user_study}. The results show that most of the generated subgoal images can represent the meaning of the text subgoals. More examples of generating figures can be seen in Figure \ref{fig:image_subtask_appendix2}
\begin{table}[ht]
\centering
\vspace{-0.5cm}
\caption{User study for the subgoal generation. The user score is the percentage of images that users consider to meet the criteria out of the total 100 images.} 
\vspace{0.5cm}
\label{goal_user_study}
\tiny
\label{tab:open_ended_subtask_list}
\resizebox{\textwidth}{!}{
\begin{tabular}{ccccccc}
\toprule
  & Close & Drink & Grab & Open & Put back & Put in\\ \midrule
Mean user score($\%$) & 66.5 & 71.75 & 55 & 66.375 & 62.125 & 64.625 \\ \midrule
 &  Sit & Stand up & Switch off & Switch on & Walk through &  \\ \midrule
   Mean user score($\%$)& 79.875 & 78.75 & 73.375 & 77.875 & 79& \\
 \bottomrule
 \end{tabular}
 }
 \end{table}

\end{document}